%% file: main.tex
\documentclass[twoside,11pt]{article}

\usepackage{amsmath, amsfonts, amsopn, mathrsfs, xcolor}
\usepackage{graphicx, epstopdf}
\usepackage{caption, subcaption, booktabs, float, multirow}
\usepackage{algorithm, algorithmic}
\usepackage{enumerate}
\usepackage{lipsum}
\usepackage{bm}
\usepackage{hyperref}
\usepackage{cleveref}

\usepackage{xurl} 
\usepackage{lineno}
\usepackage{soul}
\usepackage{fullpage}
\usepackage[auth-lg,affil-it]{authblk}
\usepackage[para]{footmisc}

\newcommand{\dif}{\mathrm{d}}
\newcommand{\tensor}[1]{\mathbf{#1}}

\newcommand{\mr}{\mathbb{R}}
\newcommand{\mc}{\mathbb{C}}

\newcommand{\bfx}{\bm{{x}}}
\newcommand{\bfy}{\bm{{y}}}

\title{Solving Seismic Wave Equations on Variable Velocity Models with Fourier Neural Operator}

\author[1,2]{Bian Li\thanks{\href{mailto:bil215@lehigh.edu}{bil215@lehigh.edu}} }
\author[2]{Hanchen Wang\thanks{\href{mailto:hanchen.wang@lanl.gov}{hanchen.wang@lanl.gov}}}
\author[2]{Shihang Feng\thanks{\href{mailto:shihang.feng@live.com}{shihang.feng@live.com}}}
\author[1]{Xiu Yang\thanks{\href{mailto:xiy518@lehigh.edu}{xiy518@lehigh.edu}}}
\author[3]{Youzuo Lin\thanks{\href{mailto:ylin@lanl.gov}{ylin@lanl.gov}}}
\affil[1]{Department of Industrial and Systems Engineering, Lehigh University}
\affil[2]{Theoretical Division, Los Alamos National Laboratory}
\affil[3]{Earth and Environmental Sciences Division, Los Alamos National Laboratory}

\begin{document}

\maketitle

\begin{abstract}
    In the study of subsurface seismic imaging, solving the acoustic wave equation is a pivotal component in existing models. The advancement of deep learning enables solving partial differential equations, including wave equations, by applying neural networks to identify the mapping between the inputs and the solution. This approach can be faster than traditional numerical methods when numerous instances are to be solved. Previous works that concentrate on solving the wave equation by neural networks consider either a single velocity model or multiple simple velocity models, which is restricted in practice. Instead, inspired by the idea of operator learning, this work leverages the Fourier neural operator (FNO) to effectively learn the \emph{frequency domain} seismic wavefields under the context of \emph{variable velocity models}. We also propose a new framework \emph{paralleled Fourier neural operator} (PFNO) for efficiently training the FNO-based solver given multiple source locations and frequencies. Numerical experiments demonstrate the high accuracy of both FNO and PFNO with complicated velocity models in the OpenFWI datasets. Furthermore, the cross-dataset generalization test verifies that PFNO adapts to out-of-distribution velocity models. Moreover, PFNO has robust performance in the presence of random noise in the labels. Finally, PFNO admits higher computational efficiency on large-scale testing datasets than the traditional finite-difference method. The aforementioned advantages endow the FNO-based solver with the potential to build powerful models for research on seismic waves.
\end{abstract}


\input{introduction}

\input{background}

\input{method}

\input{numerical}

\input{conclusion}

\input{append}

\section*{Acknowledgment}
This work was funded by the Los Alamos National Laboratory~(LANL) - Laboratory Directed Research and Development program under project number 20210542MFR.

\vskip 0.2in
\bibliographystyle{plain}
\bibliography{reference}

\end{document}

%% file: introduction.tex
\section{Introduction}
\label{sec:introduction}

The last decade has witnessed the fast development of deep learning thanks to the enormous growth in computational power as well as increasingly ample data that is accessible in many fields. According to the universal approximation theorem ~\cite{barron1994approximation,cybenko1989approximation,hornik1989multilayer}, neural networks are capable of approximating any nonlinear function satisfying certain smoothness conditions. Therefore, their high expressiveness allows them to capture highly sophisticated latent patterns from data. Furthermore, neural architectures can be exploited to learn the mappings between functions in the infinite-dimensional space. As shown in the work~\cite{chen1995universal}, neural networks can effectively approximate continuous nonlinear operators. This universal approximation theorem for operators leads to the development of DeepONet~\cite{lu2021learning}. Under the dome of operator learning, another string of research defines the Fourier neural operator (FNO) by a series of integral operators and activation functions~\cite{li2020fourier,li2020neural}.

The solution of a partial differential equation (PDE) is a type of complicated mapping that usually does not have an explicit form. In this work, we intend to solve the acoustic wave equation as wave equation forward modeling is the engine of many fields of study that are related to wave phenomena. We take the full waveform inversion (FWI) as an example, which is a powerful geophysical technique to reconstruct high-resolution subsurface seismic velocity models from seismic records~\cite{tarantola1984inversion}. It has been proven in the past decade that FWI is capable of meeting most of the industrial production demands~\cite{virieux2009overview}. However, the core of FWI involves numerically solving second-order PDEs (wave equations) twice in one iteration to calculate the gradient regarding the velocity update. The wavefield modeling step in FWI has a computational complexity of $O(N^3)$ for 2D cases and $O(N^4)$ for 3D cases. Usually, hundreds of such iterations would be needed before the algorithm converges to a satisfactory velocity model. Thus, a successful FWI implementation induces tremendous computational costs in real-data implementations, and such costs will even dramatically increase when dealing with large-scale datasets, such as the passive seismic inversion problems when a large number of sources are involved~\cite{wang2018microseismic,wang2020regularized}. As more and more seismic inversion frameworks~\cite{jin2021unsupervised,sun2021physics} perform forward modeling and the inversion concurrently, it is increasingly crucial to solving the forward problem accurately and efficiently.

It has been an emerging direction in which deep neural networks are applied to efficiently solve PDEs~\cite{berg2018unified,han2018solving,long2018pde,malek2006numerical}. This is motivated by numerous cases where the PDE admits no straightforward numerical methods or where traditional solvers are computationally expensive. Under such a context, the solution to the PDE, which is a function of time and location, is approximated by neural networks. PDE-constrained frameworks~\cite{raissi2019physics,wang2020deep,zhang2019quantifying,zhu2019physics} impose the governing equation(s) as a constraint in the training since purely data-driven models may yield solutions that violate the governing equation. Nevertheless, these frameworks require that the form of the PDE be explicitly known. As mentioned above, the wave equation is of particular interest as it governs the behavior of seismic waves. As an instance, the physics-informed neural network (PINN) was applied to solve the wave equation in the time domain~\cite{moseley2020solving,rasht2022physics}. Most importantly, the frameworks above only solve the wave equation on a fixed velocity model. This is an obvious limitation since abundant velocity models are present in reality, and we expect a model to solve the equation given different velocity models. Another direction aims to embed the PDE information into the network architecture. For example, \cite{sun2021physics,sun2020theory} harness a recurrent neural network (RNN) structure based on the finite difference of the wave equation. Particularly, the sequence of interest is formed by the wavefield values that are consecutive in time so that the time evolution of the wave directly depends on preceding states. However, the PDE information is not strictly imposed during the training. Consequently, it remains an open question whether the model can generalize to new datasets. Also, this framework relies on a regular mesh because it is built on a finite difference scheme.  

Aiming at building a mesh-independent solver incorporating variable velocity models, we hereby turn to operator-learning-based neural architectures for parametric PDEs. More specifically, the wave equation is considered a parametric PDE which takes the velocity as a variable parameter, and our goal is to map different velocity models to their corresponding wavefields, governed by the wave equation. Notably, this work solves the wave equation in the frequency domain. Due to the scattering effects caused by perplexing subsurface structures (reflected by the perplexity of the velocity model) and the causal relationship between time steps, the wavefields in the time domain can be extremely sophisticated. In comparison, the frequency domain wavefields exhibit benign properties. Specifically, the wavefields induced by different velocity models share common structures: they are composed of layers of ``circles" centered at the seismic source position. The deformed circles and other irregularities in a wavefield directly correspond to the features of the velocity model. In the frequency domain, the wave propagation is governed by the Helmholtz equation, which is the Fourier-transformed wave equation along the time axis. The Helmholtz equation has been solved with PINN in a series of papers~\cite{alkhalifah2021wavefield,song2021wavefield,song2021solving}. In addition, the U-Net~\cite{ronneberger2015u} is applied in \cite{cao2022accelerating} to perform fast wavefield interpolation of different frequencies. However, they all only solved the equation on a single velocity model. 

Inspired by the seminal work~\cite{li2020fourier}, we apply the Fourier neural operator (FNO) to learn the wavefields based on variable velocity models. There are previous works that intend to solve the Helmholtz equation using FNO~\cite{konuk2021physics,song2022high}, but these frameworks focus on other variable parameters, e.g., source location and frequency, rather than velocity. The most related work to ours is ~\cite{yang2021seismic}. This work applies FNO to perform time evolution of the wave equation based on multiple velocity models generated by the Gaussian random field, which may not yield geologically realistic features such as layered structures and fault zones. Additionally, the relatively low-frequency source wavelet has little sensitivity to the high wavenumber random scatters of the velocity models in their experiments. Hence, scattered events do not appear in the corresponding time domain wavefields under such a configuration, making the wavefield smooth and easier to predict than reality. Our work not only adapts FNO to approaching multiple realistic velocity models but also extends FNO so that it is capable of solving the Helmholtz equation in a multi-source-location and multi-frequency setting. To this end, we propose the paralleled Fourier neural operator (PFNO), where a group of FNOs approach data of different frequencies in parallel. This structure simplifies the mapping by avoiding taking the frequency as another variable parameter. The resulting wavefields predicted by PFNO capture irregularities in high-frequency data. In other words, PFNO preserves prediction accuracy while incorporating multiple sources and frequencies. 

To validate our models, we implemented comprehensive numerical tests over $6$ datasets from $3$ families in the OpenFWI datasets~\cite{deng2021openfwi}. The first objective is to verify whether FNO-based solvers make high-accuracy predictions on the wavefields. In the single source location and single frequency setting, an FNO provides satisfactory results from low-frequency to high-frequency data. However, it is not as impressive when multiple sources and frequencies are involved. In addition, a ForwardNet, which is based on an encoder-decoder architecture, is included in the comparison. In this case, PFNO outperforms FNO and ForwardNet in terms of both the mean-squared error and the wavefield visualization. Generalization is another important factor for model evaluation. To test the generalization of PFNO, we train a PFNO for each of the six datasets and test its performance on the other five ones. This cross-dataset generalization test shows that PFNO is able to generalize to unseen velocity models that share common features with the velocity models in the training set. Moreover, the wavefields are perturbed by random noise to investigate the robustness of the model. PFNO exhibits consistent performances under different levels of uncertainty. The last part of the numerical experiments addresses the computational efficiency of PFNO. We show that PFNO can predict the wavefield corresponding to a velocity model within milliseconds. Even after incorporating the training time, PFNO dominates the ``optimal" 9-point finite difference method~\cite{jo1996optimal}, a standard traditional numerical method for the Helmholtz equation, in the case of large-scale inferences. 

The paper is organized as follows. We first give some background on the Helmholtz equation and FNO in \Cref{sec:background}. \Cref{sec:method} follows to describe the FNO-based solver of the Helmholtz equation with variable velocity. Then, the performance of the models is discussed in detail in \Cref{sec:numerical}. Finally, we arrive at the discussion and the conclusion in the last two sections.

%% file: background.tex
\section{Background}
\label{sec:background}

\subsection{Helmholtz equation}
\label{subsec:helmholtz}
Prior to an in-depth discussion of the proposed method, we first briefly introduce the basics of the acoustic wave equation. Let $(x,~z) \in \mr \times \mr$ denote a point in a two-dimensional spatial domain. Specifically, $x$ is the horizontal distance, and $z$ is the depth. Then, the two-dimensional acoustic wave equation with a constant density is  
\begin{align}
    \nabla^2 \bm p(t, x, z) - \frac{1}{v(x, z)^2} \frac{\partial^2 \bm p(t, x, z)}{\partial t^2} = s(t, x_s, z_s),
    \label{eqn:time_wave}
\end{align}
where $\bm p$ denotes the acoustic pressure wavefield, $v(x,~z)$ is the velocity at location $(x,~z)$, and $s$ is the point source at location $(x_s,~z_s)$. The symbol $\nabla^2 := \frac{\partial^2 }{\partial x^2} + \frac{\partial^2 }{\partial z^2}$ is the Laplacian. \Cref{eqn:time_wave} describes the spatiotemporal pattern of how seismic energy spreads out from the source. However, it is challenging to learn the time domain wavefield, given complicated velocity models. Particularly, the subsurface structures reflected by the velocity models cause highly nonlinear scattering effects, making the pattern of the time domain wavefields exceedingly complicated. Moreover, the temporal derivative induces the dependence between snapshots of the wavefield at different time steps, which hinders the training of a neural network. Therefore, we focus on solving the Helmholtz equation:
\begin{align}
    \nabla^2 \bm u(x, z; \omega) + k^2 \bm u(x, z; \omega) = f(x_s, z_s; \omega),
    \label{eqn:helmholtz}
\end{align}
where $\omega$ is the angular frequency, and $k = \frac{\omega}{v}$ is the wavenumber. The quantify of interest is the frequency domain pressure wavefield $\bm u$. This equation can be viewed as the result of applying the Fourier transform to the time component while preserving the spatial information. Unlike $\cref{eqn:time_wave}$, the Helmholtz equation does not have a temporal derivative but revolves around frequency $\omega$. Hence, the wavefields of different frequencies are independent of each other.

\subsection{Fourier neural operator}
\label{subsec:fno}
The idea of the Fourier neural operator (FNO) stems from the Green's function of the linear differential operator $L$. Consider a linear PDE $L u = f$, where $u$ is the solution to the PDE, and $f$ is the external force.  The corresponding Green's function $G(\cdot,~\cdot)$ then gives the solution as an integral,
\begin{align}
    u(\bfx) = \int_{\mathcal{D}} G(\bfx, \bfy) f(\bfy) \dif \bfy,
    \label{eqn:green_integral}
\end{align}
where $\bfx, \bfy \in \mathcal{D} \subset \mr^d$. If there exists a function $g$ such that $G(\bfx,~\bfy) = g(\bfx - \bfy)$, \cref{eqn:green_integral} reduces to a convolution operation, and $G$ serves as the convolution kernel. For a parametric PDE, the Green's function also depends on the parameter. For example, the velocity $v$ is a parameter in the wave equation or the Helmholtz equation. In this framework, the parameter is assumed to be a function of $\bfx$. According to \cite{li2020fourier}, the kernel integral operator is defined by 
\begin{align}
    \big(K_{\theta}(v) h_i \big)(\bfx) = \int_{\mathcal{D}} \kappa_{\theta}(\bfx, \bfy, v(\bfx), v(\bfy)) \, h_i(\bfy) \dif \bfy,
    \label{eqn:kernel_integral_operator}
\end{align}
where $v$ is the PDE parameter, and the kernel integral operator is parameterized with a neural architecture with trainable parameters $\theta$ in the parameter space $\Theta$. Also, $h_i$ denotes a latent representation, updated by
\begin{align*}
    h_{i+1}(\bfx) = \sigma\left(W h_i(\bfx) + \big(K_{\theta}(v) h_i \big)(\bfx)\right).
\end{align*}
Here, $W$ denotes a linear transform. Importantly, the neural operator consists of a sequence of non-local kernel integral operators combined with the activation function $\sigma$ so that it is capable of approximating highly nonlinear operators. In this sense, the neural operator generalizes the Green's function. However, it requires more effort to approach the convolution integral in \cref{eqn:kernel_integral_operator}. Unlike~\cite{li2020neural} in which message passing in graph neural networks is employed, FNO leverages the convolution theorem which states convolution is equivalent to multiplication in the Fourier space. Therefore, FNO directly parameterizes the kernel function $\kappa_{\theta}$ in the Fourier space. Then, it Fourier-transforms the latent representations, performs the multiplication, and inverse Fourier-transforms the results back to the original space. In the implementation, the fast Fourier transform (FFT) is used for Fourier transform and inverse Fourier transform to preserve computational efficiency. 

As FNO maps the parameter $v$ to the solution $u$, it succeeds in solving PDEs with variable parameters. For comparison, frameworks such as~\cite{raissi2019physics} only solve one instance of the PDE after fixing the parameter. Another advantage of FNO compared with ~\cite{sun2020theory} is it is mesh-independent. This is because FNO takes the coordinate $\bfx$ as part of the input. When the resolution of the discretization changes, FNO generalizes to unseen locations. Our work leverages FNO to simulate the Helmholtz operator $\nabla^2 + k^2$ in \cref{eqn:helmholtz}, mapping velocity $v$ to the solution.

%% file: method.tex
\section{Method}
\label{sec:method}

\begin{figure}[!ht]
    \centering
    \includegraphics[width=0.75\textwidth]{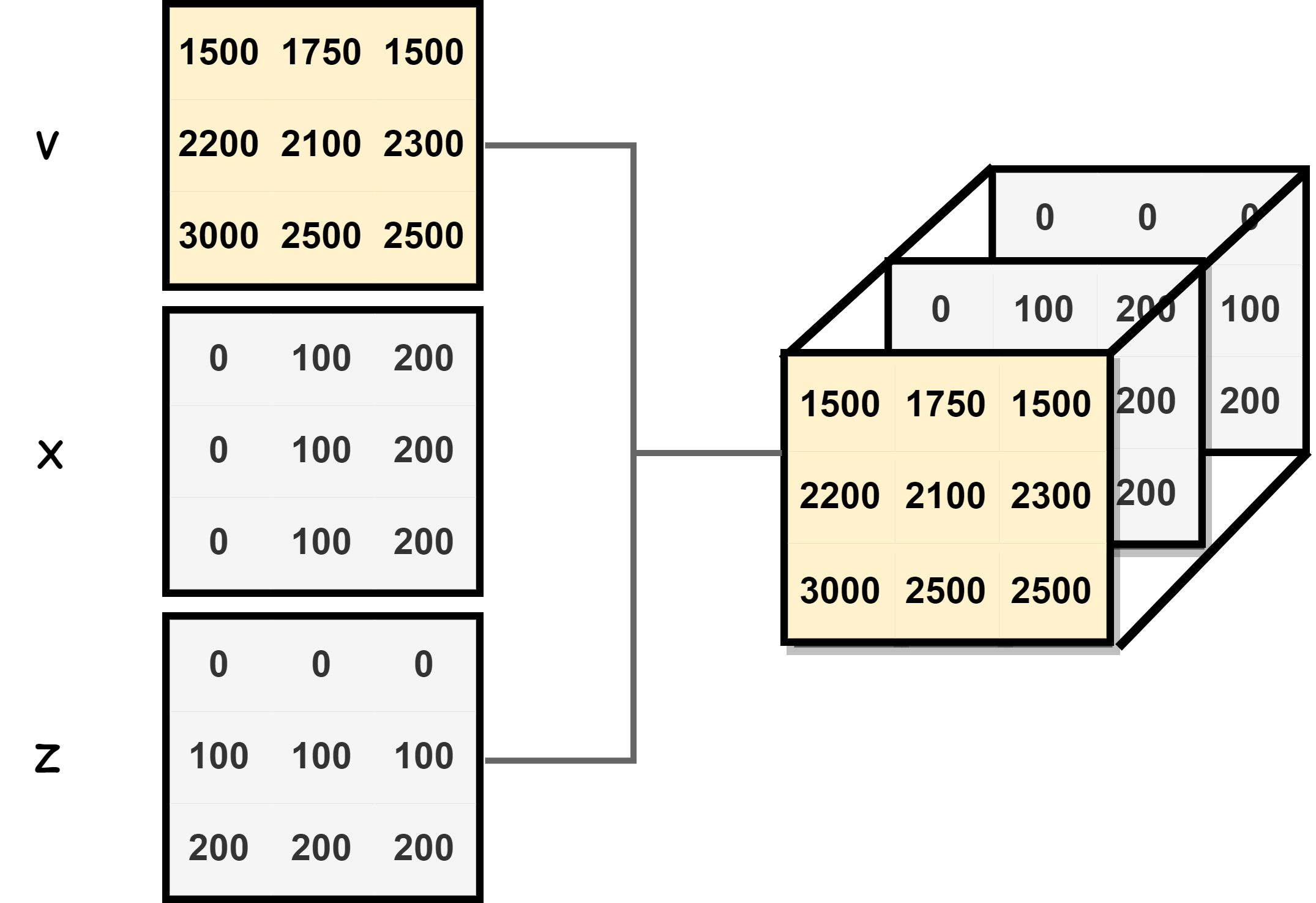}
    \caption{Data shape: an example of $n_z = n_x = 3$ with $\Delta x = \Delta z = 100$.}
    \label{fig:data_shape}
\end{figure}

Given multiple velocity models, we have the data denoted by $\{\tensor V^i,~\bm u^i \}_i^N$, wherein $\tensor V^i \in \mr^{n_z \times n_x}$ is a velocity model sample on a $n_z \times n_x$ grid. Of note, the location is represented by $\bfx = (x,~z)$. In particular, $x$ and $z$ denote the horizontal offset and the depth, respectively. The label $\bm u^i \in \mc^{n_z \times n_x}$ is the frequency domain pressure wavefield. Our goal is to solve the Helmholtz equation numerically by learning the wavefield with FNO. For convenience, we denote the shape of a data sample by $[n_z,~n_x]$. The velocity values are given on coordinates $(x,~z)$, $x$ and $z$ each giving a coordinate sample of the shape $[n_z,~n_x]$. Since FNO incorporates the coordinates into its input, a complete input is a 3-dimensional tensor of the shape $[n_z,~n_x,~3]$, as illustrated in \Cref{fig:data_shape}. It is worth mentioning that the grid can be irregular as long as the velocity values match the coordinates. The last dimension contains $v,~x,~z$ values, making it an analog to the dimension of channels as in image data. Specifically, we have $3$ ``channels" here. 

The forward modeling is formulated by the following optimization problem:
\begin{align*}
    \min_{\theta \in \Theta} \quad \frac{1}{N}\sum_{i=1}^N \left\Vert \bm \hat{\bm u}_\theta^i -  \bm u^i \right\Vert_2,
\end{align*}
with $\hat{\bm u}_\theta^i = \text{FNO}(\tensor T^i; \theta)$ and $\tensor T^i = \tensor V^i \oplus \tensor X^i \oplus \tensor Z^i$, where $\oplus$ denotes concatenation. 

\begin{figure*}[!ht]
    \centering
    \includegraphics[width=0.75\textwidth]{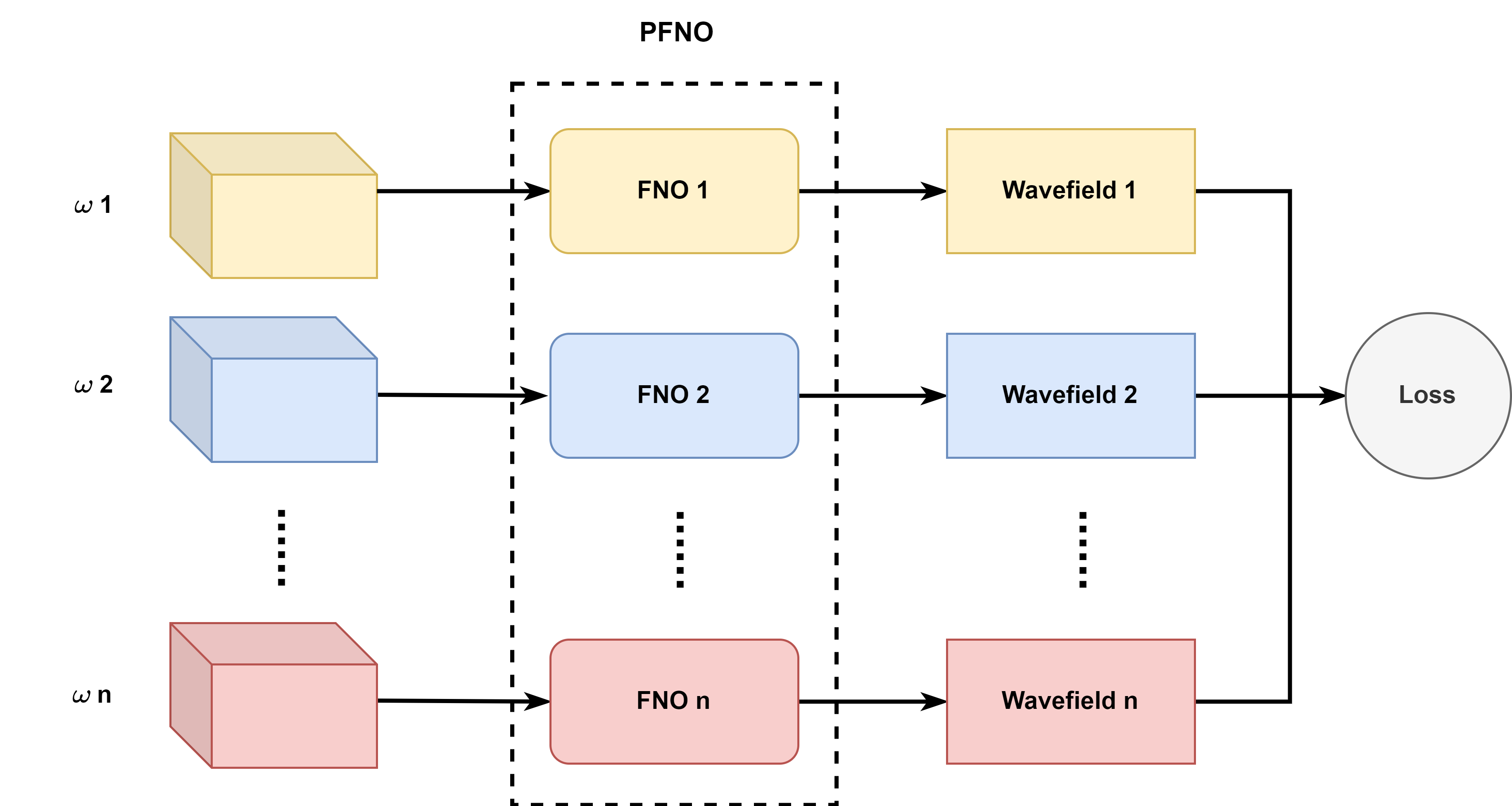}
    \caption{Structure of PFNO: each FNO receives data of only one frequency.}
    \label{fig:PFNO}
\end{figure*}

In real-life applications, multiple sources at different locations are usually required. Seismic data simulated by multiple sources has a better representation (or illumination) of the velocity model compared to the single source case because different source locations provide various incident waves that illuminate the same spatial position. Additionally, multi-frequency simulation is essential as well. Lower frequency components in the data are only sensitive to the longer wavelength parts of the velocity model, which represents the smoothed background. In comparison, higher frequencies are more sensitive to the shorter wavelengths, which represent the sharp scatters in the velocity model. Thus, the model should be capable of solving the Helmholtz equation with multiple source locations and multiple seismic frequencies in order to have an accurate seismic data representation of the velocity model. Accordingly, FNO incorporates $\bm x_s,~\omega$ as another two variable parameters. Regarding the source location, we use binary encoding: the source locations are encoded by $1$, whereas all other locations are $0$. On the other hand, the frequency is set as a constant across the spatial domain since it relies on neither the coordinates nor the velocity. As a result, a data sample ends up in the shape $[n_s,~n_\omega,~n_z,~n_x,~5]$. A naive way of training is to feed these 5-dimensional tensors into a single FNO. Nevertheless, different source locations and frequencies significantly increase the complexity of the mapping between the inputs and the wavefields. To partially counteract the increasing mapping complexity, we hereby propose the paralleled Fourier neural operator (PFNO), a structure in which we stack multiple FNOs in parallel and let each of them specialize in only one frequency. In other words, each FNO only takes two variable parameters, the source location and the velocity, instead of three. \Cref{fig:PFNO} depicts the data flow through PFNO. 

%% file: numerical.tex
\section{Numerical Experiments}
\label{sec:numerical} 

\subsection{Datasets}
\begin{figure}[!ht]
  \centering
  \subfloat[]{
  \includegraphics[width=0.33\textwidth]{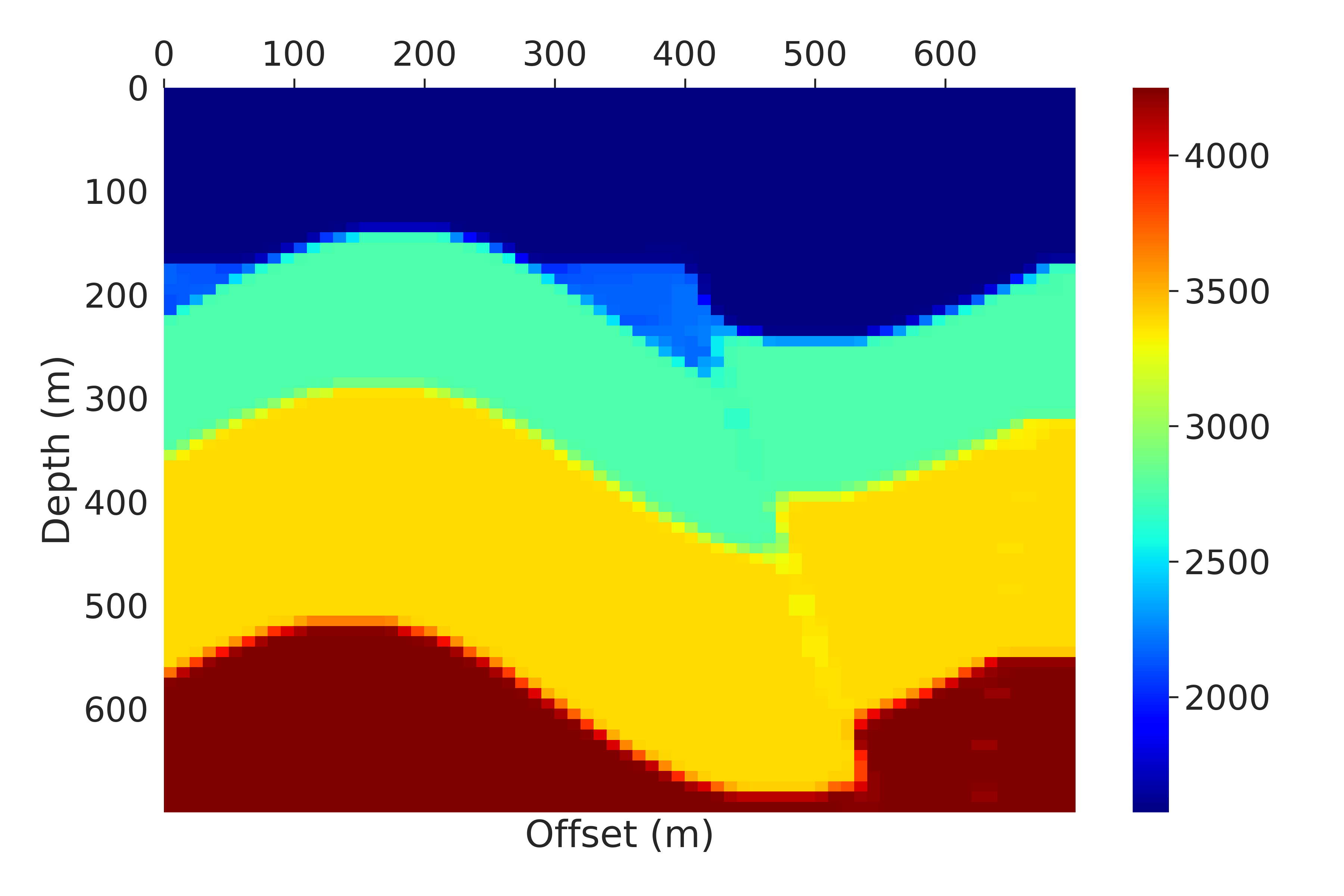}}
  \subfloat[]{
  \includegraphics[width=0.33\textwidth]{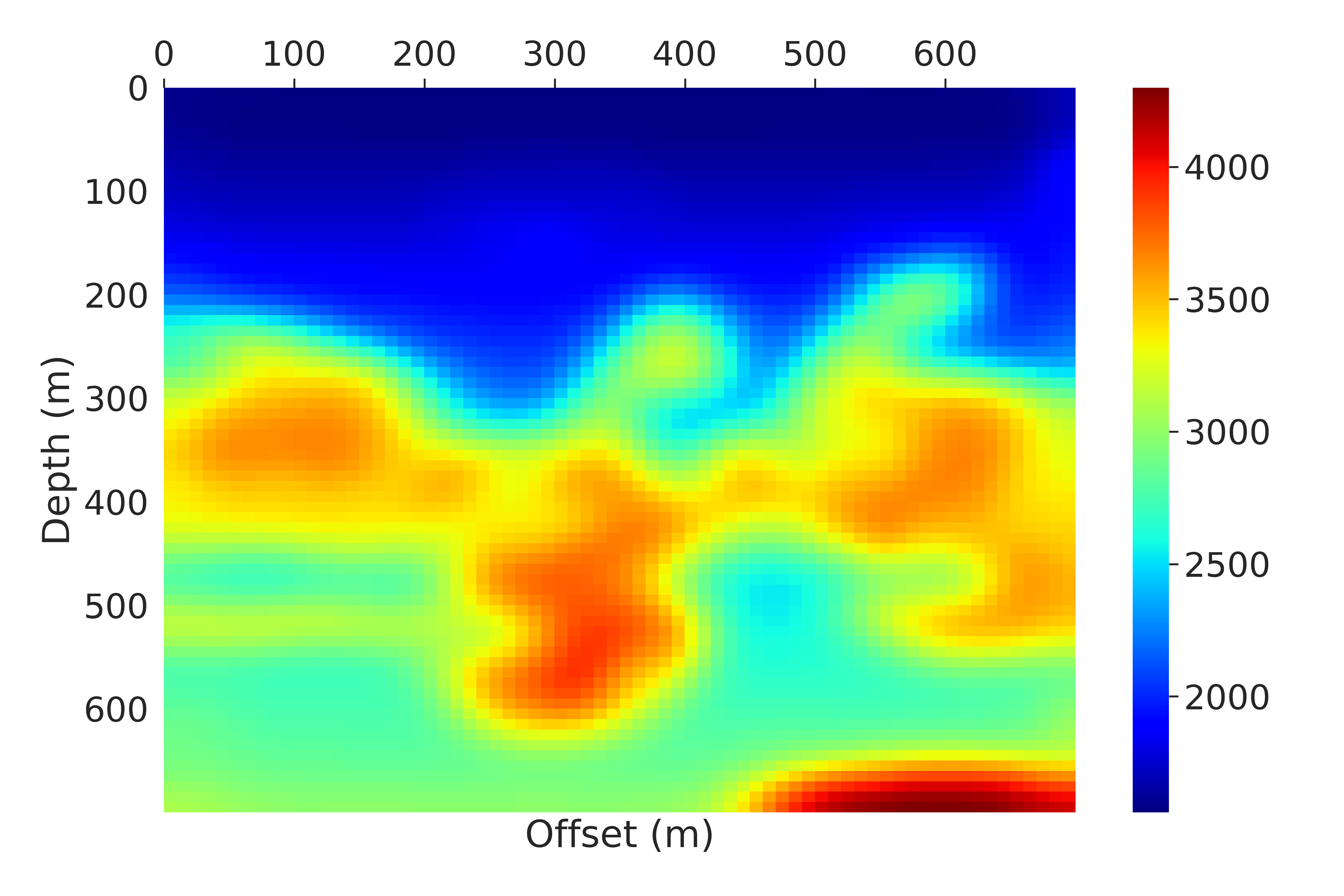}}
  \subfloat[]{
  \includegraphics[width=0.33\textwidth]{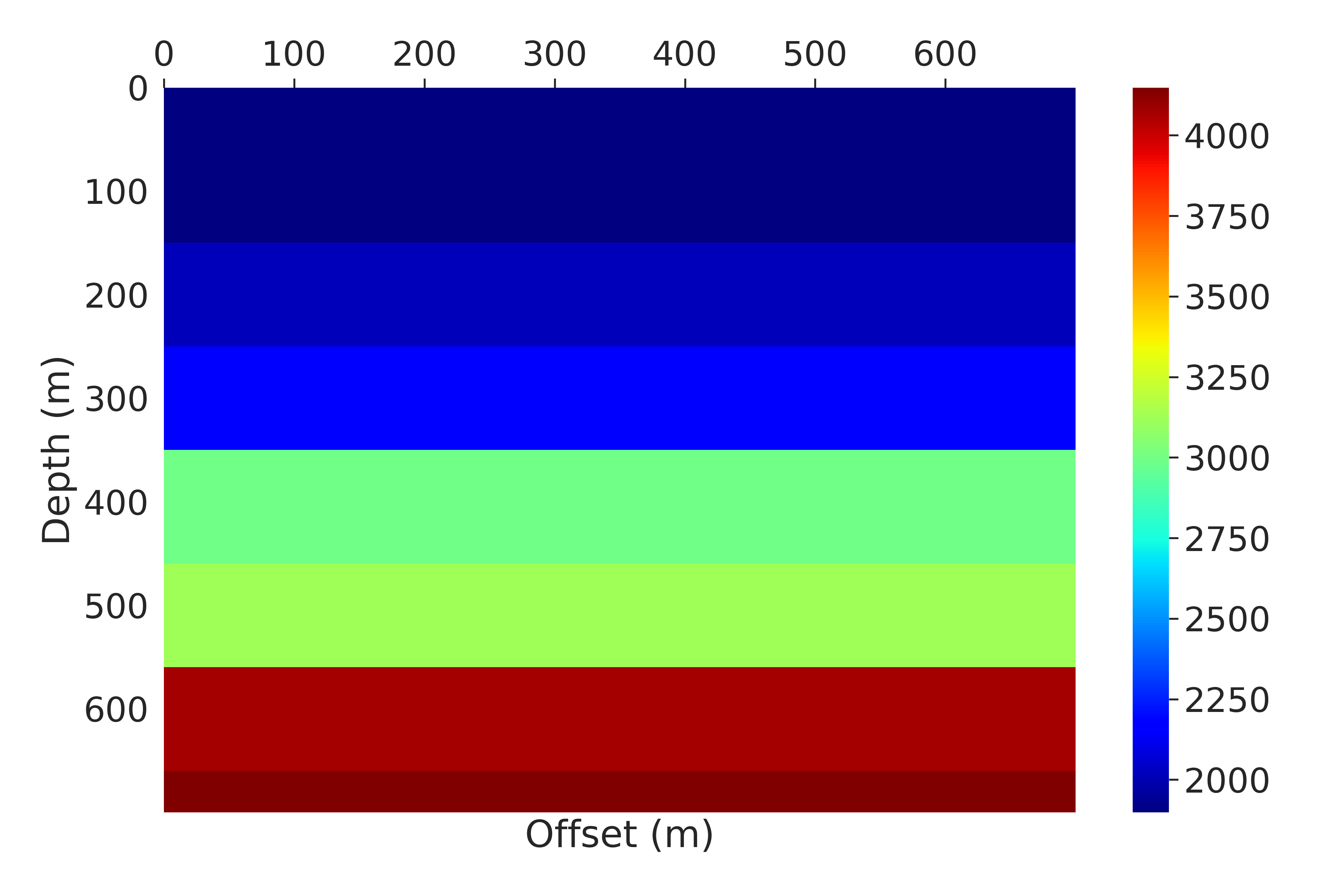}} \\
  \subfloat[]{
  \includegraphics[width=0.33\textwidth]{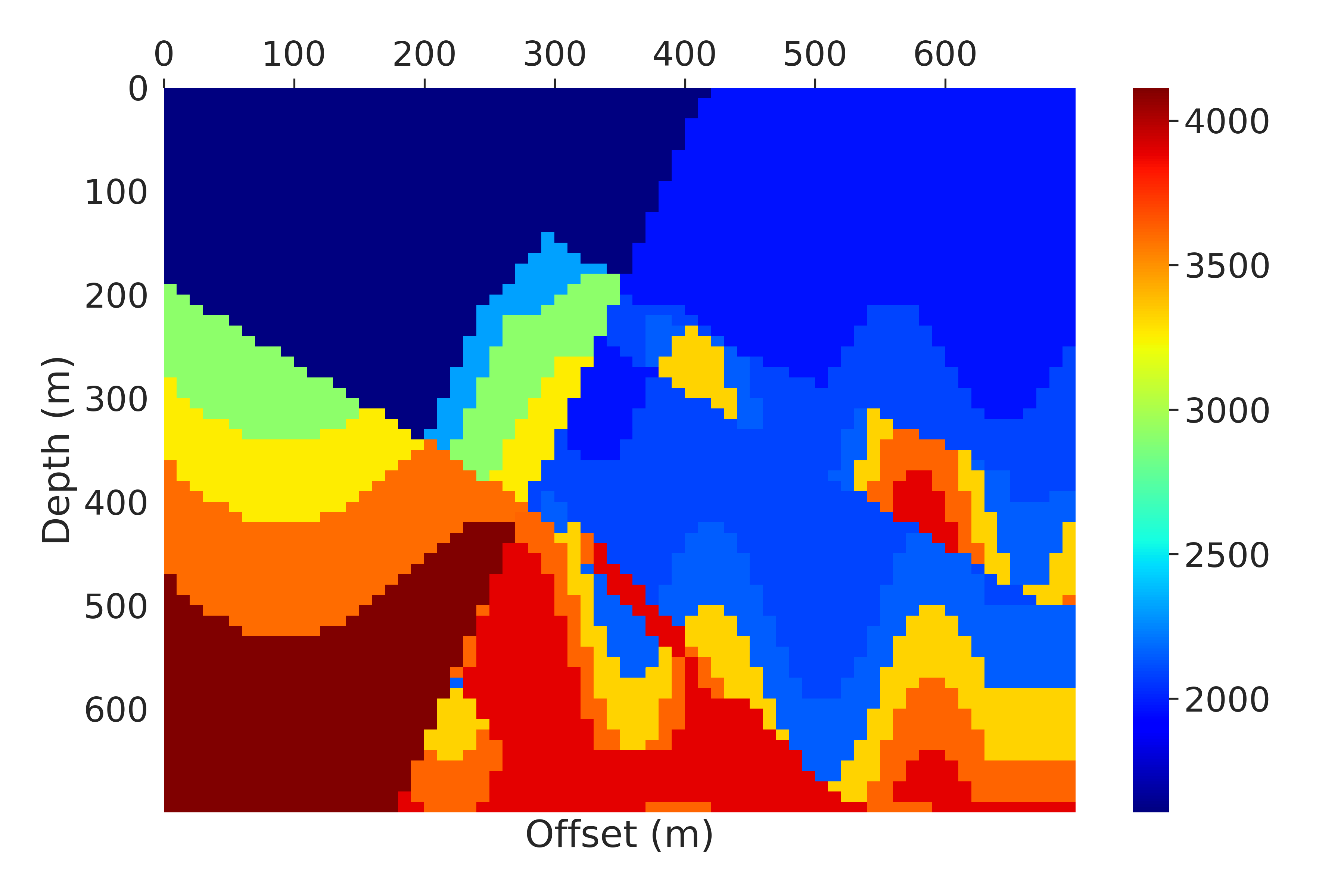}}
  \subfloat[]{
  \includegraphics[width=0.33\textwidth]{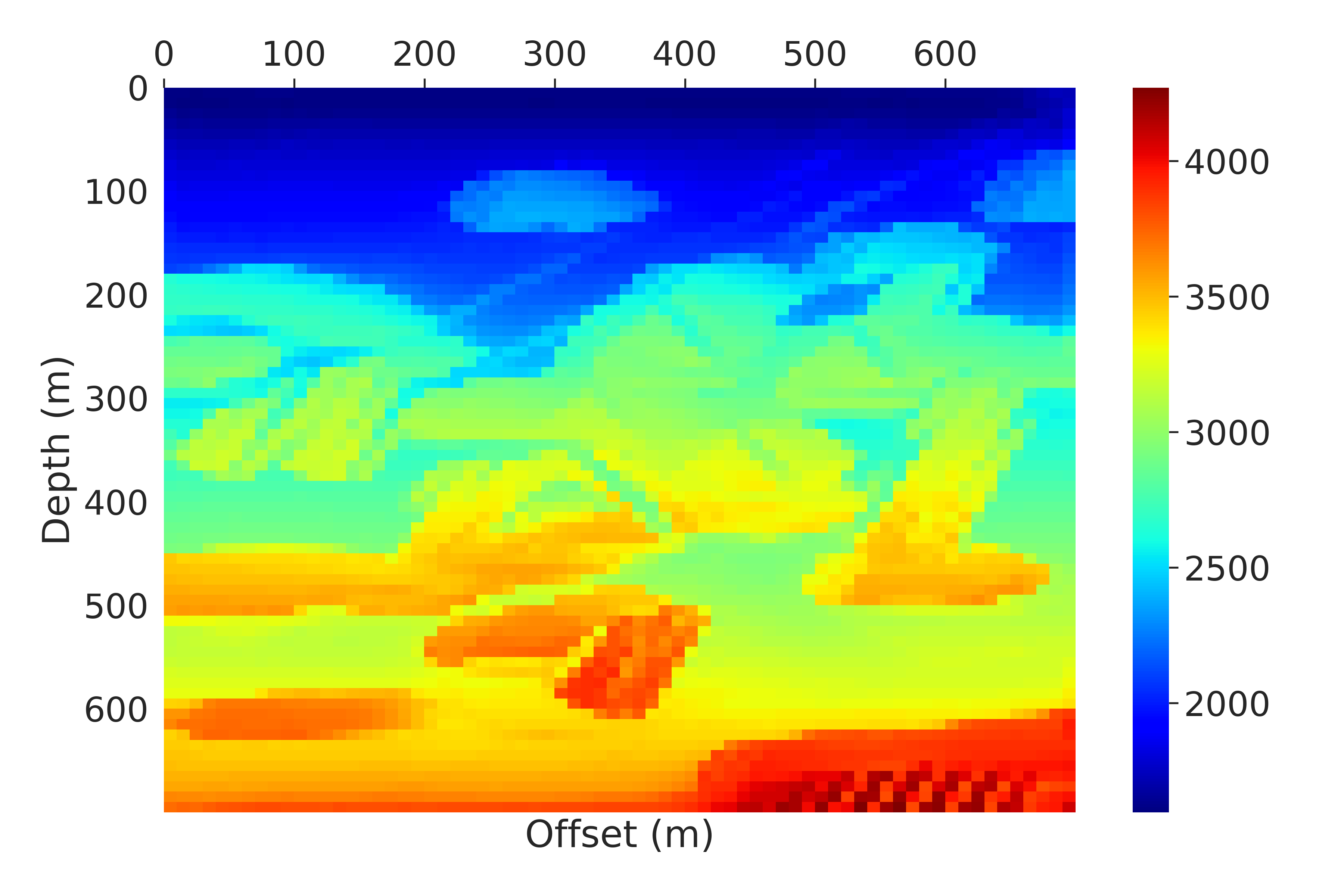}}
  \subfloat[]{
  \includegraphics[width=0.33\textwidth]{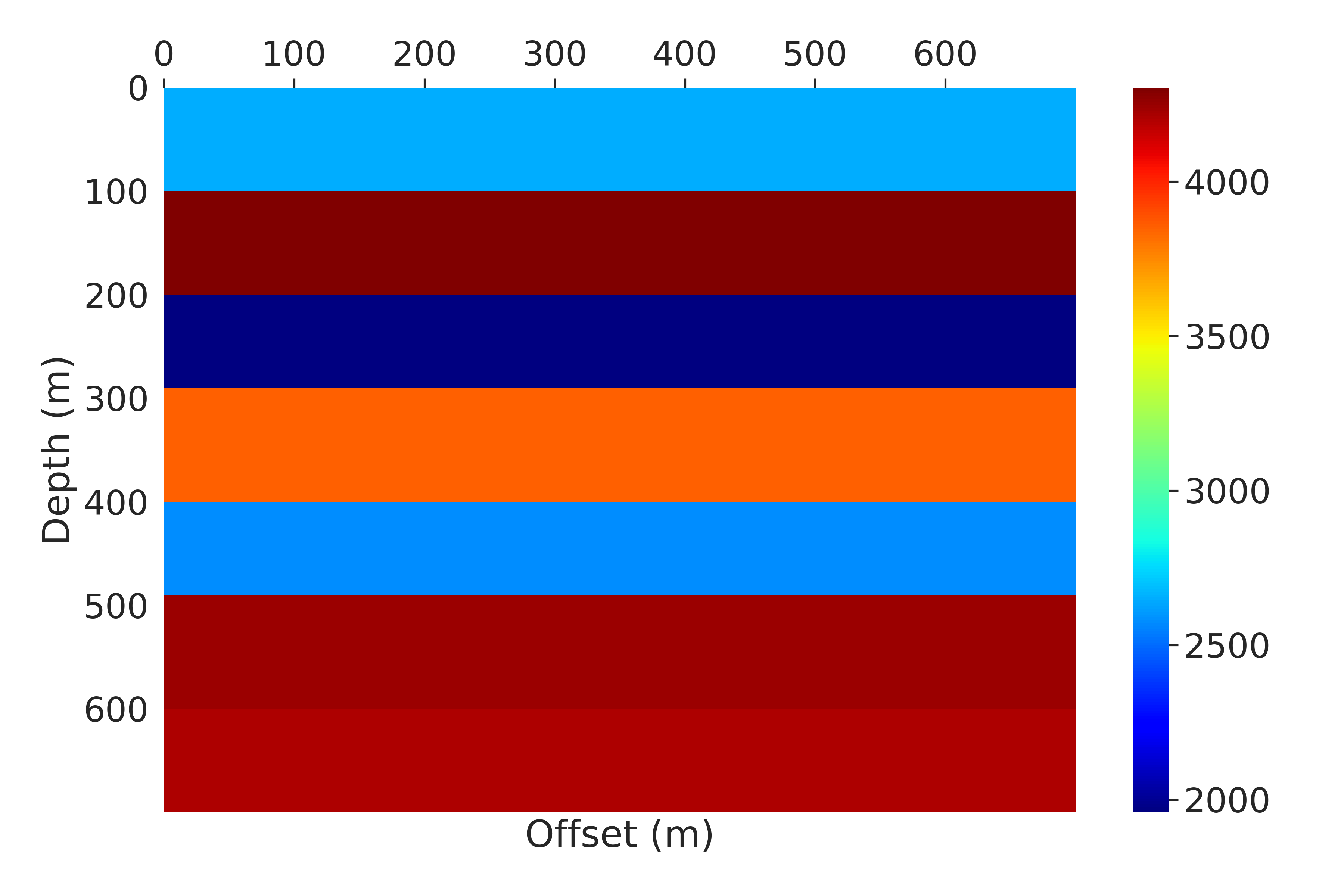}}
  \caption{Velocity models: (a) \emph{CurveFault-A}; (b) \emph{Style-A}; (c) \emph{FlatVel-A}; (d) \emph{CurveFault-B}; (e) \emph{Style-B}; (f) \emph{FlatVel-B}. } 
  \label{fig:velocity}
\end{figure}

In this section, we demonstrate the performance of FNO and PFNO using $6$ datasets from the dataset collection OpenFWI, namely \emph{CurveFault-A} \& \emph{CurveFault-B}, \emph{Style-A} \& \emph{Style-B}, and \emph{FlatVel-A} \& \emph{FlatVel-B}. More specifically, \emph{CurveFault-A} and \emph{CurveFault-B} contain discontinuities (geological faults) in the velocity models. These faults are induced by shifting the rock layers. Of note, \emph{CurveFault-B} exhibits more complicated changes of the velocities than \emph{CurveFault-A}. In contrast, \emph{FlatVel-A} and \emph{FlatVel-B} have flat interfaces without faults, but the velocity values are randomly distributed across different depths in \emph{FlatVel-B}, which increases the complexity. The velocity models in \emph{Style-A} and \emph{Style-B} are generated with style transfer, where the Marmousi model~\cite{brougois1990marmousi} serves as the style image while the content images are drawn from the COCO dataset~\cite{lin2014microsoft}. The difference between them is that the velocity models in \emph{Style-A} are smoother. In the above datasets, the spatial domain is defined on a $70 \times 70$ grid with $\Delta x = \Delta z = 10$ m, and all velocity values fall in the interval $[1500 \text{ m/s}, 4500 \text{ m/s}]$. \Cref{fig:velocity} displays examples from each of the datasets.

Based on the velocity models, we solve \cref{eqn:time_wave} by a high-order finite difference scheme~\cite{jin2021unsupervised} to obtain the labels (ground truth wavefields). Let $p_{x, z}^t := p(t, x, z)$. The temporal derivative approximated with the second-order central difference 
\begin{align*}
    \frac{\partial^2 p}{\partial t^2} \approx \frac{p_{x,z}^{t+\Delta t} - 2p_{x,z}^{t} + p_{x,z}^{t-\Delta t}}{\Delta t^2}.
\end{align*}
To guarantee accuracy, the Laplacian is discretized with the fourth-order central difference,
\begin{align*}
	\nabla^2 p 
	= \frac{\partial^2 p}{\partial x^2} + \frac{\partial^2 p}{\partial z^2} \nonumber  \approx \frac{1}{\Delta x^2} \sum_{i=-2}^{2} c_i p_{x+i, z}^t + \frac{1}{\Delta z^2} \sum_{i=-2}^{2} c_i p_{x, z+i}^t,
\end{align*}
where $c_{-1} = c_1 = -\frac{1}{12}$, $c_{-2} = c_2 = \frac{4}{3}$, $c_0 = -\frac{5}{2}$ are the coefficients. For convenience, we write $x+i := x + i \Delta x$ and $z+i := z + i \Delta z$. With $v_{x, z} := v(x, z), s_{x, z}^t := s(t, x, z)$, \cref{eqn:time_wave} can be discretized as follows, 
\begin{align*}
     s_{x,z}^t = &\frac{1}{\Delta x^2} \sum_{i=-2}^{2} c_i p_{x+i, z}^t + \frac{1}{\Delta z^2} \sum_{i=-2}^{2} c_i p_{x, z+i}^t \\ &- \frac{p_{x,z}^{t+\Delta t} - 2p_{x,z}^{t} + p_{x,z}^{t-\Delta t}}{v_{x, z}^2 \Delta t^2}.
\end{align*}
To satisfy the Courant-Friedrichs-Lewy (CFL) condition for numerical stability, we take $\Delta t = 0.001$ and let the finite difference scheme solve to $t = 1$. The source is generated by a Ricker wavelet whose peak frequency is set at $15$Hz. Moreover, $120$ absorbing layers with damping parameters are applied around the spatial domain to avoid reflection artifacts. Lastly, we apply the Fourier transform to the $t$ dimension and convert the time domain wavefields to the frequency domain wavefields which will be the labels. Notably, the frequency domain wavefields are complex-valued. When training FNOs, we split the complex numbers into real and imaginary parts. The training loss is then obtained by averaging the losses from the two parts.    

\begin{figure}[!ht]
    \centering
    \includegraphics[width= 0.7\textwidth]{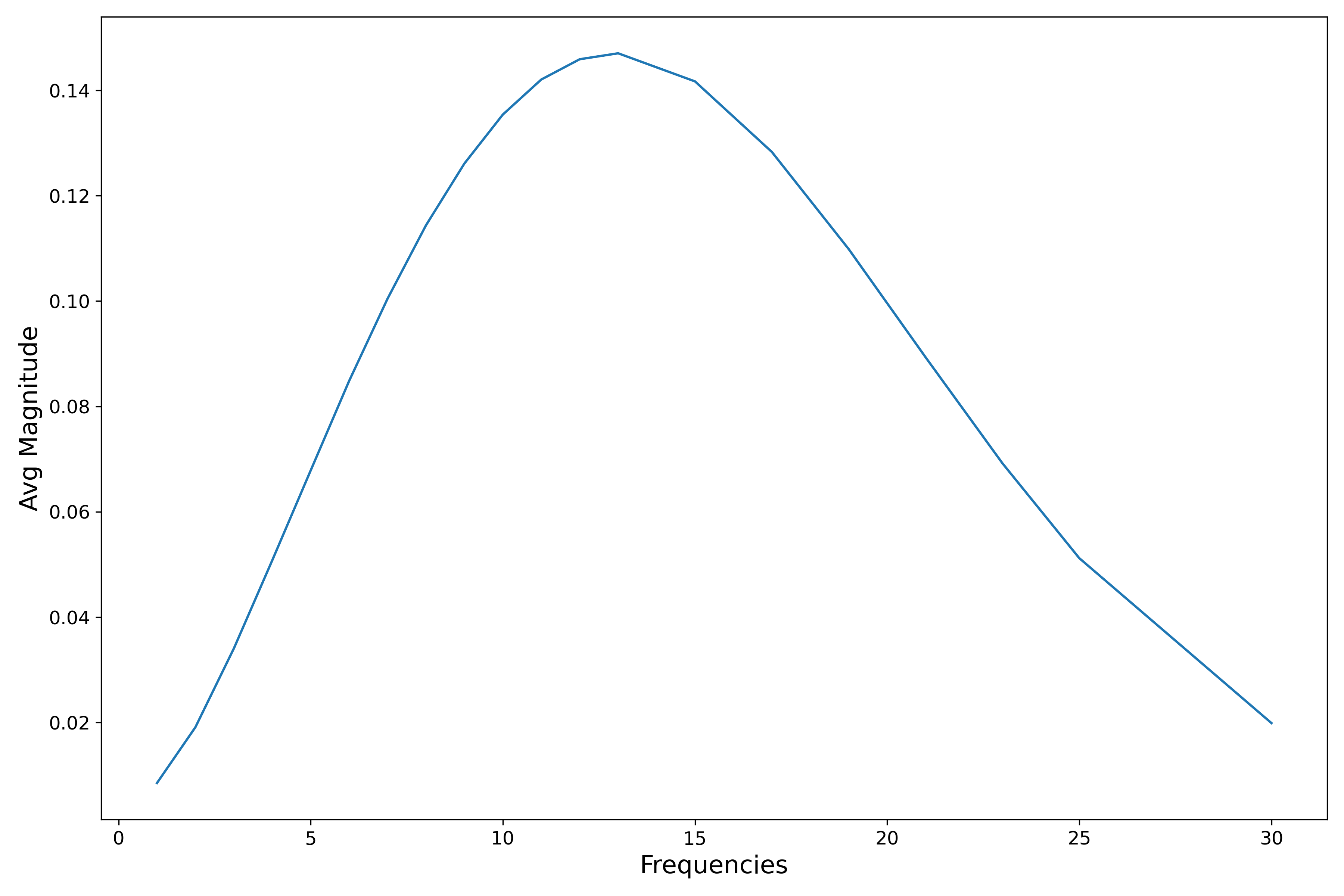}
    \caption{Spectrum of interest for \emph{CurveFault-A}: the magnitude is averaged over the dimension of space, source, and sample.}
    \label{fig:spectrum}
\end{figure}

As mentioned in \Cref{sec:method}, we intend to incorporate multiple source locations and frequencies. Particularly, the datasets have five sources at a depth of $10$m below the surface. Horizontally, they are distributed evenly from $0$m to $690$m. On the other hand, the selection of the frequencies is slightly more sophisticated. With a fast Fourier transform (FFT) from the time domain, there are $1,000$ frequencies corresponding to the $1,000$ time steps. However, it is computationally prohibitive to learn the wavefields of all the frequencies. Hence, a common technique in the field of signal processing is adopted in this work. Specifically, we attend to the magnitude of the wavefields since it reflects the energy level of the frequency domain signal. Then, we only focus on the spectrum that contributes to the majority of the total energy. Based on \emph{CurveFault-A}, we plot the magnitude against the frequencies in \Cref{fig:spectrum}. Over $90\%$ of the energy is contributed by the signals of the spectrum $1$Hz $\sim$ $30$Hz. Of note, the magnitude has an almost identical distribution over frequencies in other datasets. Therefore, this range is chosen to be the spectrum of interest. 

\subsection{Wavefield prediction}
\label{subsec:wavefield_prediction}
The motivation for applying FNO/PFNO is to obtain a numerical solver that can quickly map velocity models to their corresponding wavefields. In this part, we demonstrate the performance of the FNO-based models on the aforementioned datasets. In the following experiments, we use $15,000$ velocity models for training and $3,000$ for testing. Moreover, we allow $600$ epochs in the training with a batch size of $16$. The initial learning rate is set to $0.0016$ which is decayed by a factor of $0.5$ every $125$ epochs. For the optimization purpose, the AdamW optimizer~\cite{loshchilov2017decoupled} is applied with $\beta_1 = 0.9,~\beta_2 = 0.999$ and weight decay $0.0001$. Following the setting in~\cite{li2020fourier}, we admit $4$ Fourier layers in FNO and only $12$ Fourier modes for truncation. The other hyperparameter in FNO is the width of each Fourier layer. In PFNO, the dimensions are specified by the rule: $1)$ $32$ if the frequency is in $[1\text{Hz},~15\text{Hz}]$; $2)$ 64 if the frequency is in $[16\text{Hz},~25\text{Hz}]$; $3)$ 96 if the frequency is in $[26\text{Hz},~30\text{Hz}]$. This is consistent with the fact that the higher the frequency, the more complicated the wavefield. 

\begin{figure}[!ht]
  \centering
  \subfloat[]{
  \includegraphics[width= \textwidth, height= 0.35\textheight]{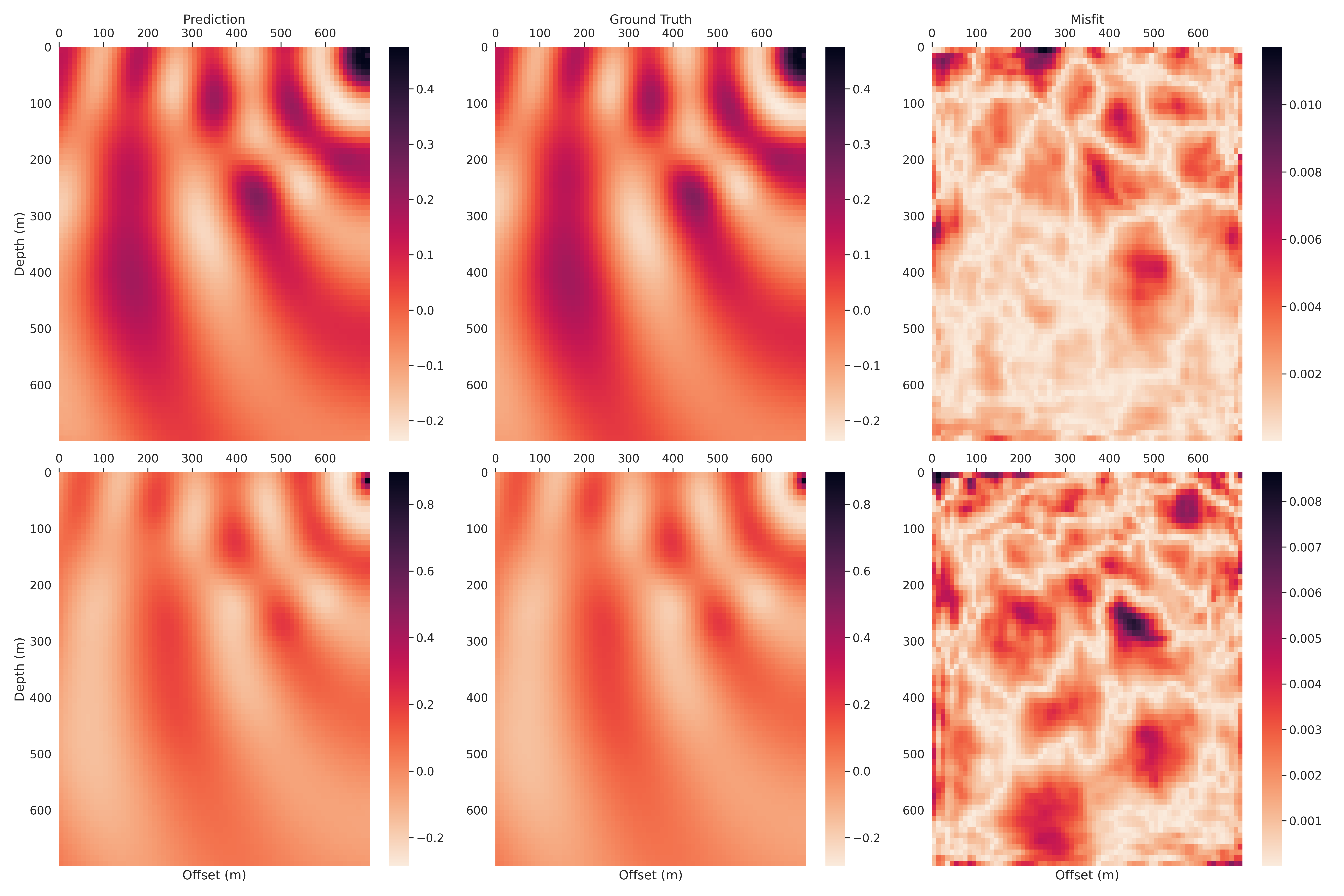}}\\
  \subfloat[]{
  \includegraphics[width= \textwidth, height= 0.35\textheight]{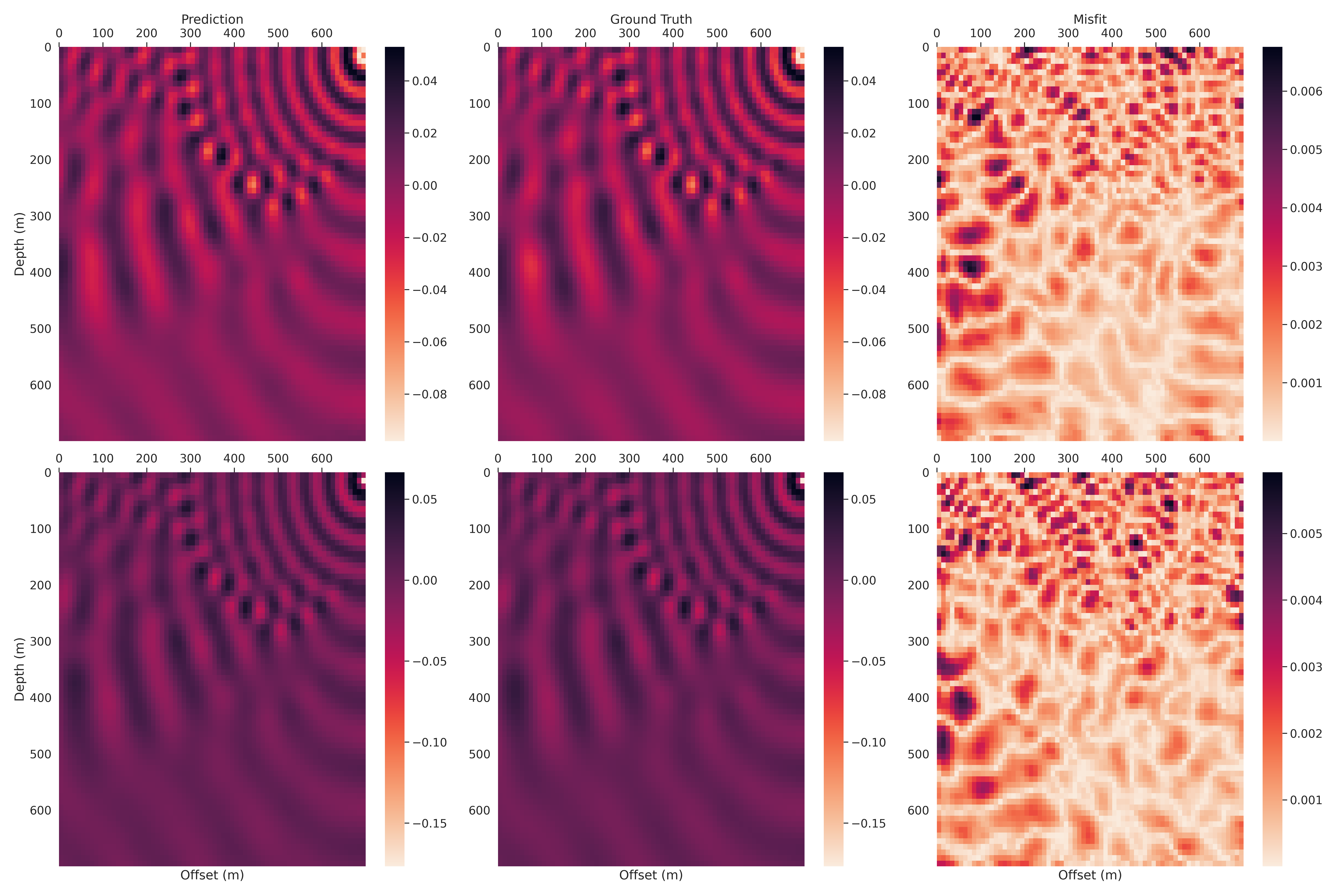}}
  \caption{Wavefield with a single source location and a single frequency based on a velocity model from \emph{CurveFault-A}: (a) $10$Hz; (b) $30$Hz; the first row of each sub-figure shows the real part and the second row shows the imaginary part.} 
  \label{fig:wavefield_freqs}
\end{figure}

Starting with a simple case of only one source location and one frequency, we train two FNOs based on the  \emph{CurveFault-A} dataset. In this experiment, the source is located in the right upper corner of the domain. For comparison purposes, the two FNOs are trained on data of $10$Hz and $30$Hz, respectively. In addition, the width is $32$ for the $10$Hz data, whereas we use $96$ for $30$Hz data in that it is much more challenging to fit. As \Cref{fig:wavefield_freqs} verifies, the wavefields of $30$Hz data have smaller wavelengths as well as more prominent scattering effects. Consequently, the FNO trained on $10$Hz data outperforms its counterpart in terms of the relative misfit (absolute difference between the prediction and the label). Additionally, the imaginary part has highly similar patterns to the real part regardless of the data frequency, but they differ in magnitude. 

The mapping between the inputs and the wavefield becomes significantly more complicated when multiple source locations and frequencies are incorporated. In our numerical experiments, we find that the frequency has a greater impact on the performance of FNO than the source location. Therefore, we propose the PFNO structure where multiple FNOs are paralleled. Each of these FNOs only approaches data of a single frequency (but still of multiple source locations). Hence, the source location and the velocity are the variable parameters for each FNO. In the numerical experiments, the datasets have the aforementioned $5$ source locations and the following 10 frequencies: $\{1, 3, 5, 7, 9, 12, 15, 19, 25, 30 \}\text{Hz}$. We choose smaller intervals for lower frequencies and larger intervals for higher ones in the experiments mainly to solve seismic imaging problems like FWI, where the lower frequency components are more important than the higher ones. It is because the lower frequencies represent the fundamental low-wavenumber part of subsurface kinematics, and lack of such information leads to the higher frequencies cycle-skipping problems~\cite{brossier2009seismic}.

\begin{figure}[!ht]
  \centering
  \subfloat[]{\label{fig:wavefields_CurveFault_A}
  \includegraphics[width= \textwidth, height= 0.22\textheight]{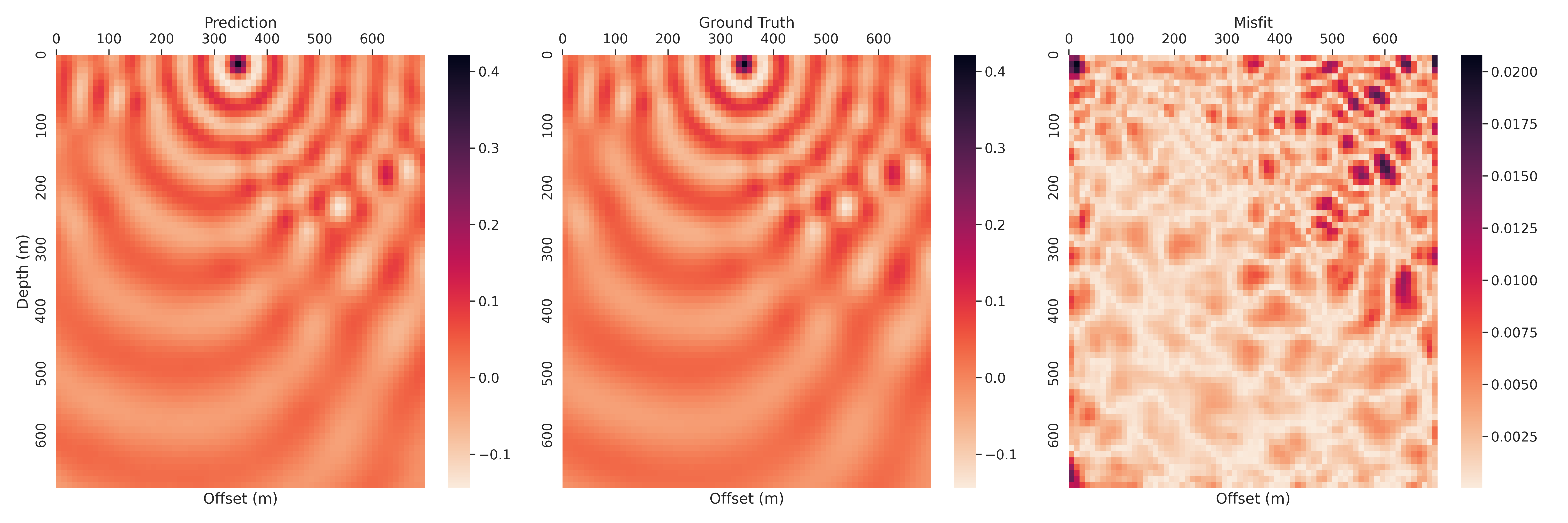}}\\
  \subfloat[]{\label{fig:wavefields_CurveFault_B}
  \includegraphics[width= \textwidth, height= 0.22\textheight]{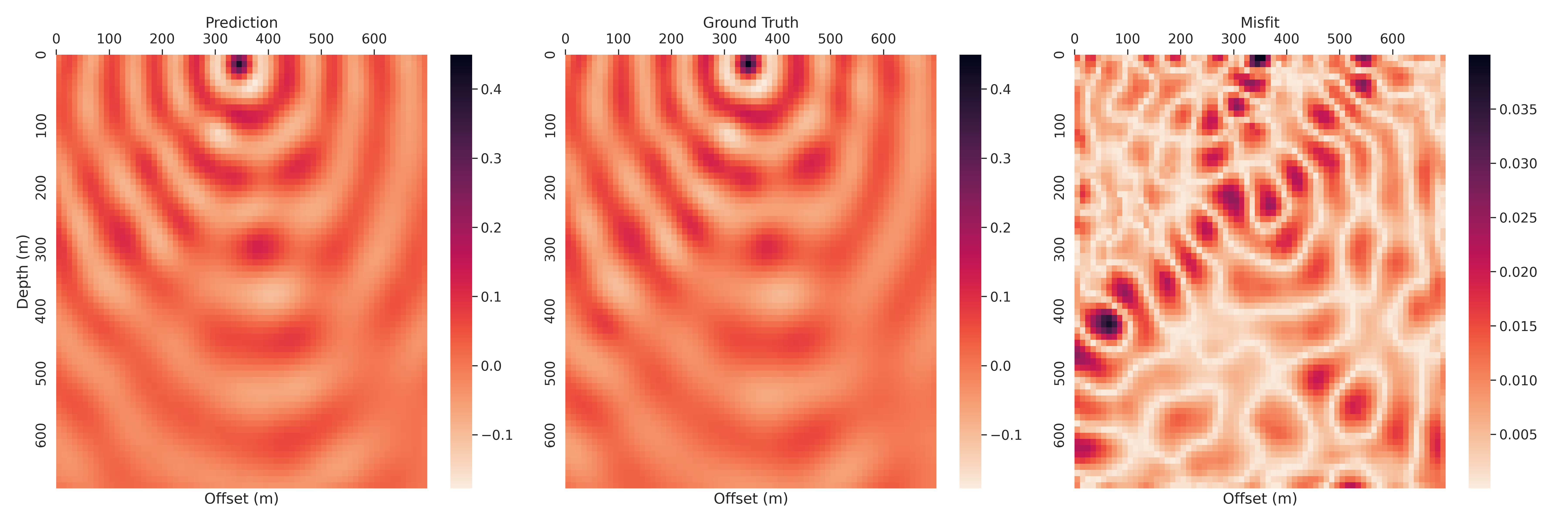}}
  \caption{Real part of 25Hz wavefields - \emph{CurveFault}: (a) \emph{CurveFault-A}; (b) \emph{CurveFault-B}.}
  \label{fig:wavefields_CurveFault}
\end{figure}

\begin{figure}[!ht]
  \centering
  \subfloat[]{\label{fig:wavefields_Style_A}
  \includegraphics[width= \textwidth, height= 0.22\textheight]{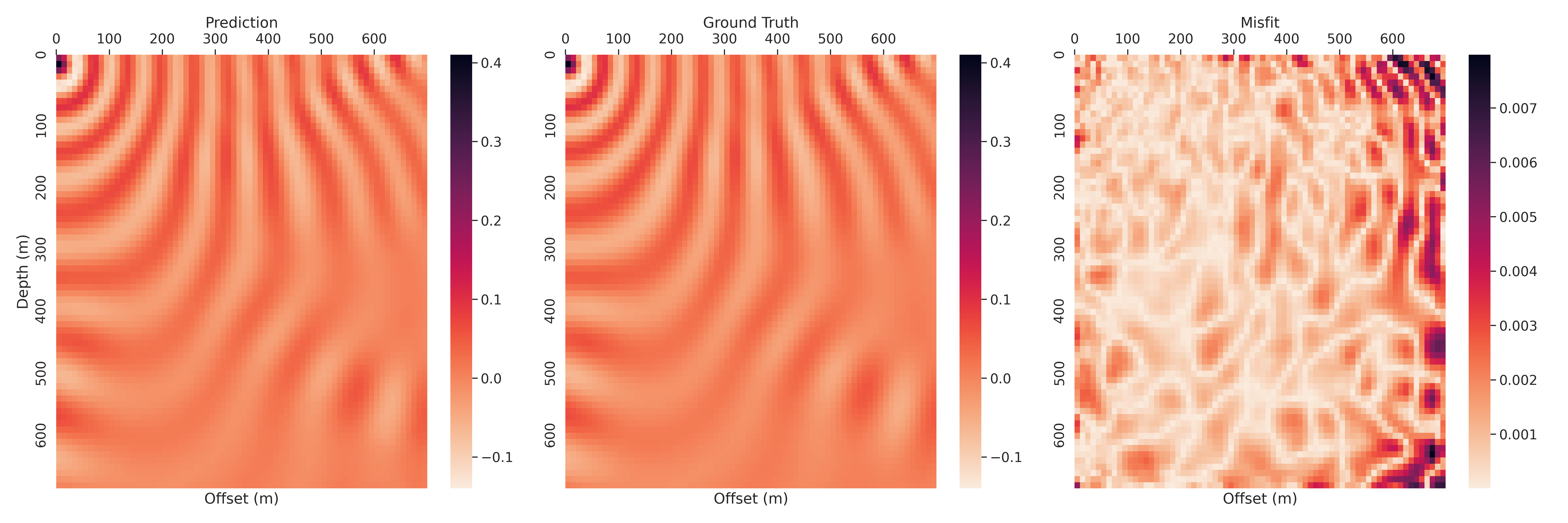}}\\
  \subfloat[]{\label{fig:wavefields_Style_B}
  \includegraphics[width= \textwidth, height= 0.22\textheight]{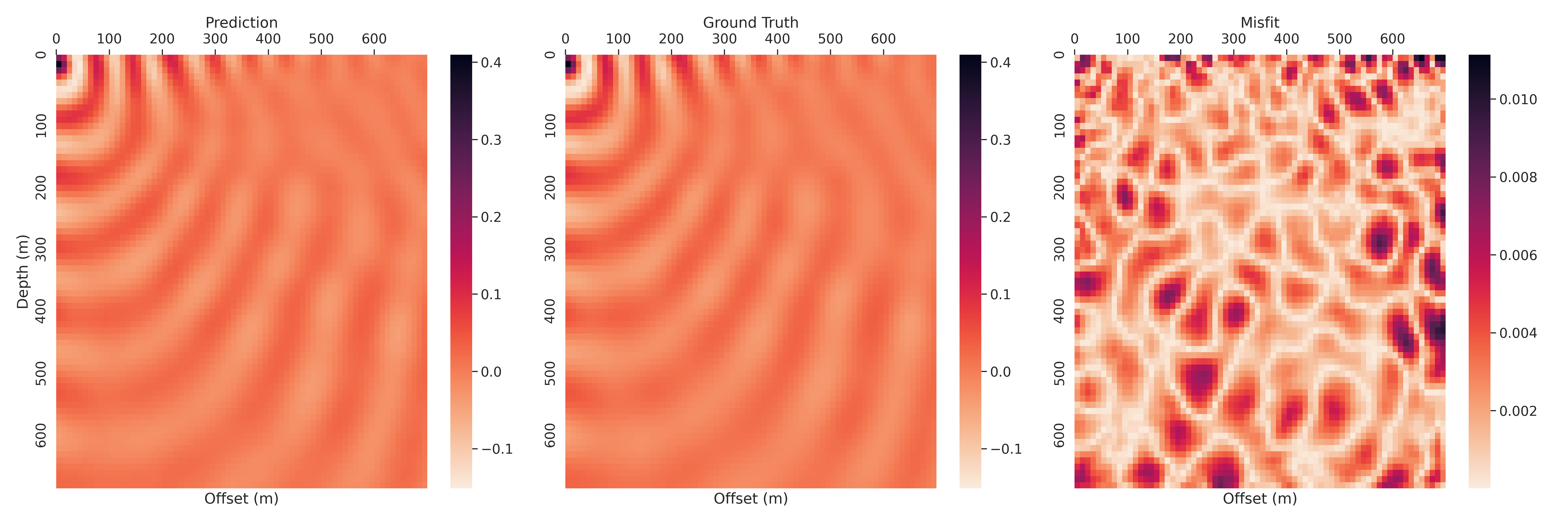}}
  \caption{Real part of 25Hz wavefields - \emph{Style}: (a) \emph{Style-A}; (b) \emph{Style-B}.}
  \label{fig:wavefields_Style}
\end{figure}

\begin{figure}[!ht]
  \centering
  \subfloat[]{\label{fig:wavefields_FlatVel_A}
  \includegraphics[width= \textwidth, height= 0.22\textheight]{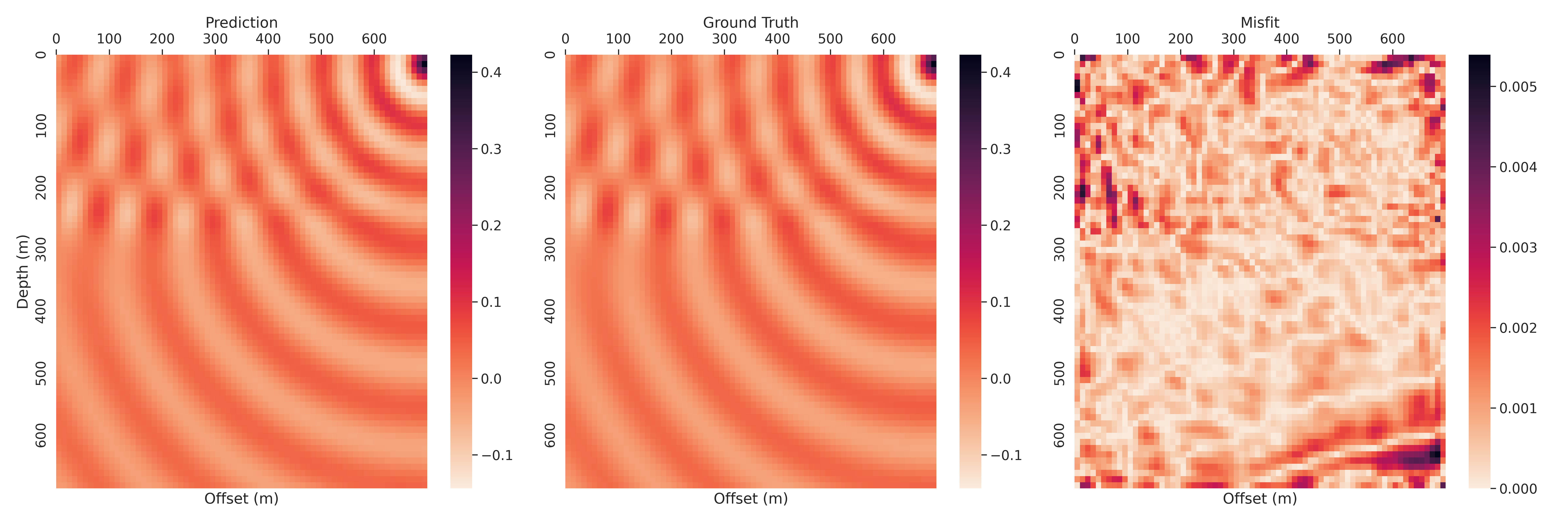}}\\
  \subfloat[]{\label{fig:wavefields_FlatVel_B}
  \includegraphics[width= \textwidth, height= 0.22\textheight]{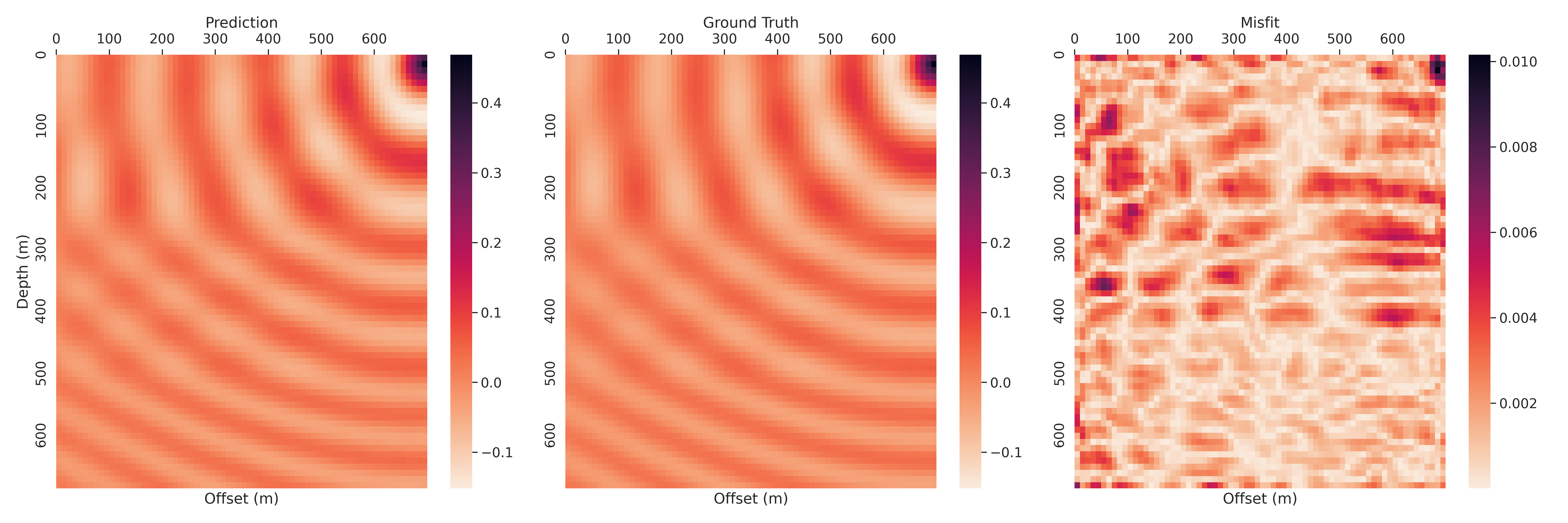}}
  \caption{Real part of 25Hz wavefields - \emph{FlatVel}: (a) \emph{FlatVel-A}; (b) \emph{FlatVel-B}.}
  \label{fig:wavefields_FlatVel}
\end{figure}

Figures \ref{fig:wavefields_CurveFault},~\ref{fig:wavefields_Style}, and~\ref{fig:wavefields_FlatVel} visualize the performance of PFNO on \emph{CurveFault} datasets, \emph{Style} datasets, and \emph{FlatVel} datasets, respectively. For demonstration purposes, only the real part of the $25$Hz wavefields is shown since the imaginary part is similar. As depicted in \Cref{fig:wavefields_CurveFault}, the geological faults induce a number of scattering artifacts that form the cobwebs in some local areas. Given smooth features in the velocity models,  \Cref{fig:wavefields_Style_A} exhibits smooth curves in the wavefields as expected. In comparison, ruffled layers manifest in \Cref{fig:wavefields_Style_B}. The irregularities in \Cref{fig:wavefields_FlatVel} reveal the reflections of energy between layer interfaces, although these interfaces are flat. Since \emph{Style} and \emph{FlatVel} have relatively simpler wavefields, PFNO performs better on these two families than on \emph{CurveFault}. Notably, the resolution of the wavefield at a certain point is inverse-proportional to its wavelength and the velocity value at that point if we fix the frequency. Thus, the blurry parts are caused by the small wavenumber at the current frequency. In addition, we also visualize the time domain wavefields reconstructed by inverse Fourier transform in \cref{sec:time_domain_wavefield}.

\begin{figure}[!ht]
    \centering
    \subfloat[]{\label{fig:model_comparison_FNO}
    \includegraphics[width= 0.42\textwidth, height= 0.22\textheight]{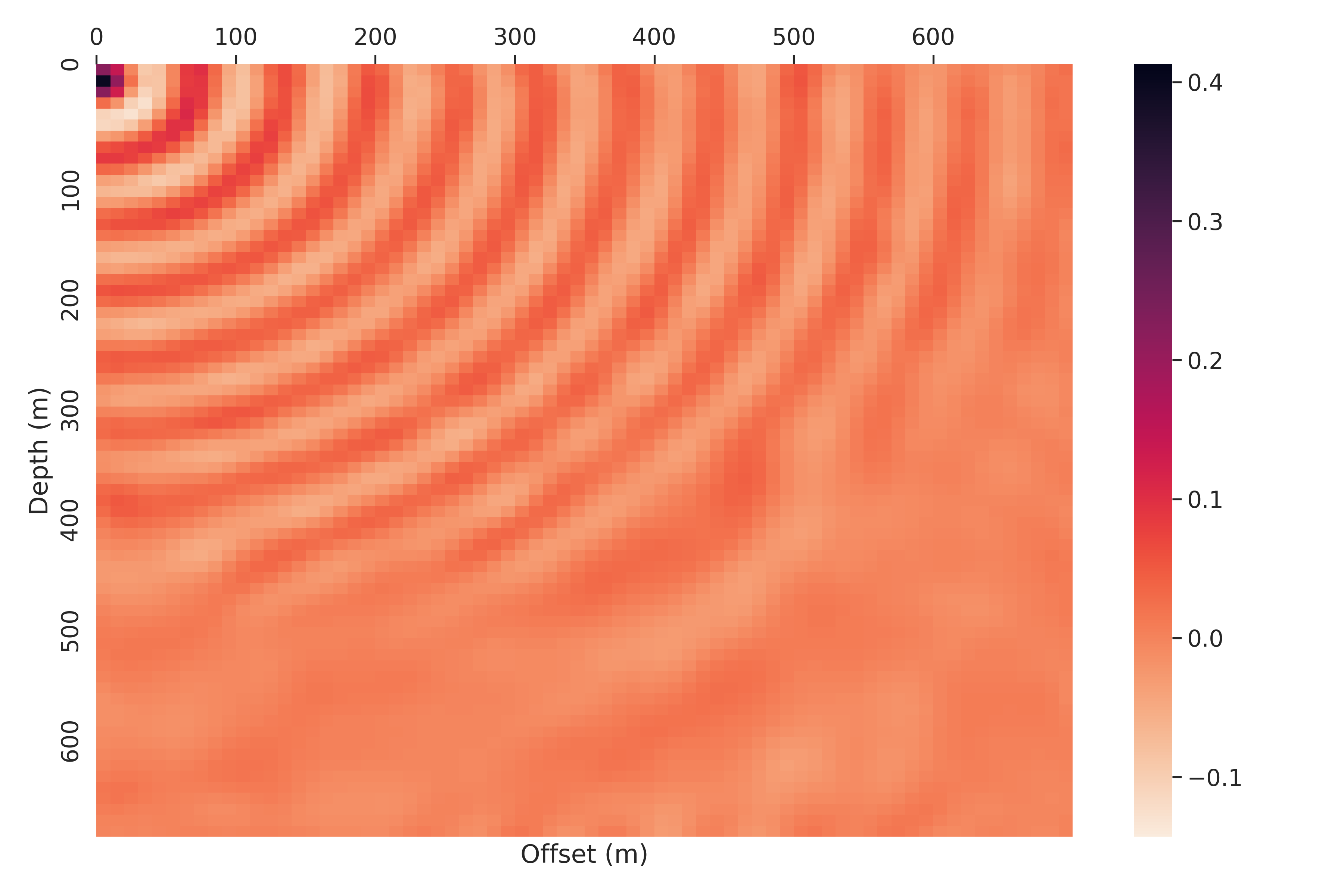}}
    \subfloat[]{\label{fig:model_comparison_ForwardNet}
    \includegraphics[width= 0.42\textwidth, height= 0.22\textheight]{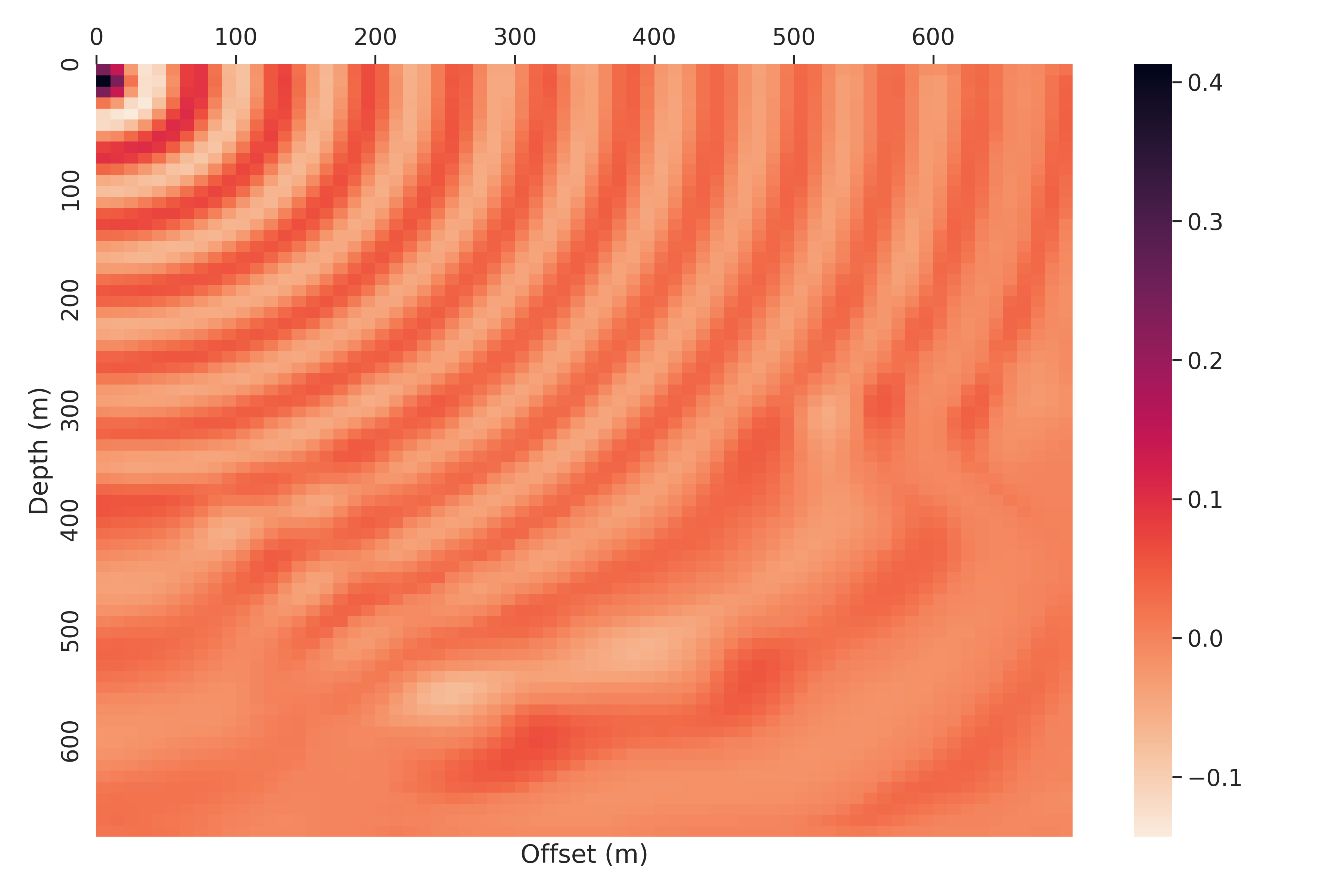}} \\
    \subfloat[]{\label{fig:model_comparison_PFNO}
    \includegraphics[width= 0.42\textwidth, height= 0.22\textheight]{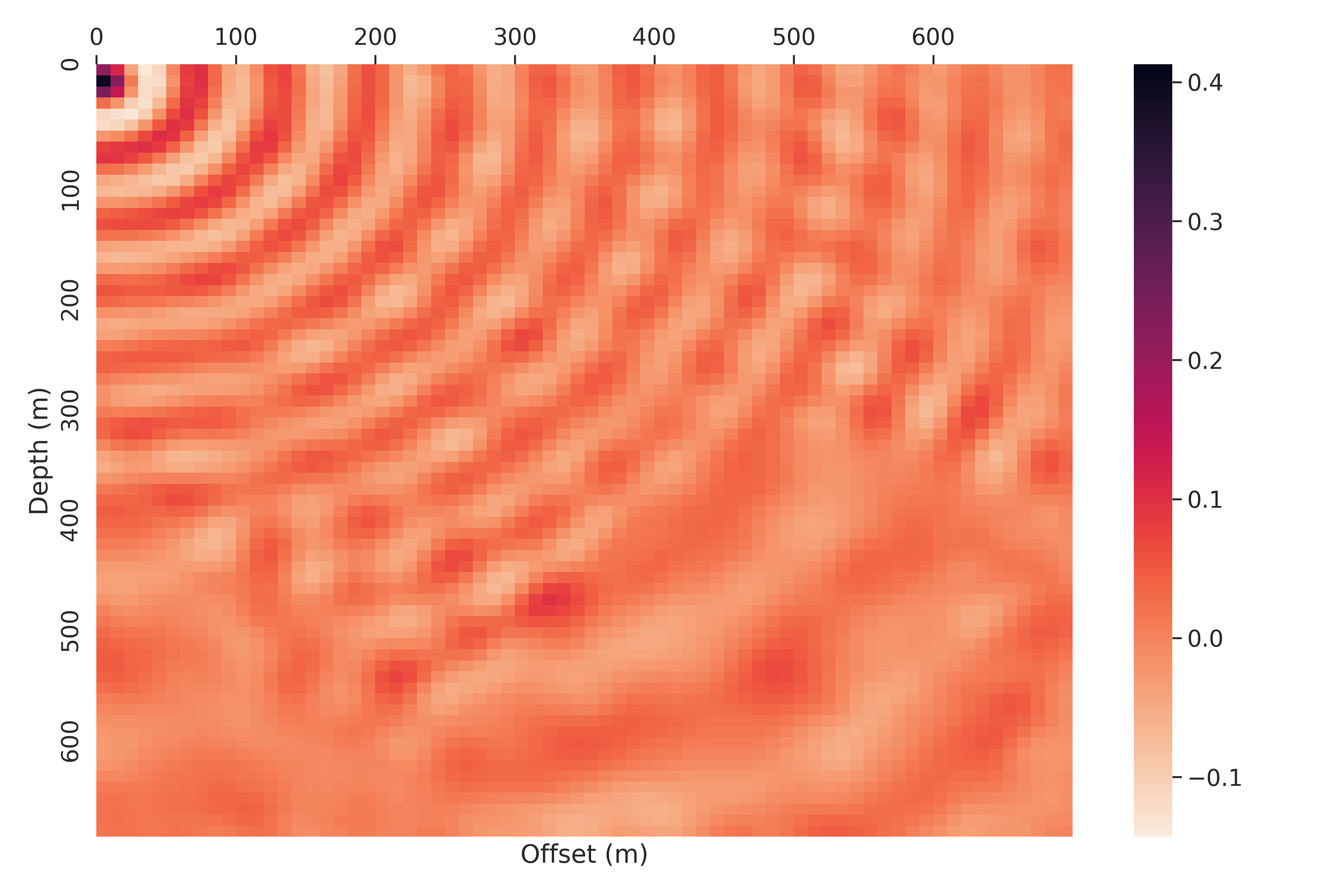}}
    \subfloat[]{\label{fig:model_comparison_label}
    \includegraphics[width= 0.42\textwidth, height= 0.22\textheight]{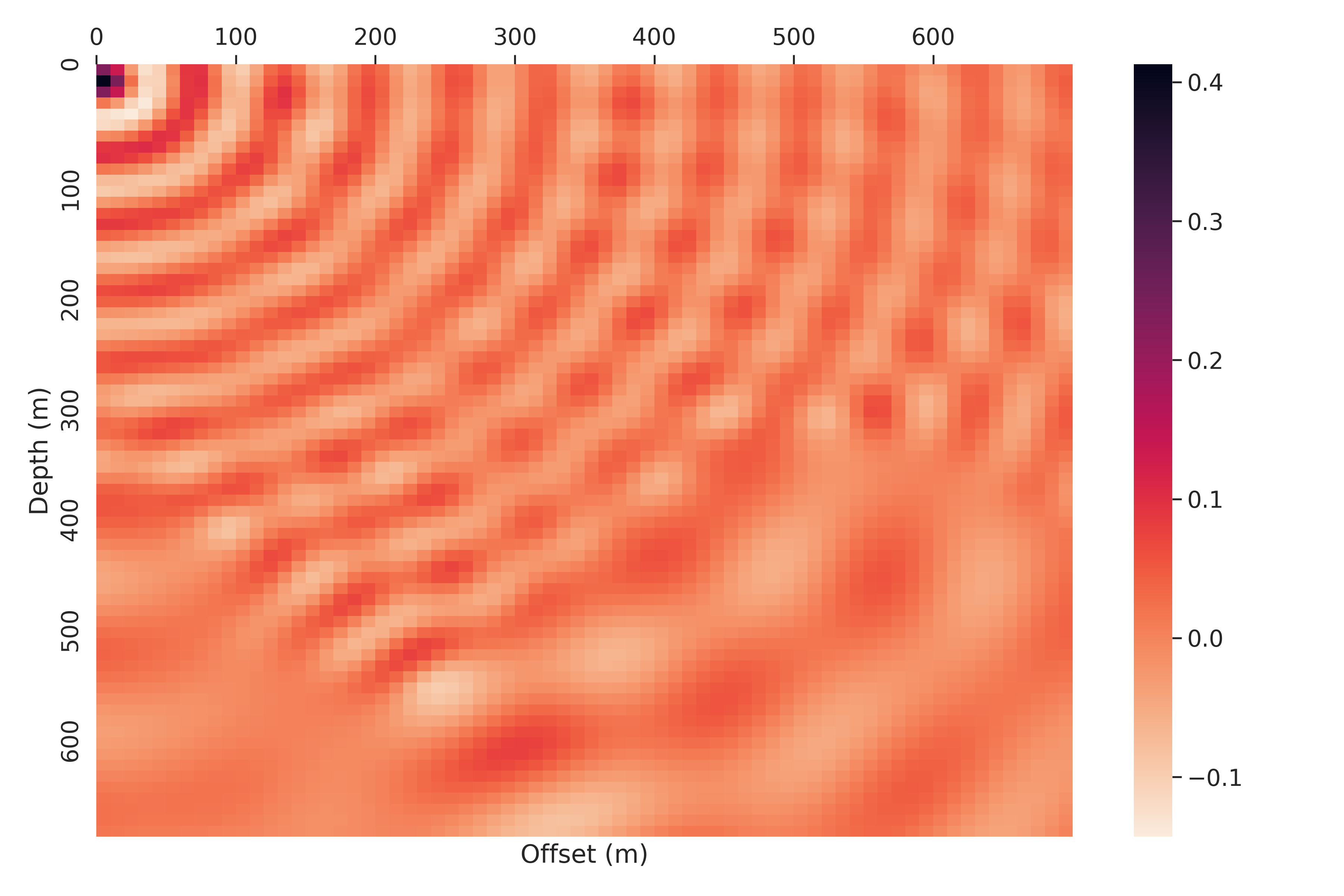}} \\
    \caption{Model performance comparison: (a) FNO; (b) ForwardNet; (c) PFNO; (d) Ground Truth.}
    \label{fig:model_comparison}
\end{figure}

As a comparison, we also implement a single FNO of width $96$ that takes source location, frequency, and velocity into its inputs. In addition, an encoder-decoder architecture is introduced as an image-to-image method. This architecture is inspired by InversionNet~\cite{wu2019inversionnet}. With the purpose of solving the forward problem, an encoder of $11$ convolutional layers maps the velocity model to a latent space, and then a decoder also of $11$ convolutional layers maps the latent representation to the wavefield. For convenience, it is named ForwardNet. The specific structure and hyperparameters of ForwardNet can be found in~\Cref{sec:forwardnet_architecture}.  Similarly, we examine the performance of FNO and ForwardNet on the $6$ datasets and compare them with PFNO. Here, the mean-squared error (MSE) is employed to measure how PFNO performs on the testing set. As \Cref{tab:comparison} shows, PFNO consistently outperforms the other two models on all datasets. To provide an intuitive comparison, \Cref{fig:model_comparison} visualizes the predicted wavefields and the ground truth using a $25$Hz example from \emph{CurveFault-A}. Highly agitated patterns of the layer edges are recognized by PFNO, whereas FNO and ForwardNet only capture the smooth part. Moreover, the wavefields predicted by FNO turn blurry as the wave spreads rightward, but PFNO still gives sharp scatterings, as the ground truth suggests. 
\begin{table}[!ht]
  \centering
  \caption{Performance of the models, measured by MSE}
    \resizebox{0.5\textwidth}{!}{\begin{tabular}{cccc}
    \hline\hline
        & FNO   & ForwardNet & PFNO \\
    \midrule
    \emph{CurveFault-A} & 4.231e-05 & 4.441e-05 & \textbf{3.357e-05} \\
    \emph{CurveFault-B} & 2.213e-04 & 2.588e-04 & \textbf{1.959e-04} \\
    \emph{Style-A} & 2.262e-05 & 3.304e-05 & \textbf{1.713e-05} \\
    \emph{Style-B} & 1.702e-05 & 2.575e-05 & \textbf{8.478e-06} \\
    \emph{FlatVel-A} & 8.280e-06 & 1.372e-05 & \textbf{3.209e-06} \\
    \emph{FlatVel-B} & 3.316e-05 & 3.591e-05 & \textbf{2.302e-05} \\
    \hline\hline
    \end{tabular}}%
  \label{tab:comparison}%
\end{table}%

\subsection{Generalization}
The model's ability to generalize to other datasets is of importance. If a trained model only approximates certain patterns within a dataset instead of the intrinsic properties of the Helmholtz operator, it needs to be retrained once the features of the velocity model change, which offsets the benefit of fast inference. To examine the generalization of PFNO, we hereby perform a cross-dataset generalization test. More specifically, we test the performance of each trained PFNO in \cref{subsec:wavefield_prediction} on the other $5$ datasets. In Table~\ref{tab:generalization}, the diagonal marks the intra-dataset test results. As expected, the smallest error for each testing set appears on the diagonal. Since \emph{CurveFault-A} and \emph{CurveFault-B} share similar features in their velocity models, PFNO trained on either one generalizes to the other. Furthermore, PFNOs trained on them can adapt to all other datasets except \emph{FlatVel-B} because the \emph{CurveFault} family admits features seen in the other two families, such as curves and layered structures. Interestingly, the \emph{Style-A}-trained PFNO has a mild generalization to \emph{CurveFault-A}, but it fails to generalize to \emph{CurveFault-B}. Also, it can generalize to \emph{Style-B} despite sharper features in it. In contrast, its \emph{Style-B}-trained counterpart only adapts to \emph{Style-A}. This is presumably caused by the unique sharp features in \emph{Style-B}. Among the $3$ families, \emph{FlatVel} seems to be biased to the other two distributions because it contains only flat layers and no lateral variation. Thus, PFNOs trained on them are unable to generalize to the others. Although the \emph{FlatVel-B}-trained PFNO makes accurate predictions on \emph{FlatVel-A}, the opposite direction does not apply. The reason roots in the randomly distributed velocities across layers in \emph{FlatVel-B}. 
\begin{table*}[!ht]
  \centering
  \caption{Cross-dataset generalization: the rows are the training sets and the columns are the testing sets; errors are measured by MSE.}
    \resizebox{0.8\textwidth}{!}{\begin{tabular}{ccccccc}
    \hline\hline
      & \emph{CurveFault-A} & \emph{CurveFault-B} & \emph{Style-A} & \emph{Style-B} & \emph{FlatVel-A} & \emph{FlatVel-B} \\
    \midrule
    \emph{CurveFault-A} & \textbf{3.357e-05} & 8.873e-04 & 2.067e-04 & 1.269e-04 & 8.963e-05 & 3.451e-03 \\
    \emph{CurveFault-B} & 1.070e-04 & \textbf{1.959e-04} & 6.748e-05 & 4.835e-05 & 8.869e-05 & 2.034e-03 \\
    \emph{Style-A} & 8.221e-04 & 1.853e-03 & \textbf{1.713e-05} & 2.834e-05 & 2.851e-03 & 6.607e-03 \\
    \emph{Style-B} & 1.232e-02 & 2.768e-02 & 2.294e-04 & \textbf{8.478e-06} & 2.268e-02 & 3.550e-01 \\
    \emph{FlatVel-A} & 3.040e-03 & 4.793e-03 & 2.377e-03 & 1.070e-03 & \textbf{3.209e-06} & 1.185e-03 \\
    \emph{FlatVel-B} & 3.015e-03 & 4.704e-03 & 2.382e-03 & 1.054e-03 & 1.185e-05 & \textbf{2.302e-05} \\
    \hline\hline
    \end{tabular}}
  \label{tab:generalization}%
\end{table*}%

Another pivotal generalization test is to examine how FNO/PFNO adapts to a smooth velocity model. The motivation lies in the fact that FWI methods usually start with a smooth velocity model. By iterative updates, these methods gradually find sharper features. Therefore, FNO-based solvers must be able to accurately predict the wavefields in the early stage if they are to be applied as the forward modeling module in an FWI method. 

\begin{figure}[!ht]
    \centering
    \subfloat[]{
    \includegraphics[width= 0.32\textwidth, height= 0.16\textheight]{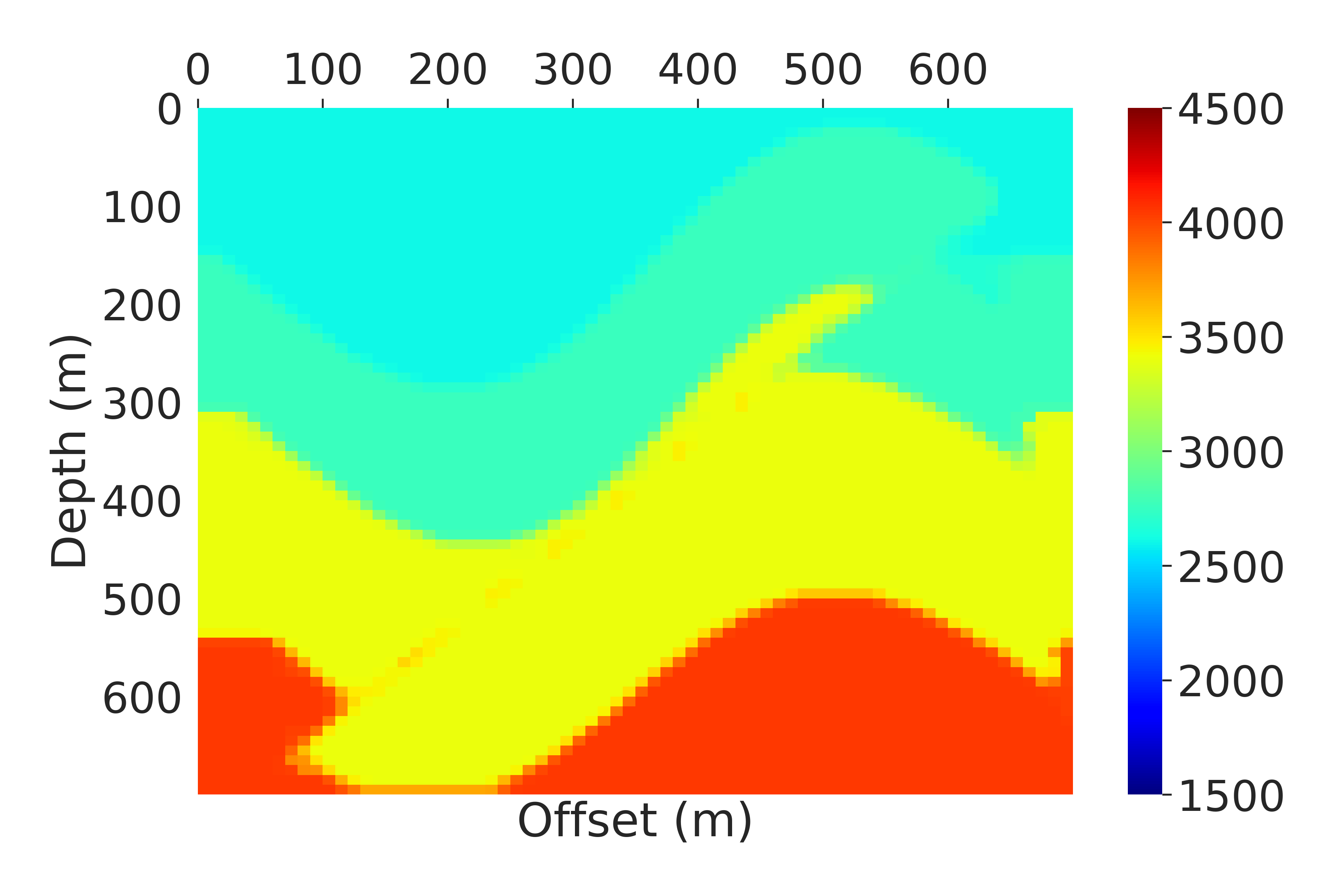}}
    \subfloat[]{\label{fig:gen_smooth_descent_10}
    \includegraphics[width= 0.32\textwidth, height= 0.16\textheight]{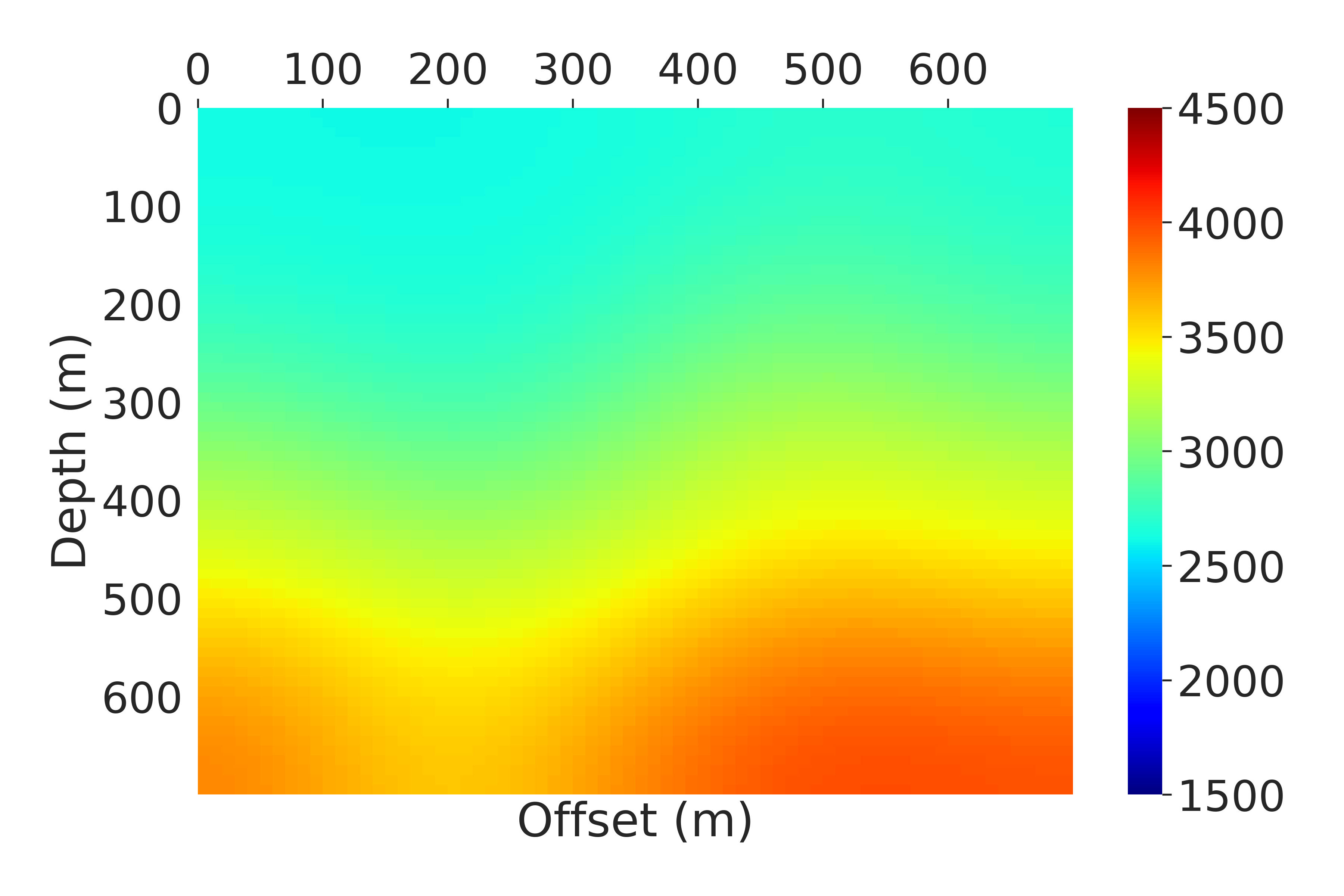}} 
    \subfloat[]{\label{fig:gen_smooth_descent_mse}
    \includegraphics[width= 0.32\textwidth, height= 0.16\textheight]{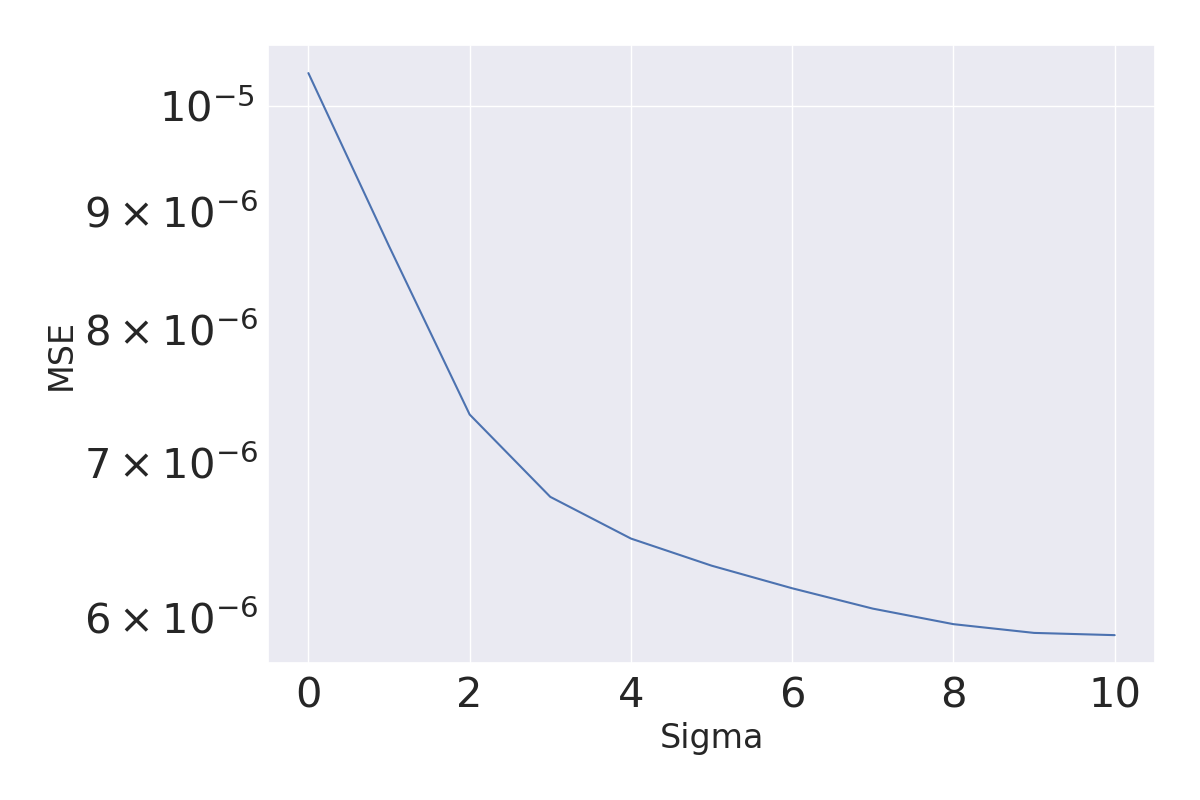}}
    \caption{Generalization on smooth velocity models: (a) (original) velocity model with $\text{sigma} = 0$; (b) velocity model with $\text{sigma} = 10$; (c) MSE against sigma, where sigma is the standard deviation of the Gaussian kernel.}
    \label{fig:gen_smooth_descent}
\end{figure}

With the PFNO trained on \emph{CurveFault-A}, we test its performance on velocity models that are smoothed by Gaussian filters. Particularly, the degree of smoothness is controlled by the standard deviation of the Gaussian kernel. As we smooth the velocity model, it becomes less similar to the training velocity models, and the neural network has never seen such smoothed velocity models in the training stage, which is usually described as out-of-distribution. Meanwhile, the smoothness leads to simpler wavefields as the high wavenumber part of the velocity models has been smoothed out. The generalization is subject to the combination of these two factors. \Cref{fig:gen_smooth_descent_mse} shows that the error drops to a lower level. As can be seen in \Cref{fig:gen_smooth_descent_10}, the velocity model is so smooth that the highest-frequency component in the wavefield is no longer sensitive to the smoothed faults. In this case, the impact of simpler wavefields outweighs the impact of out-of-distribution.

However, it is possible to see an increase in the total error due to a phase shift at a far offset in the shallow layers. The main reason for such a phase shift is the low average velocity along the wavepath. The relationship of wavelength $\lambda$, frequency $f$, and velocity $v$ is $\lambda= \frac{v}{f}$, which results in a high wavenumber (small $\lambda$) wavefield in the low-velocity zone when $f$ is fixed. Once applied a Gaussian filter, the out-of-distribution velocity models cause wavefield prediction misfits, and such misfits will accumulate along the wavepaths. A neighborhood with a large wavenumber is more sensitive to this inaccuracy, thus showing a more significant phase shift.  
Hence, when FNO/PFNO predicts the wavefield of a smoothed and out-of-distribution velocity model, it is most likely to see a wavefield phase shift in the far-offset low-velocity area, given a fixed frequency. 

\begin{figure}[!ht]
    \centering
    \subfloat[]{\label{fig:gen_smooth_v8_s0_o}
    \includegraphics[width= 0.192\textwidth, height= 0.128\textheight]{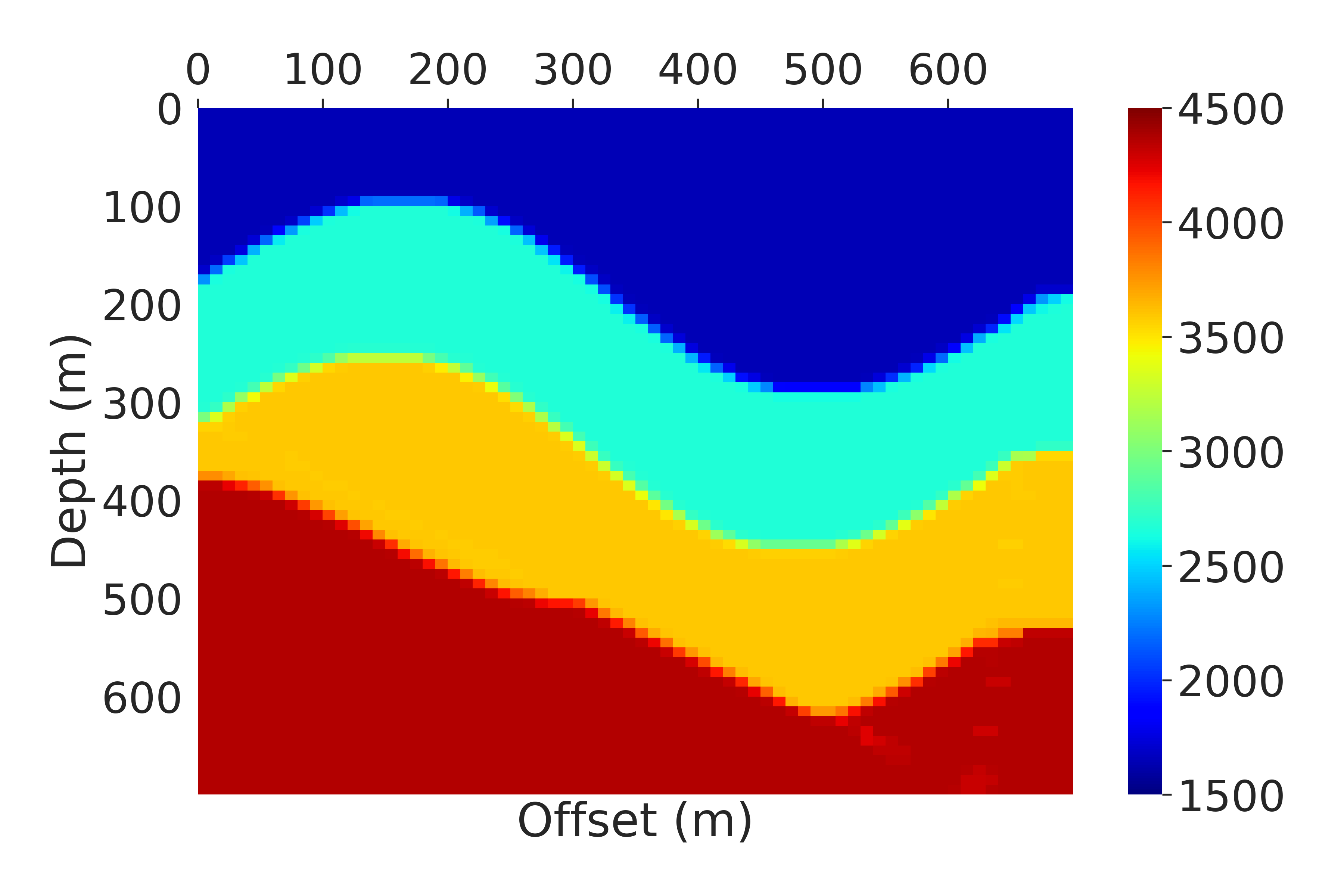}}
    \subfloat[]{\label{fig:gen_smooth_misfit0_o}
    \includegraphics[width= 0.192\textwidth, height= 0.128\textheight]{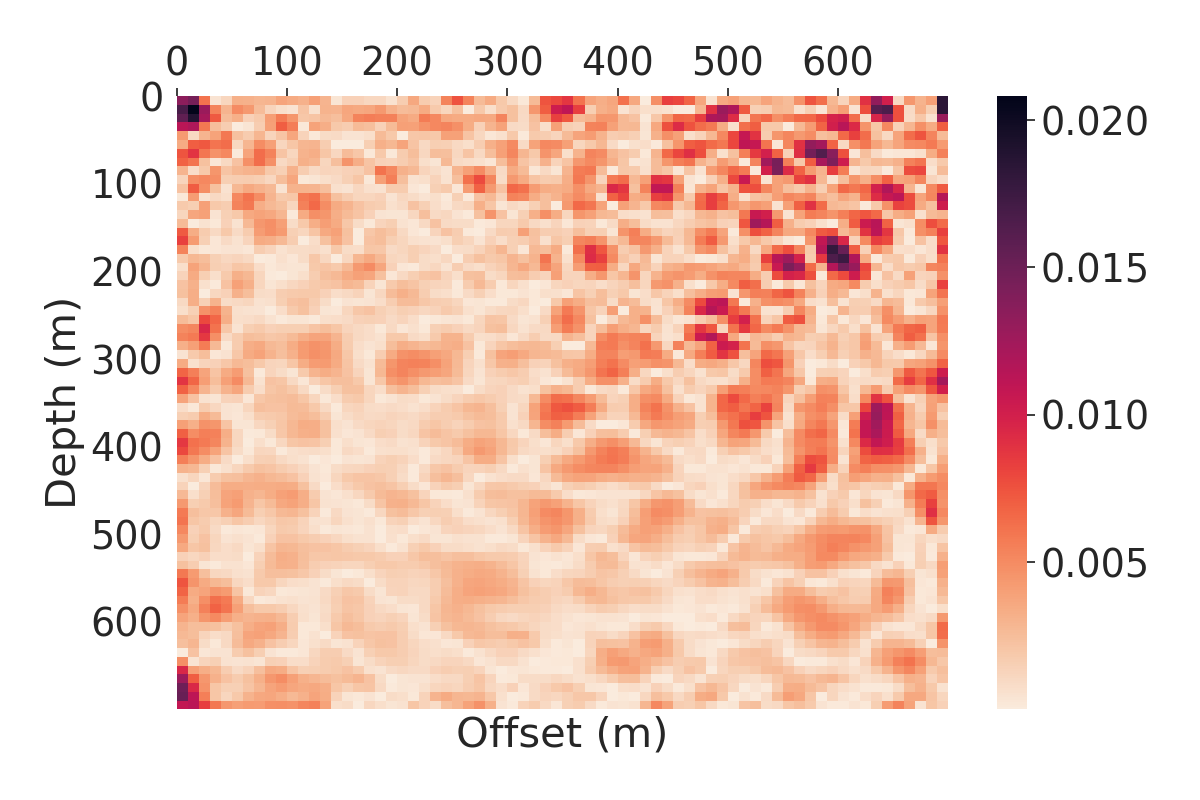}} 
    \subfloat[]{\label{fig:gen_smooth_v8_s10_o}
    \includegraphics[width= 0.192\textwidth, height= 0.128\textheight]{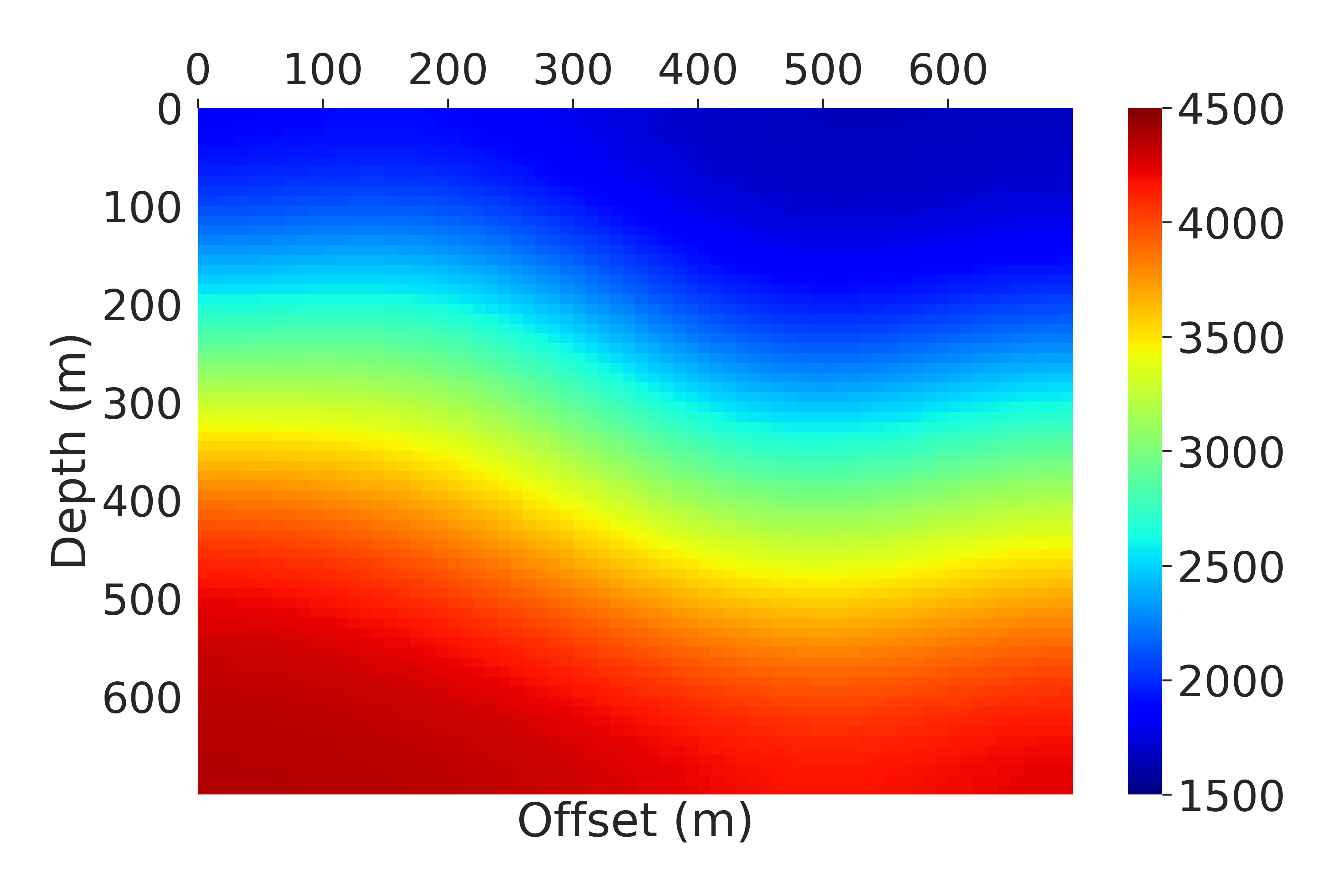}} 
    \subfloat[]{\label{fig:gen_smooth_misfit10_o}
    \includegraphics[width= 0.192\textwidth, height= 0.128\textheight]{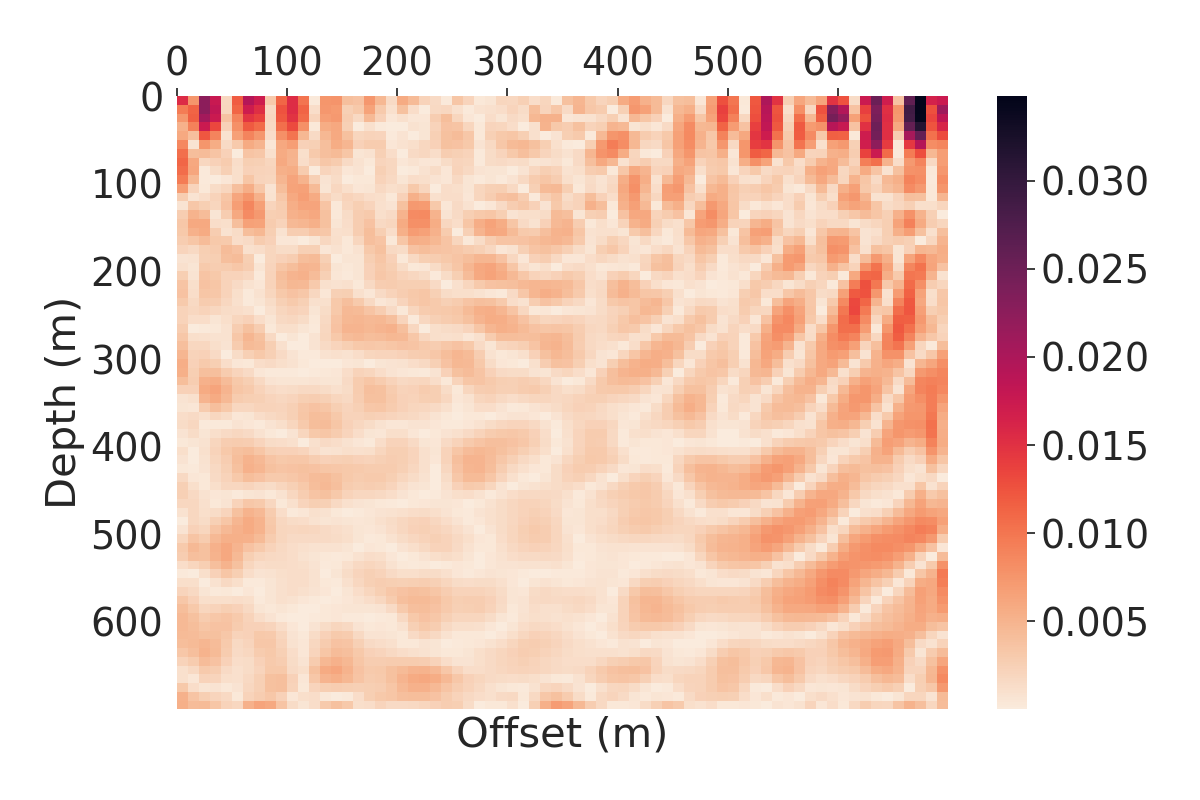}} 
    \subfloat[]{\label{fig:gen_smooth_mse_o}
    \includegraphics[width= 0.192\textwidth, height= 0.128\textheight]{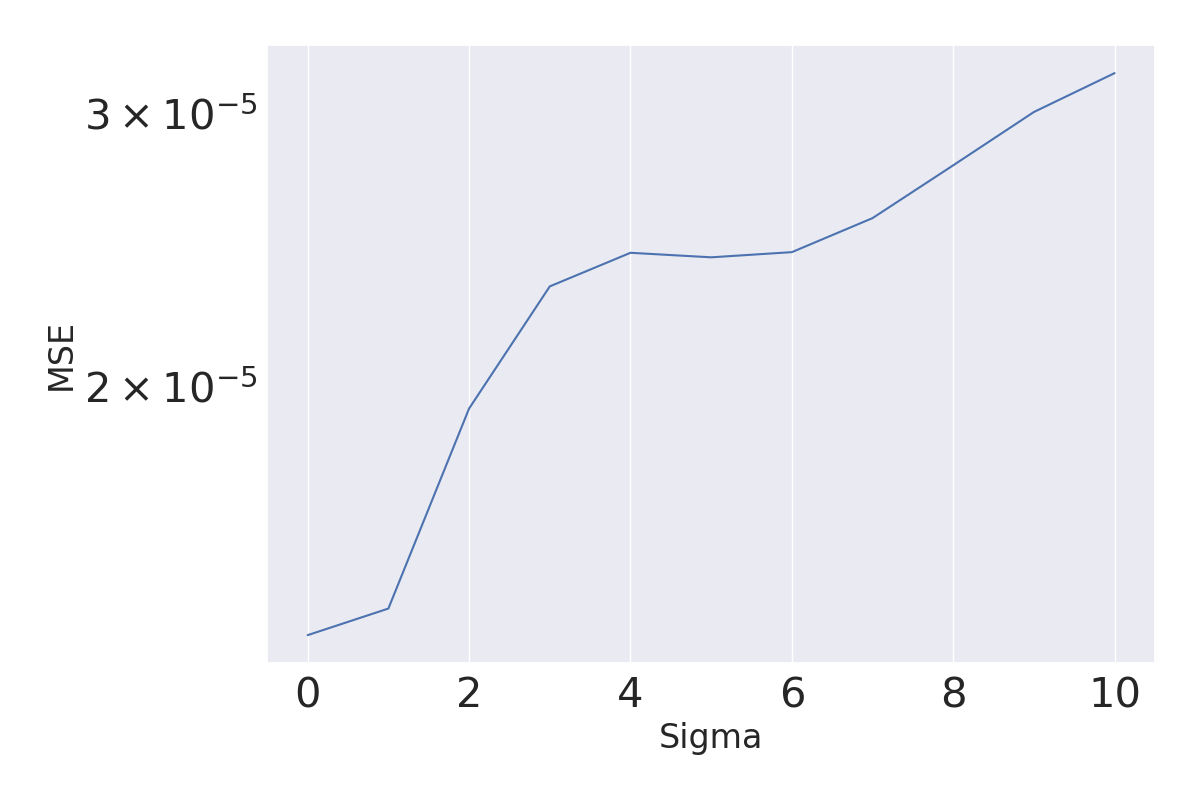}}
    \newline
    \subfloat[]{\label{fig:gen_smooth_v8_s0_m}
    \includegraphics[width= 0.192\textwidth, height= 0.128\textheight]{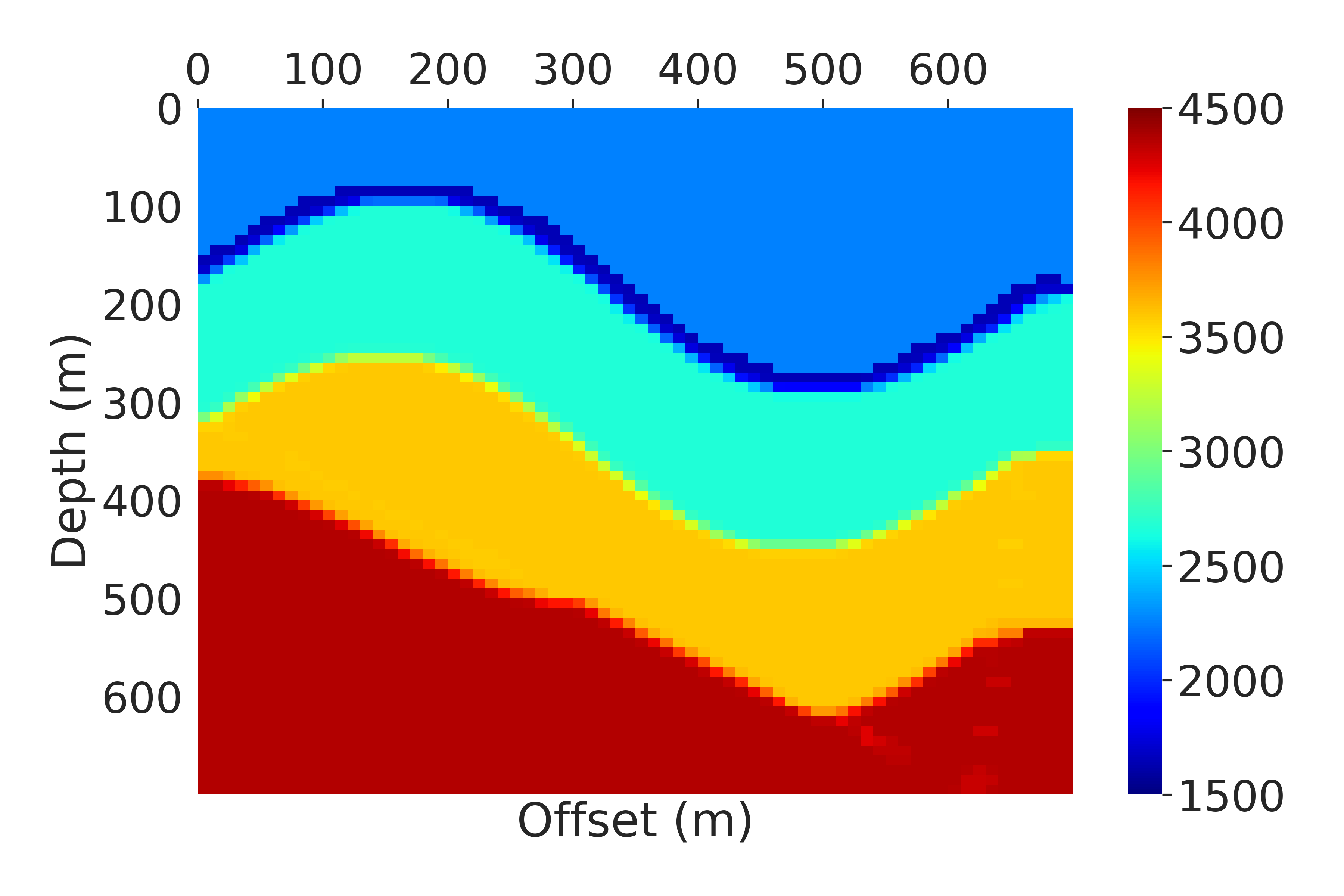}}
    \subfloat[]{\label{fig:gen_smooth_misfit0_m}
    \includegraphics[width= 0.192\textwidth, height= 0.128\textheight]{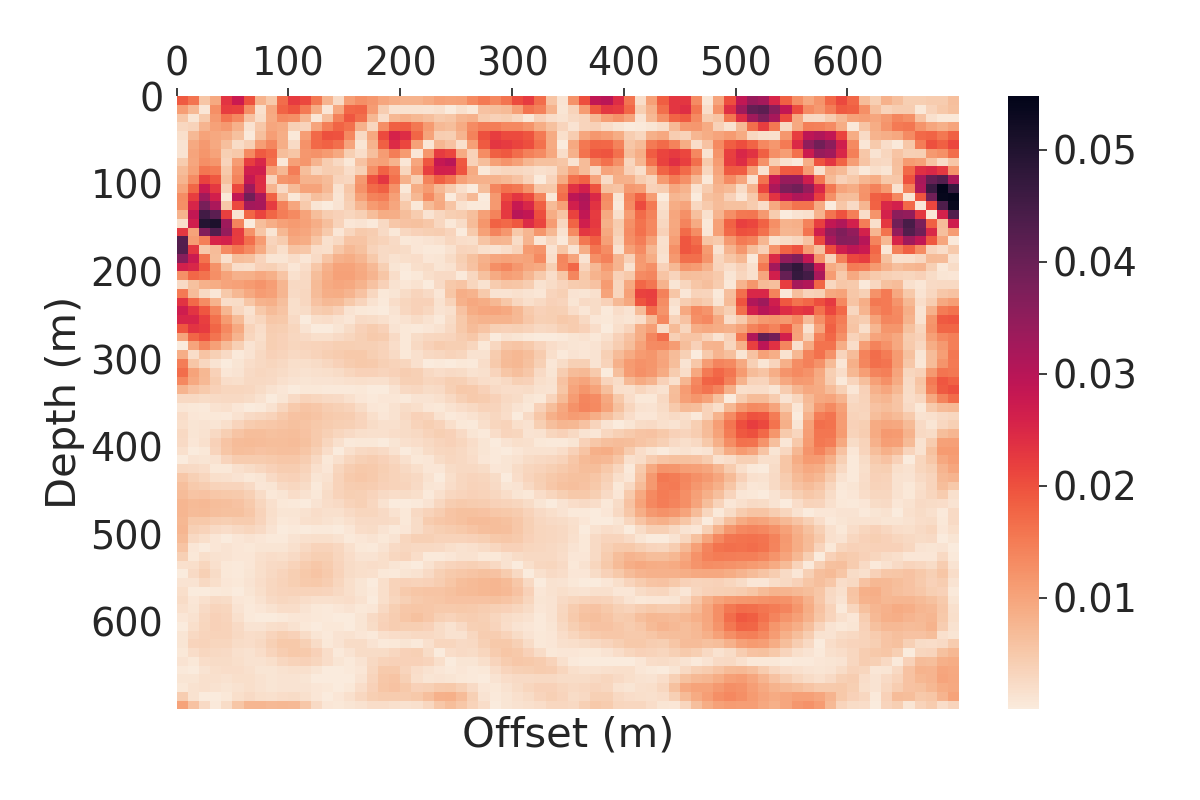}} 
    \subfloat[]{\label{fig:gen_smooth_v8_s10_m}
    \includegraphics[width= 0.192\textwidth, height= 0.128\textheight]{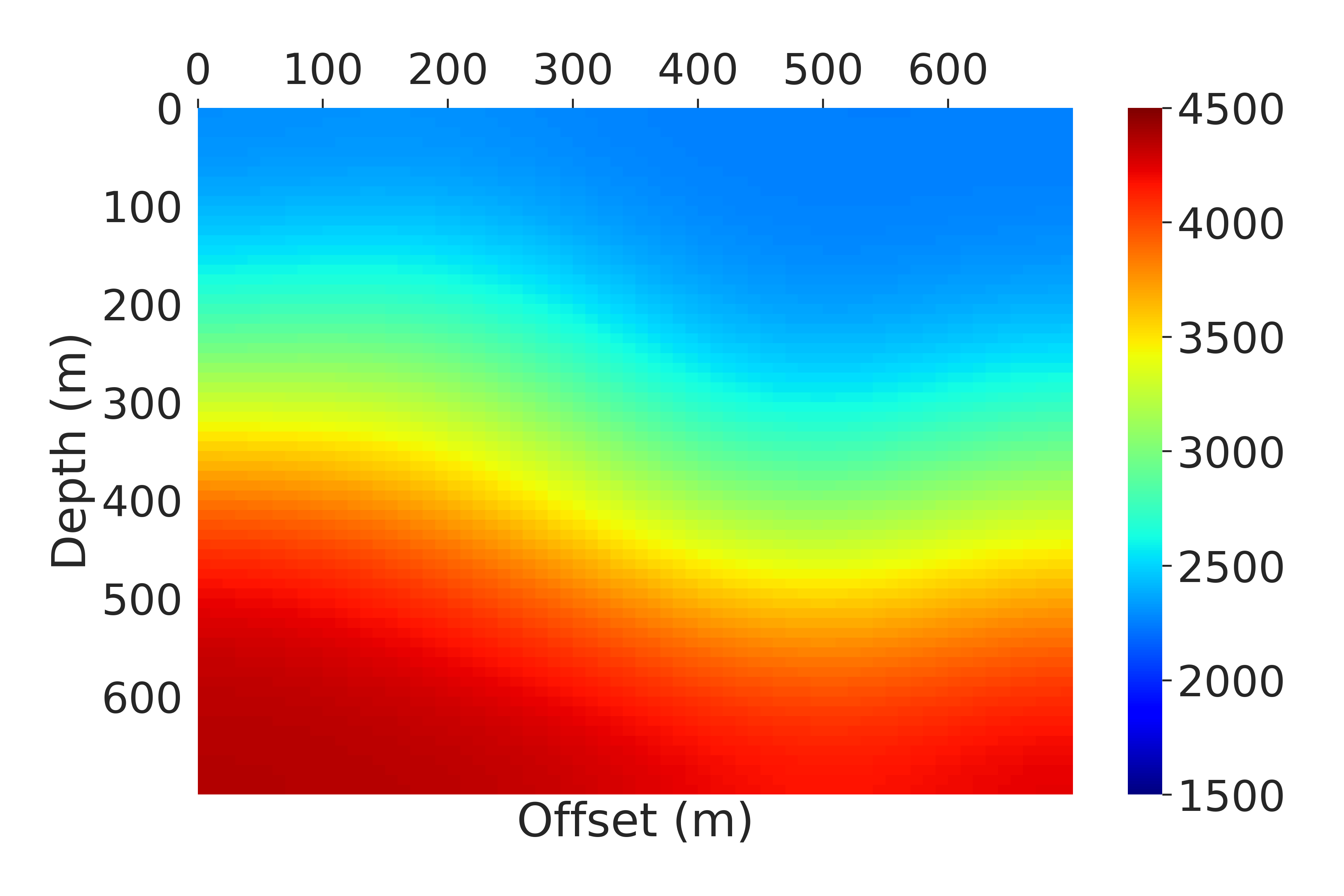}} 
    \subfloat[]{\label{fig:gen_smooth_misfit10_m}
    \includegraphics[width= 0.192\textwidth, height= 0.128\textheight]{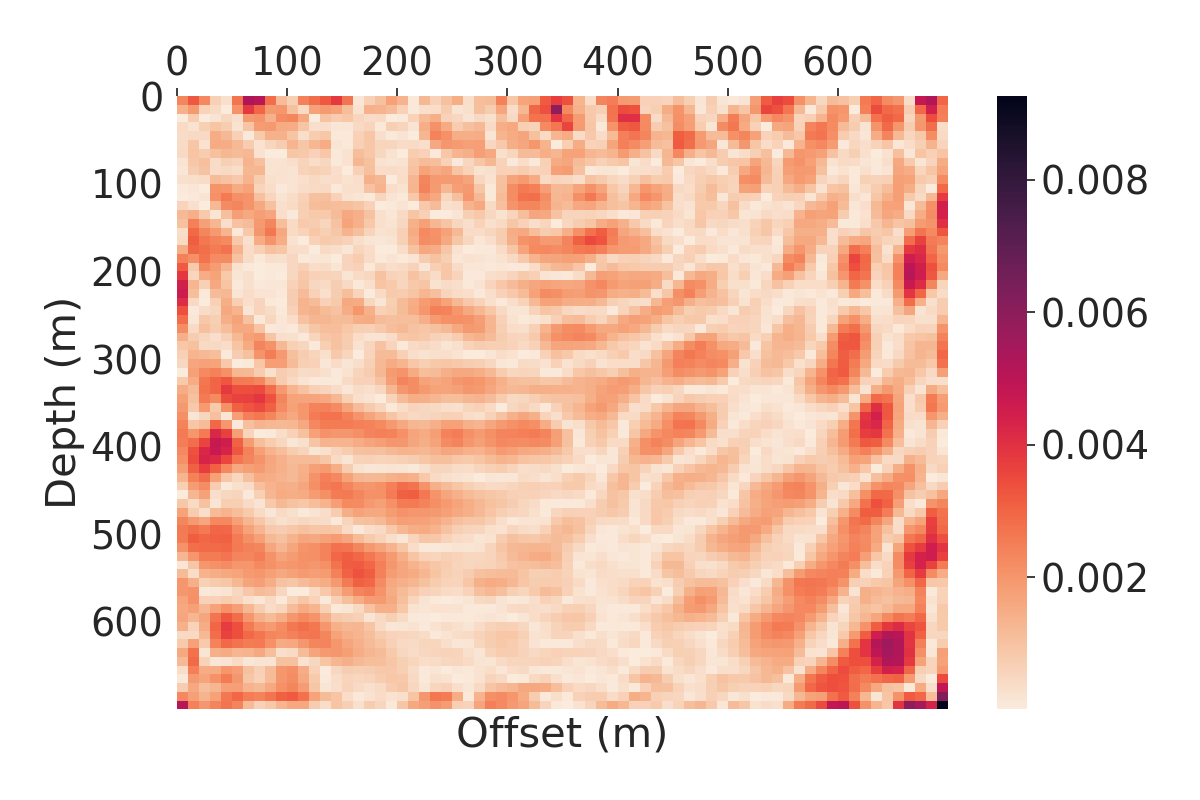}} 
    \subfloat[]{\label{fig:gen_smooth_mse_m}
    \includegraphics[width= 0.192\textwidth, height= 0.128\textheight]{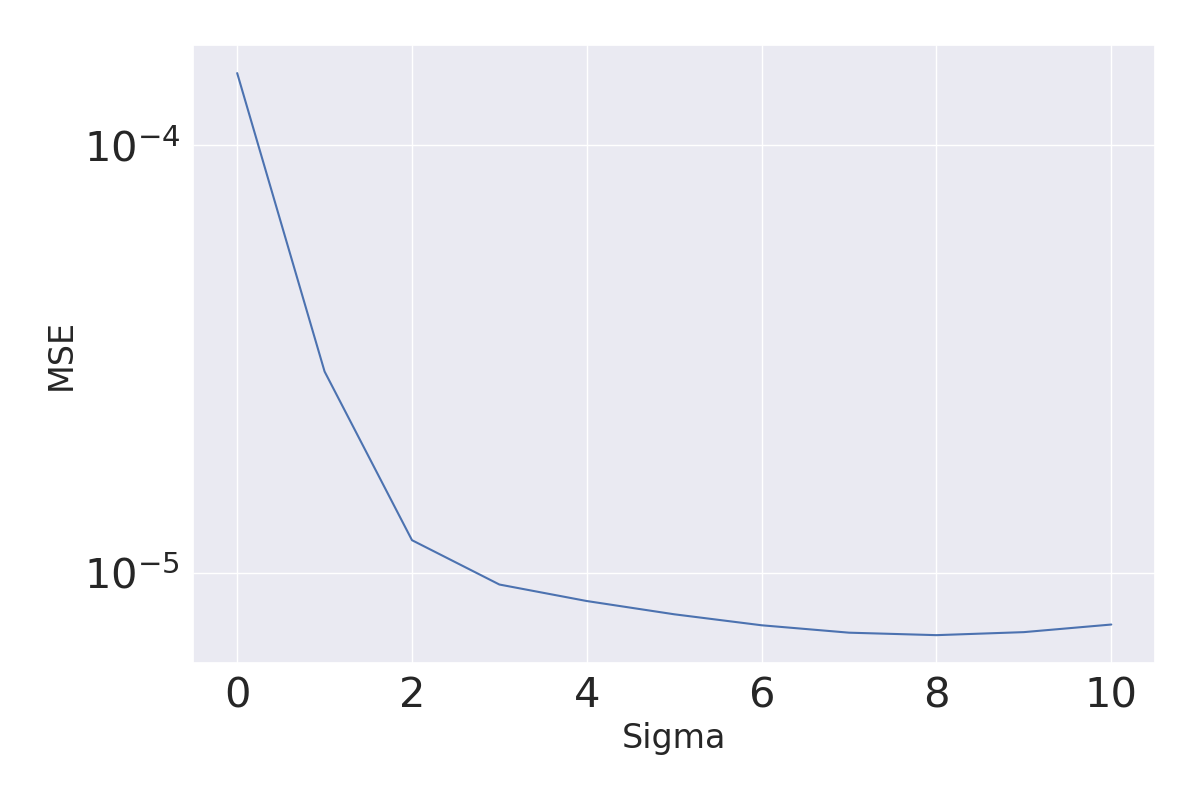}}
    \caption{Generalization test on modified smooth velocity models: (a) original velocity model with first layer $v=1645$m/s and $\text{sigma} = 0$; (b) wavefield prediction misfit of $\textit{(a)}$ at $25$Hz; (c) smoothed velocity model with $\text{sigma} = 10$; (d) wavefield prediction misfit of $\textit{(c)}$ at $25$Hz; (e) MSE against sigma of unmodified velocities, where sigma is the standard deviation of the Gaussian kernel; (f) modified velocity model with first layer $v=2250$m/s and $\text{sigma} = 0$; (g) wavefield prediction misfit of $\textit{(f)}$ at $25$Hz; (h) smoothed modified velocity model with $\text{sigma} = 10$; (i) wavefield prediction misfit of $\textit{(h)}$ at $25$Hz; (j) MSE against sigma of modified velocities.}
    \label{fig:gen_smooth_modify}
\end{figure}

To verify our claim, we perform the following experiment: 1) pick a velocity model, as shown in \Cref{fig:gen_smooth_v8_s0_o}. It has a low-velocity first layer, and its wavefield misfit after the Gaussian smoothing is dominant by the far-offset phase shift in the first layer (top left/right corners in \Cref{fig:gen_smooth_misfit0_o}), which results in the total error increasing, as shown in \Cref{fig:gen_smooth_mse_o}. 2) Modify the original velocity model by increasing the first layer velocity from $1645$m/s to $2250$m/s. 3) Compare the prediction of the modified velocity with the original one. We find the phase shift is significantly suppressed when the velocity is modified, and the total error decreases with respect to the Gaussian filter size. Thus, it is grounded to maintain that the phase shift is more significant in a low-frequency area. To use our proposed FNO/PFNO in an FWI iteration, one can start with lower-frequency components to maintain a relatively small wavenumber, since the velocity cannot be modified as in the test.

Although this work focuses on whether our FNO-based solvers generalize on velocity, it is also worth exploring their generalization on frequency. The detail can be found in \Cref{sec:generalization_freq}.

\subsection{Robustness}
In reality, the measurement of the seismic pressure is prone to the errors of the receivers (sensors). This requires our neural network model to perform consistently under a certain level of uncertainty. To test the robustness of the model, we inject random noise into the labels and training PFNOs with these noisy labels. In particular, zero-mean Gaussian i.i.d. random variables are added to the labels in \emph{CurveFault-A}. To control the uncertainty level, we specify different standard deviations when generating the Gaussian random variables. To ensure that the data has a similar magnitude, we select these $5$ frequencies in the training: $[11, 13, 15, 17, 19]$Hz. 

\begin{figure}[!ht]
    \centering
    \includegraphics[width= 0.8\textwidth]{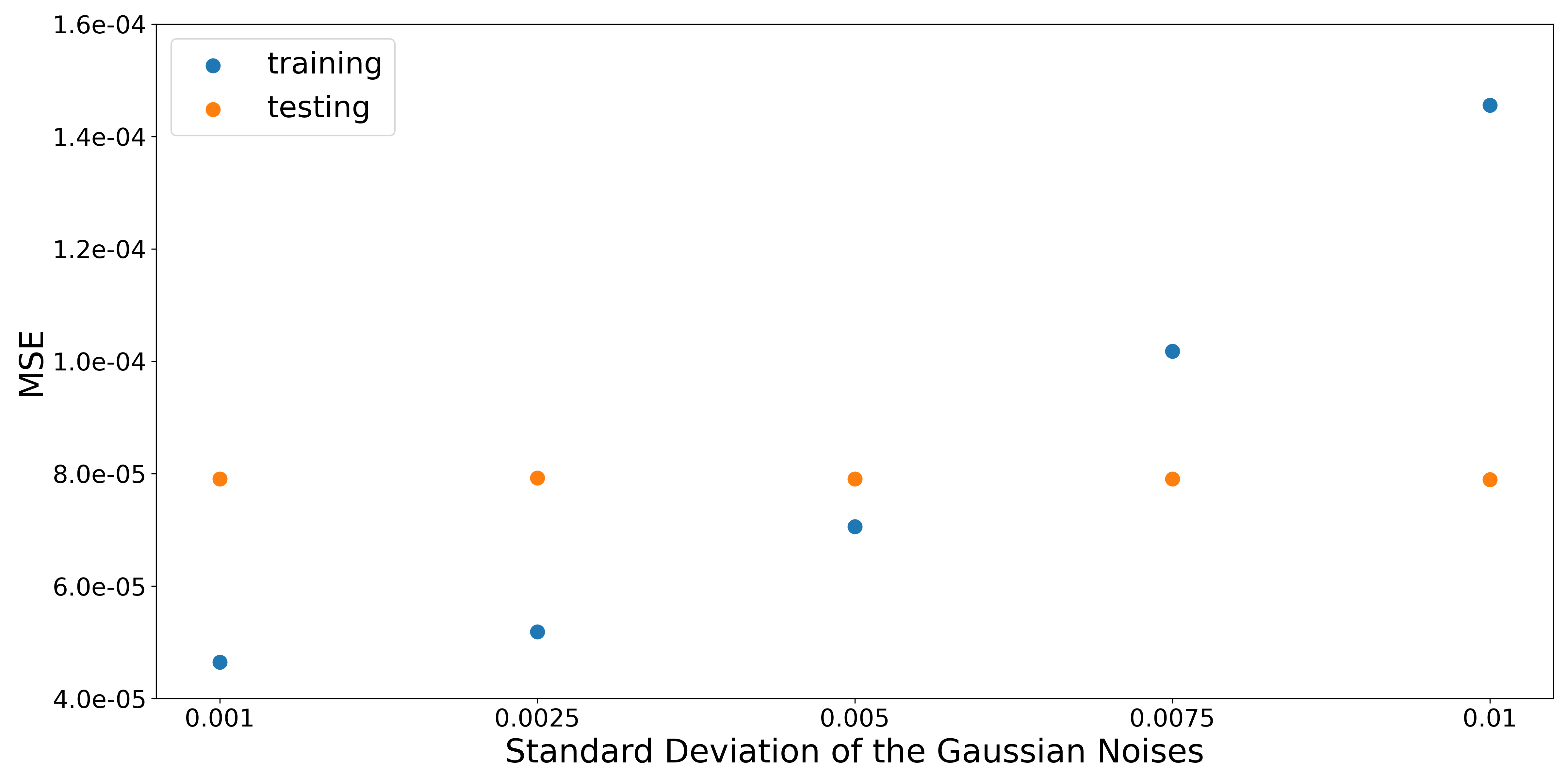}
    \caption{Robustness test: PFNO is trained with different level of Gaussian noises added to the labels}
    \label{fig:robust}
\end{figure}

The result is exhibited in \Cref{fig:robust}. It is not surprising to observe an upward trend in the training MSE. Since the random noise constantly perturbs the labels, it is more challenging to learn the mapping. As a strong contrast, the testing MSE is barely impacted by the stochasticity up to $0.01$ standard deviation which is approximately $10\%$ of the average magnitudes. Instead of ascending, the testing MSE even decreases slightly. An explanation is that stochasticity actually serves as a regularization. This benefits the training of PFNO because the random noises prevent the optimizer from driving towards sharp minima. Hence, PFNO generalizes slightly better. 

\subsection{Computational efficiency}
One major motivation for applying deep learning models over traditional methods is their fast inference. In the case of a large-scale dataset, i.e., tens of thousands of velocity models, it is time-consuming to solve the Helmholtz equation with traditional methods. In comparison, FNO and PFNO predict a wavefield within several milliseconds on average. Nevertheless, the apparent drawback is that they demand hours for training which is the typical off-line efforts for DL-based scientific computing like PINN and DeepONet. Therefore, we take both the training and inference time into consideration when comparing the efficiency of FNO and PFNO with a traditional method, the ``optimal" 9-point finite difference method~\cite{jo1996optimal}.   

\begin{figure}[!ht]
    \centering
    \subfloat[]{\label{fig:efficiency_case1}
    \includegraphics[width= 0.49\textwidth]{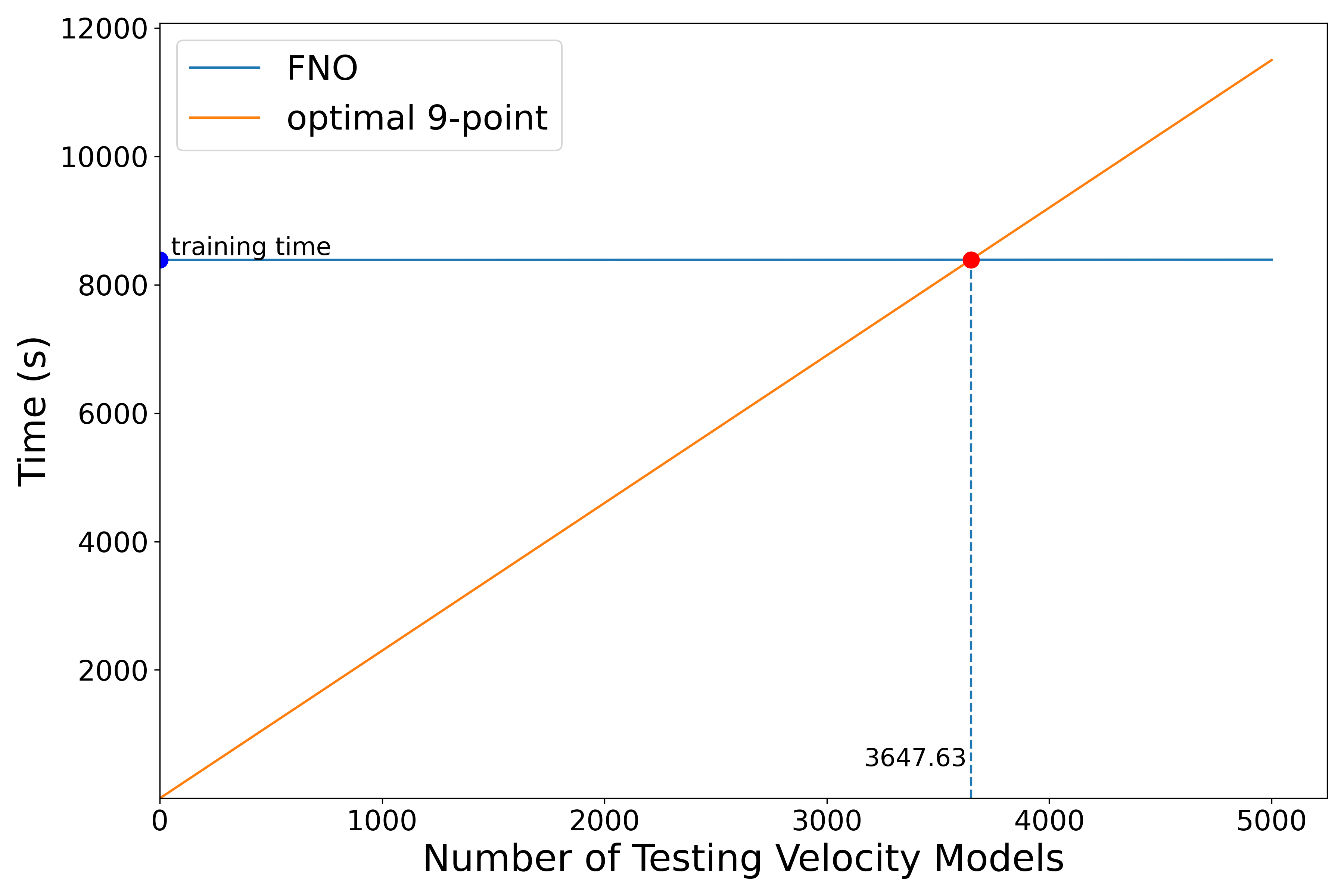}}
    \subfloat[]{\label{fig:efficiency_case2}
    \includegraphics[width= 0.49\textwidth]{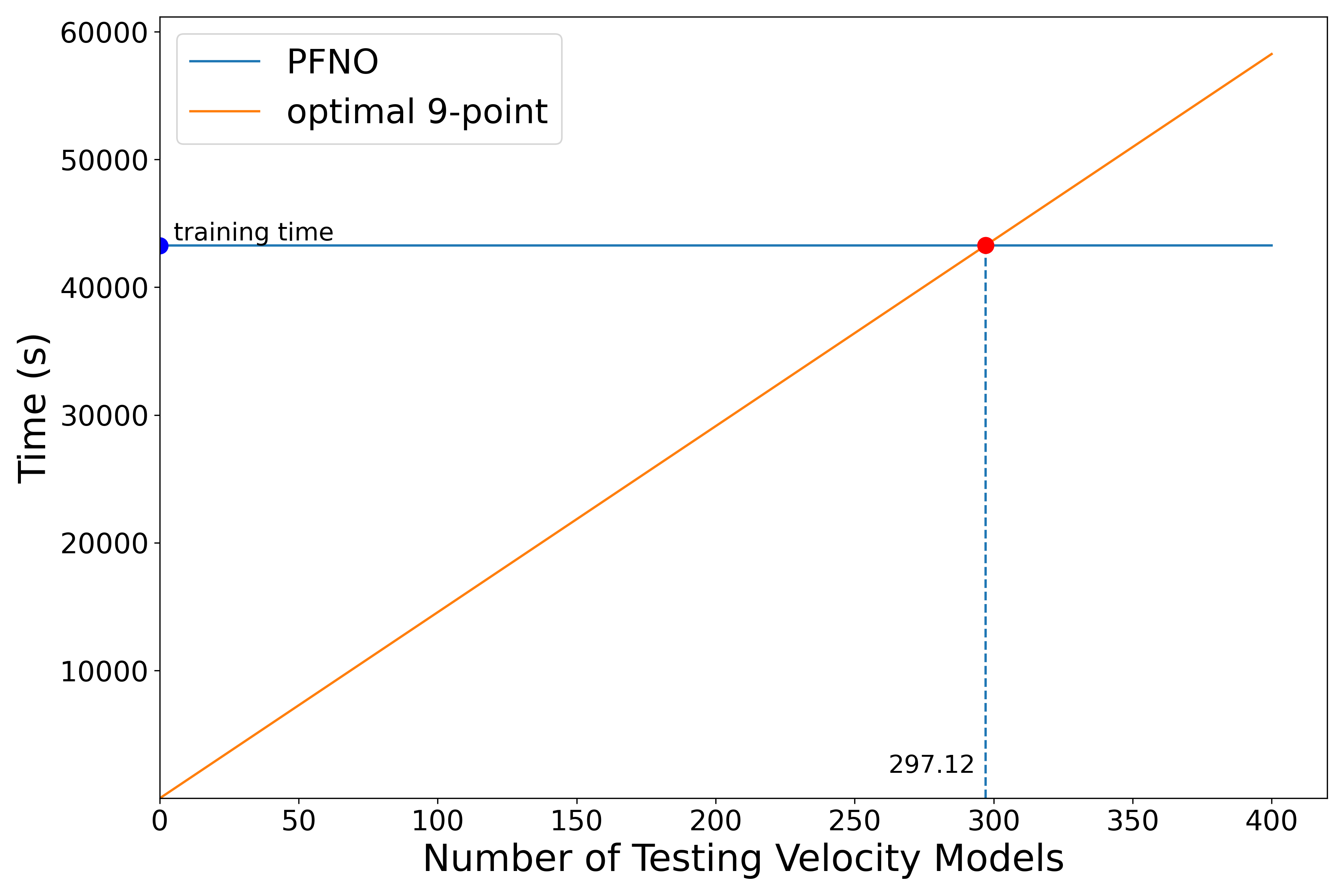}} 
    \caption{Efficiency comparison based on velocity models from \emph{CurveFault-A}: (a) case 1; (b) case 2; the training time of both FNO and PFNO corresponds to $15,000$ training samples of the velocity model.}
    \label{fig:efficiency}
\end{figure}

The comparison is performed in two settings: $1)$ single source location and single frequency; $2)$ $5$ source locations and $10$ frequencies. In the first case, an FNO is trained with 1 GPU, while we exploit $4$ GPUs to train a PFNO in the second case. However, both cases use only $1$ GPU for inference. Again, there are $15,000$ velocity models for training and $3,000$ for inference. In the inference, the batch size of both FNO and PFNO is set to $64$. Not surprisingly, FNO outputs a wavefield in 4.23e-04 seconds on average in case $1$. On the other hand, the ``optimal" 9-point method needs about $2.30$ seconds to solve the equation based on a velocity model. The speed advantage of PFNO is more significant with multiple sources and frequencies. For the multi-source and multi-frequency experiment setup on the same velocity model, PFNO averages 9.63e-03 seconds to predict the wavefield, while in contrast, the ``optimal" 9-point method takes $145.65$ seconds on average. Considering that FNO and PFNO took $2.33$ hours and $12.02$ hours for training, respectively, we consolidate them with their inference time, which makes the comparison more reasonable. As \Cref{fig:efficiency_case1} illustrates, FNO has a negligible marginal increase in the wall time as the number of velocity models increases, leading to an almost flat line. Moreover, FNO is more efficient if more than $3,648$ velocity models are to be approached, even though the training time is incorporated. A similar conclusion can be drawn from \Cref{fig:efficiency_case2}, only that PFNO is even more preferable. Provided that we have more than $298$ velocity models in the dataset, PFNO prevails over the traditional method in terms of speed. From the evidence above, one may verify the efficiency of FNO and PFNO in large-scale inferences. 

\begin{figure}[!ht]
    \centering
    \includegraphics[width= 0.7\textwidth]{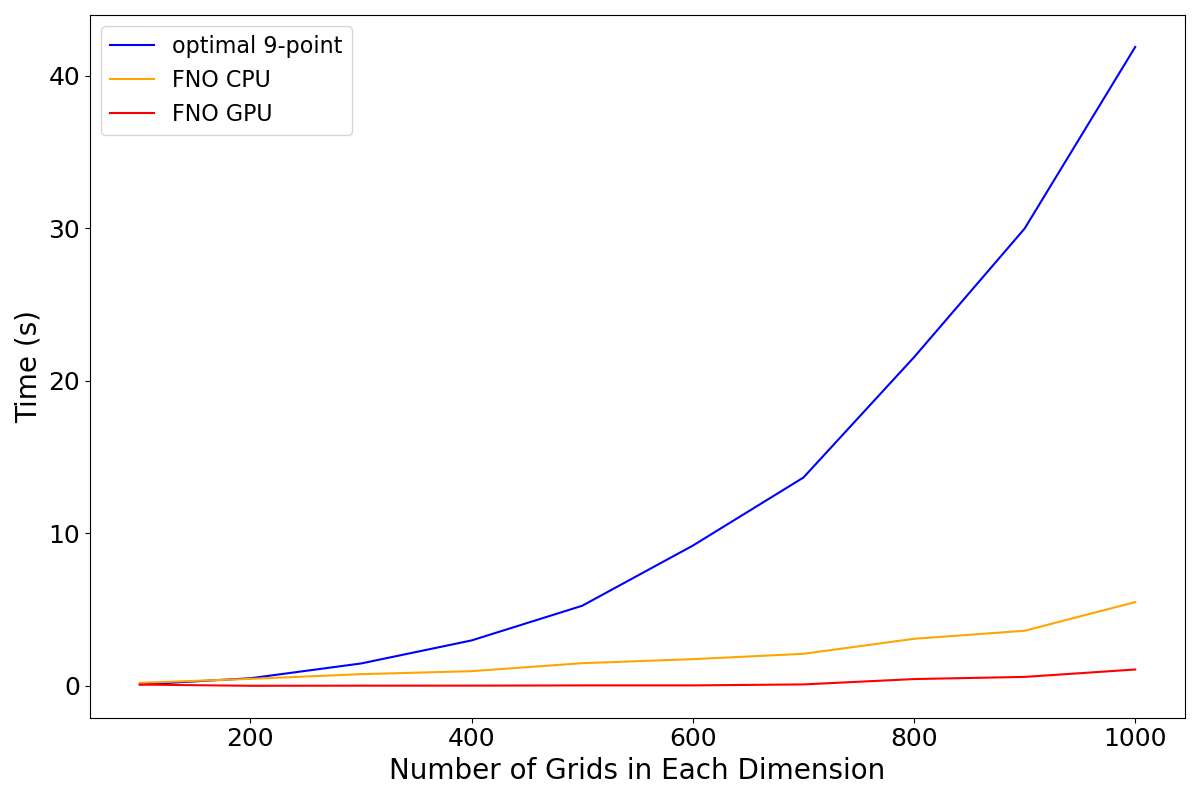}
    \caption{Scalable inference of FNO-based solvers: the domain size is quantified by the number of grids in each dimension. The running time of FNO under the CPU setting is also incorporated for a fair comparison with the ``optimal" 9-point method.}
    \label{fig:scale}
\end{figure}

Furthermore, in practice, the domain size also has an impact on the inference speed. By varying the domain size, we show that FNO dominates the ``optimal" 9-point method regarding the inference speed when $n_z = n_x \geq 200$. Unlike the multi-velocity-model setting above, this comparison only focuses $1$ velocity model, generated by interpolating an instance in \emph{CurveFault-A}. \Cref{fig:scale} reveals that the time of the ``optimal" 9-point method soars up with the increasing domain size. Although the theoretical time complexity of the method is $O(n^3)$, the sparsity and the special structure of the impedance matrix yields a almost quadratic complexity in practice. In contrast, the inference speed of the FNO-based solver remains almost constant under the GPU setting. 

%% file: conclusion.tex
\section{Conclusion}
In this work, the Fourier neural operator (FNO) is applied to efficiently solve the Helmholtz equation under the context of variable velocity. Furthermore, we extend FNO to the multi-source and multi-frequency setting and propose the paralleled Fourier neural operator (PFNO) that consists of a group of FNOs. The paralleled structure distributes the data of different frequencies to different FNOs such that each FNO concentrates on a single frequency but multiple sources. PFNO achieves high accuracy even on the complicated high-frequency wavefields, while FNO has less impressive performance. In the numerical experiments on the $6$ OpenFWI datasets, we demonstrate the accuracy of PFNO with wavefields visualization and show that it outperforms FNO and ForwardNet on all the $6$ datasets. Moreover, a trained PFNO is able to generalize to other datasets, provided that the velocity models share similar features. In terms of generalization on frequency, our experiments show only a weak ability to generalize to unseen frequencies, which might be a limitation of the current models. We will leave the realization of generalization on frequency to future work. In the case of noisy data, PFNO shows robustness against the noise. This strength makes PFNO reliable against measurement errors in practice. Finally, the ``optimal" 9-point method is compared with FNO in terms of their computational efficiency. FNO embraces fast inferences as expected, taking milliseconds on average to solve for the wavefield corresponding to a velocity model. Even when the training time is incorporated, PFNO still gains an edge over the ``optimal" 9-point method, given a large number of velocity models. In addition, the inference of the FNO-based is scalable in large spatial domains. 

With the strengths discussed above, FNO or PFNO can be a building block of an accurate and fast numerical solver for the Helmholtz equation when the model is expected to solve a large number of velocity models.

%% file: append.tex
\appendix

\section{Reconstructed Time Domain Wavefields}
\label{sec:time_domain_wavefield}
\begin{figure}[!ht]
  \centering
  \subfloat[]{
  \includegraphics[width= \textwidth, height= 0.22\textheight]{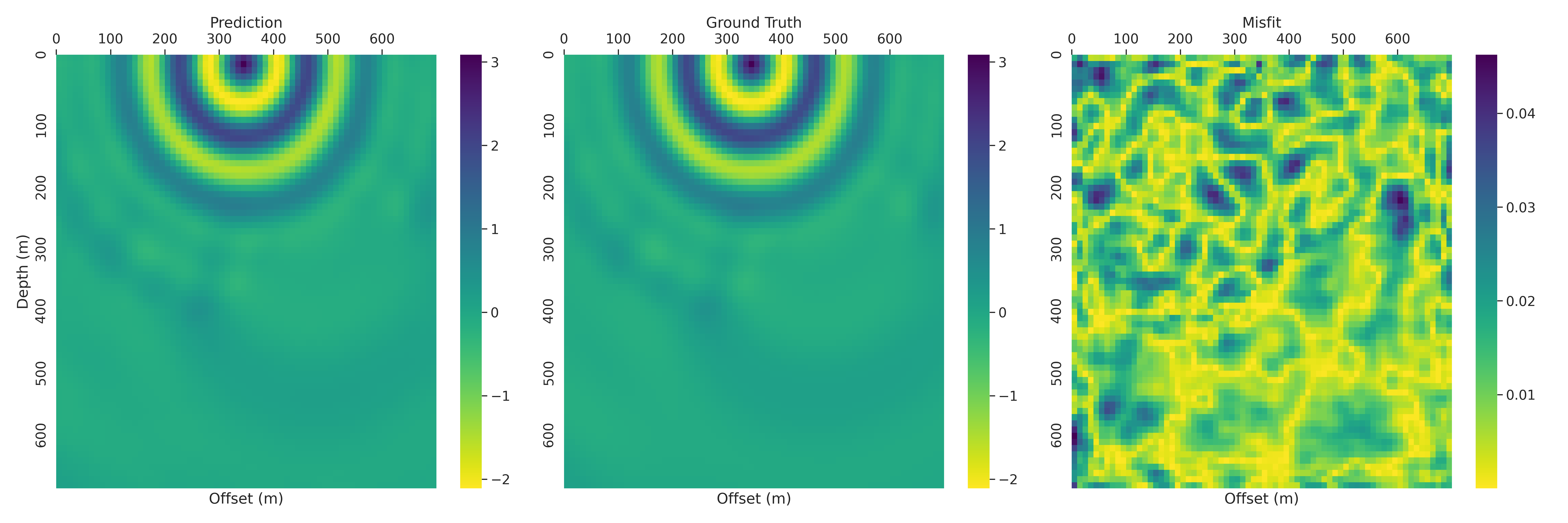}}\\
  \subfloat[]{
  \includegraphics[width= \textwidth, height= 0.22\textheight]{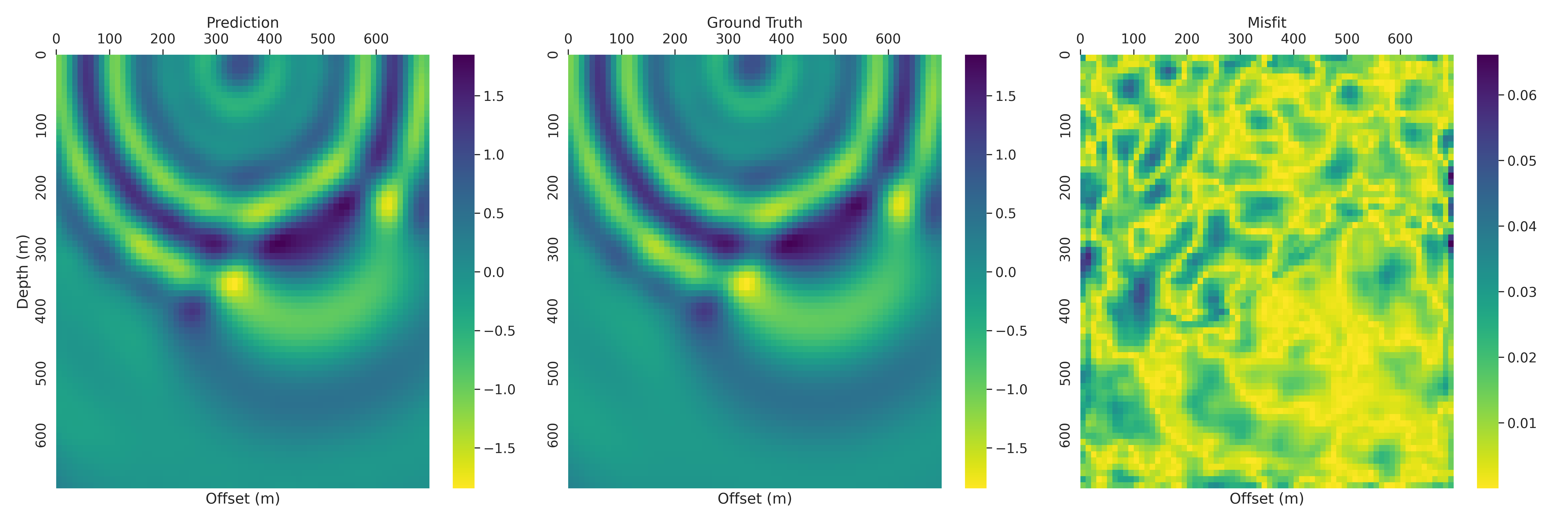}} \\
  \subfloat[]{
  \includegraphics[width= \textwidth, height= 0.22\textheight]{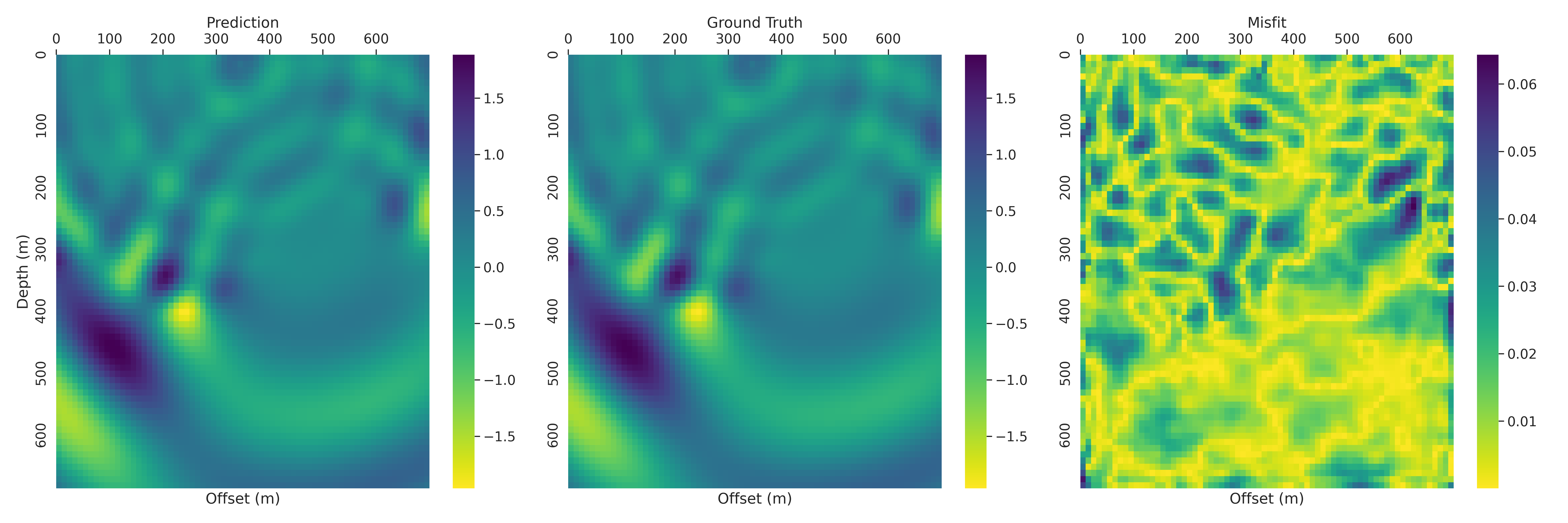}}
  \caption{Time domain wavefields reconstructed by inverse Fourier transform after frequency truncation: $(a)$ wavefields at $150$ millisecond; $(b)$ wavefields at $250$ millisecond; $(c)$ wavefields at $350$ millisecond. The ground truth is also obtained after frequency truncation. }
  \label{fig:wavefields_timeDomain}
\end{figure}

The wave propagation in the time domain describes how the energy spreads outward from the source in a straightforward way. Hence, we attempt to visualize the time domain wavefields at different time points. Since time domain wavefields can be obtained by applying inverse Fourier transform to frequency domain wavefields, we train a PFNO on data with a single source location and $10$ frequencies, $11 \sim 20$Hz. The frequencies are truncated  due to the limit of the computational resources although there should be $1000$ frequencies theoretically if the goal is to perfectly reconstruct the time domain wavefields with $\Delta t = 0.001$ seconds up to $1$ second. For a fair comparison, we filter out other frequencies in the ground truth as well. Thereafter, the predictions of the PFNO and truncated ground truth are transformed to the time domain and the real part of them is preserved, leading to the reconstructed time domain wavefields. As explained above, the reconstructed wavefields are not the same as the true time domain wavefields because of the missing frequencies. An example is given in \Cref{fig:wavefields_timeDomain}. As time elapses, the regular shape of the wave turns irregular as scatters show up. 

\begin{figure}[!ht]
  \centering
  \subfloat[]{
  \includegraphics[width= 0.7\textwidth, height= 0.26\textheight]{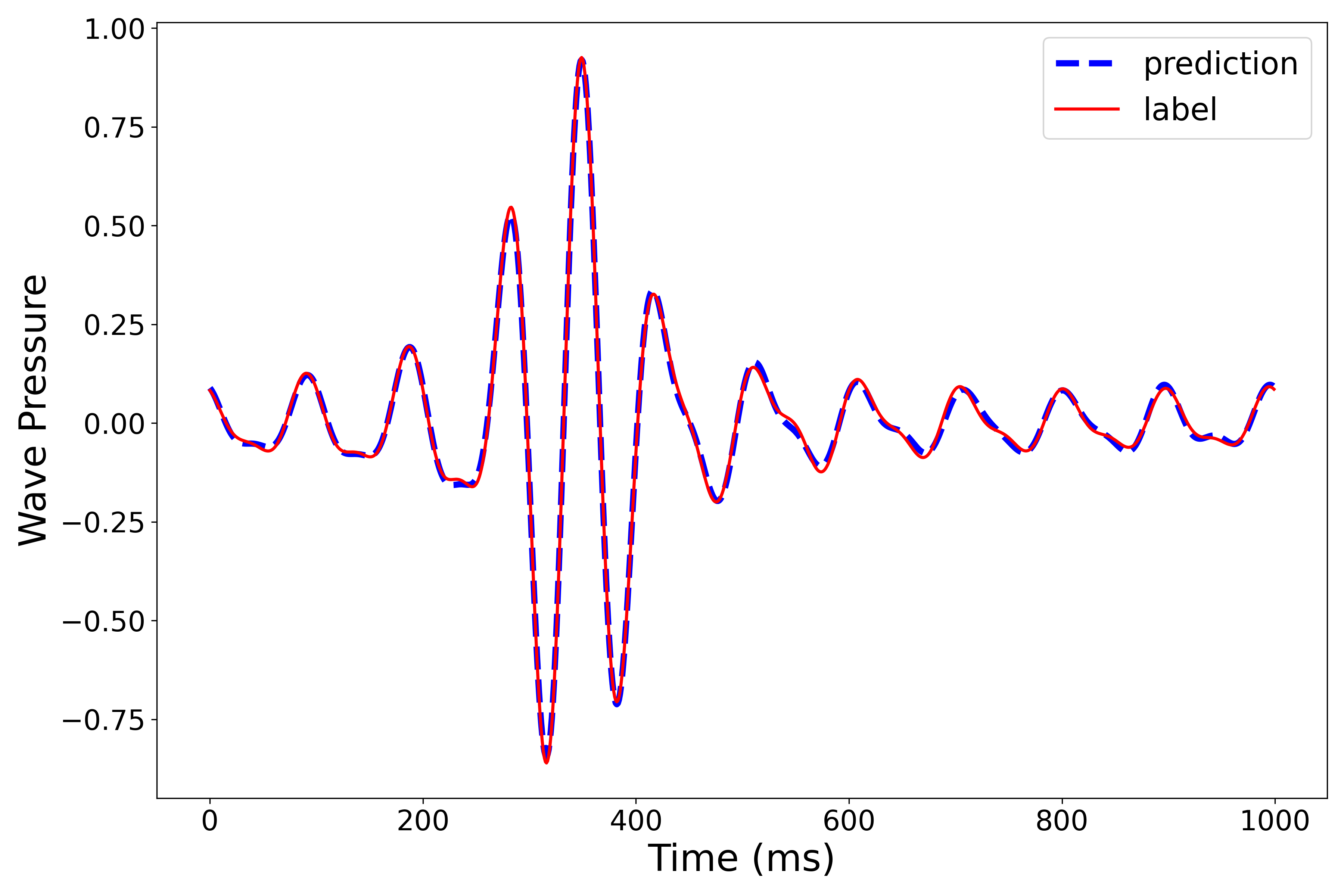}} \\
  \subfloat[]{
  \includegraphics[width= 0.7\textwidth, height= 0.26\textheight]{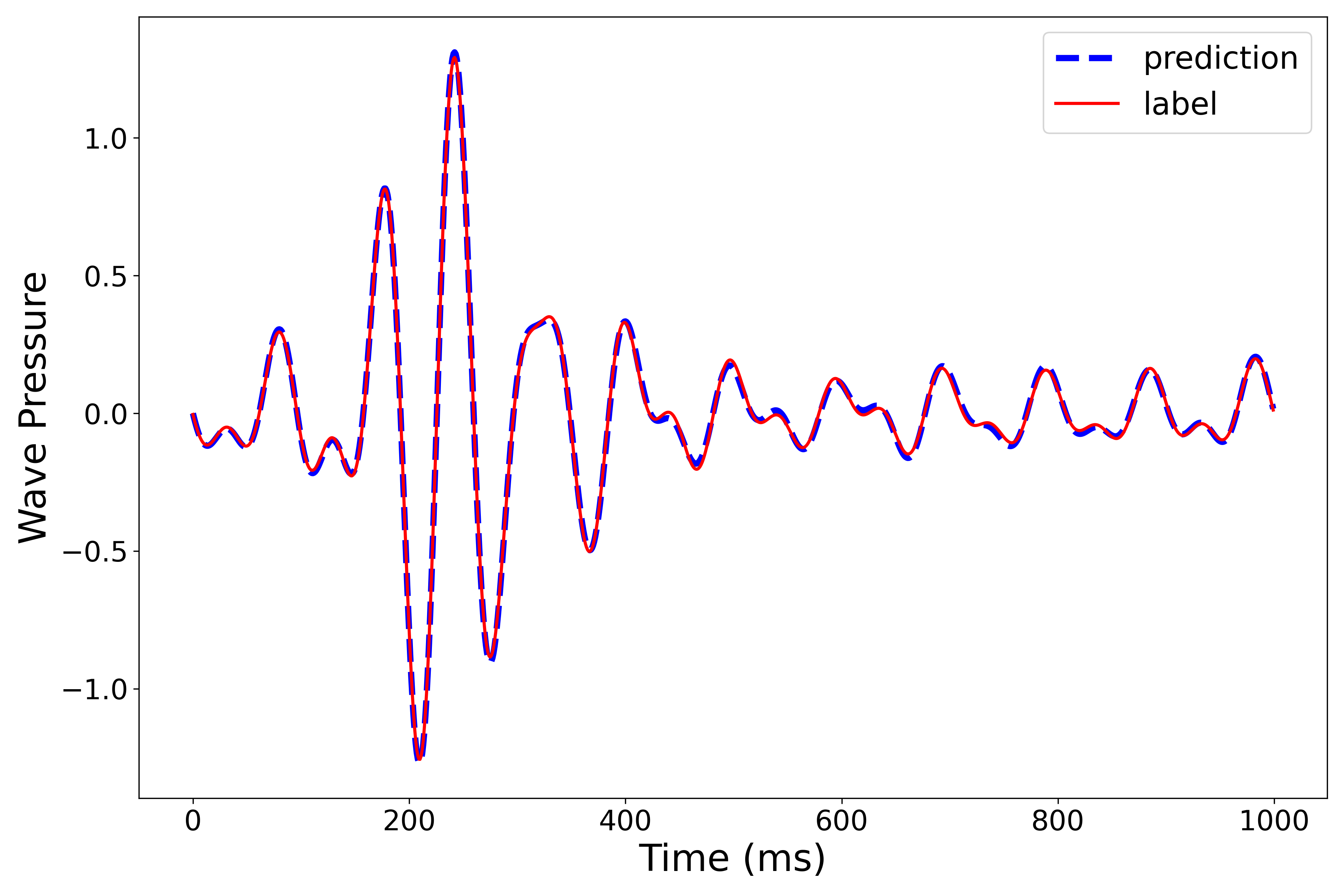}}
  \caption{Trace plot of the reconstructed wavefields at two locations: $(a)$ trace plot at location $(x, ~z) = (190, ~550)$; $(b)$ trace plot at location $(x, ~z) = (600, ~110)$.}
  \label{fig:wavefields_timeDomain_trace}
\end{figure}

To better observe the error in the time domain, we select $2$ locations, i.e, $(x, ~z) = (190, ~550)$ and $(x, ~z) = (600, ~110)$, for which the trace plot \Cref{fig:wavefields_timeDomain_trace} depicts the time-varying pattern of the wave pressure at these two locations. It clearly shows that the error remains at a low level in a long time span. 

\section{ForwardNet Architecture}
\label{sec:forwardnet_architecture}
The architecture of the encoder is shown in~\Cref{tab:encoder}, where conv stands for a convolutional layer. Each convolutional layer consists of a convolutional, a batch normalization~\cite{ioffe2015batch}, and a leaky ReLU activation function~\cite{nair2010rectified} with slope 0.2. Of note, the kernel size is fixed at 3 for each convolution. The columns in\_channel and out\_channel represent the number of channels applied in each layer. Also, we show the changes in data size throughout the layers in the last two columns. As a result, the latent representations have the dimension $512\times 1 \times 1$. 
\begin{table}[!ht]
  \centering
  \caption{Encoder}
    \begin{tabular}{ccccccc}
    \hline\hline
    layer & in\_channel & out\_channel & stride & padding  & in\_size & out\_size \\
    \midrule
    conv  & 3     & 32    & 2     & 1     & 70$\times$70 & 35$\times$35 \\
    conv  & 32    & 32    & 1     & 1     & 35$\times$35 & 35$\times$35 \\
    conv  & 32    & 64    & 2     & 1     & 35$\times$35 & 18$\times$18 \\
    conv  & 64    & 64    & 1     & 1     & 18$\times$18 & 18$\times$18 \\
    conv  & 64    & 128   & 2     & 1     & 18$\times$18 & 9$\times$9 \\
    conv  & 128   & 128   & 1     & 1     & 9$\times$9   & 9$\times$9 \\
    conv  & 128   & 256   & 2     & 1     & 9$\times$9   & 5$\times$5 \\
    conv  & 256   & 256   & 1     & 1     & 5$\times$5   & 5$\times$5 \\
    conv  & 256   & 512   & 2     & 1     & 5$\times$5   & 3$\times$3 \\
    conv  & 512   & 512   & 1     & 1     & 3$\times$3   & 3$\times$3 \\
    conv  & 512   & 512   & 2     & 0     & 3$\times$3   & 1$\times$1 \\
    \hline\hline
    \end{tabular}%
  \label{tab:encoder}%
\end{table}%

On the other hand, \Cref{tab:decoder} reveals the details of the decoder structure. The aforementioned convolutional layer is also a part of the building block in the decoder. In this table, the kernel is set in size 3 as well. The stride and the padding are hidden since they are fixed at 1 and 1, respectively, in all convolutions. Moreover, up\_conv stands for the combination of an up-sampling operation and a convolutional layer. To avoid the ``checkerboard effect", ForwardNet employs up-sampling instead of deconvolution (transposed convolution). The up-sampling operation is implemented with the nearest-value mode and has a parameter scale that controls the output size.  Besides up\_conv and conv, the crop layer removes the surrounding part of the intermediate representation and retrieves the middle component to obtain the 70$\times$70 size that matches the wavefield. Finally, the output channel is 2 because the wavefield is complex-valued. The two channels correspond to the real part and the imaginary part. 
\begin{table}[!ht]
  \centering
  \caption{Decoder}
    \begin{tabular}{cccccc}
    \hline\hline
    layer & in\_channel & out\_channel & scale & in\_size & out\_size \\
    \midrule
    up\_conv & 512   & 512   & 5     & 1$\times$1   & 5$\times$5 \\
    conv  & 512   & 512   &       & 5$\times$5   & 5$\times$5 \\
    up\_conv & 512   & 256   & 2     & 5$\times$5   & 10$\times$10 \\
    conv  & 256   & 256   &       & 10$\times$10 & 10$\times$10 \\
    up\_conv & 256   & 128   & 2     & 10$\times$10 & 20$\times$20 \\
    conv  & 128   & 128   &       & 20$\times$20 & 20$\times$20 \\
    up\_conv & 128   & 64    & 2     & 20$\times$20 & 40$\times$40 \\
    conv  & 64    & 64    &       & 40$\times$40 & 40$\times$40 \\
    up\_conv & 64    & 32    & 2     & 40$\times$40 & 80$\times$80 \\
    conv  & 32    & 32    &       & 80$\times$80 & 80$\times$80 \\
    crop  &       &       &       & 80$\times$80 & 70$\times$70 \\
    conv  & 32    & 2     &       & 70$\times$70 & 70$\times$70 \\
    \hline\hline
    \end{tabular}%
  \label{tab:decoder}%
\end{table}%

\section{Generalization on Frequency}
\label{sec:generalization_freq}
Since PFNO does not actually take frequency as a variable parameter, our experiments thereof target an FNO with a width of $128$. It is trained with multiple frequencies, and only $1$ source location is included in the training data to dampen the impact of the source location.

\begin{figure}[!ht]
    \centering
    \subfloat[]{\label{fig:gen_freq_3_4}
    \includegraphics[width= 0.32\textwidth, height= 0.2\textheight]{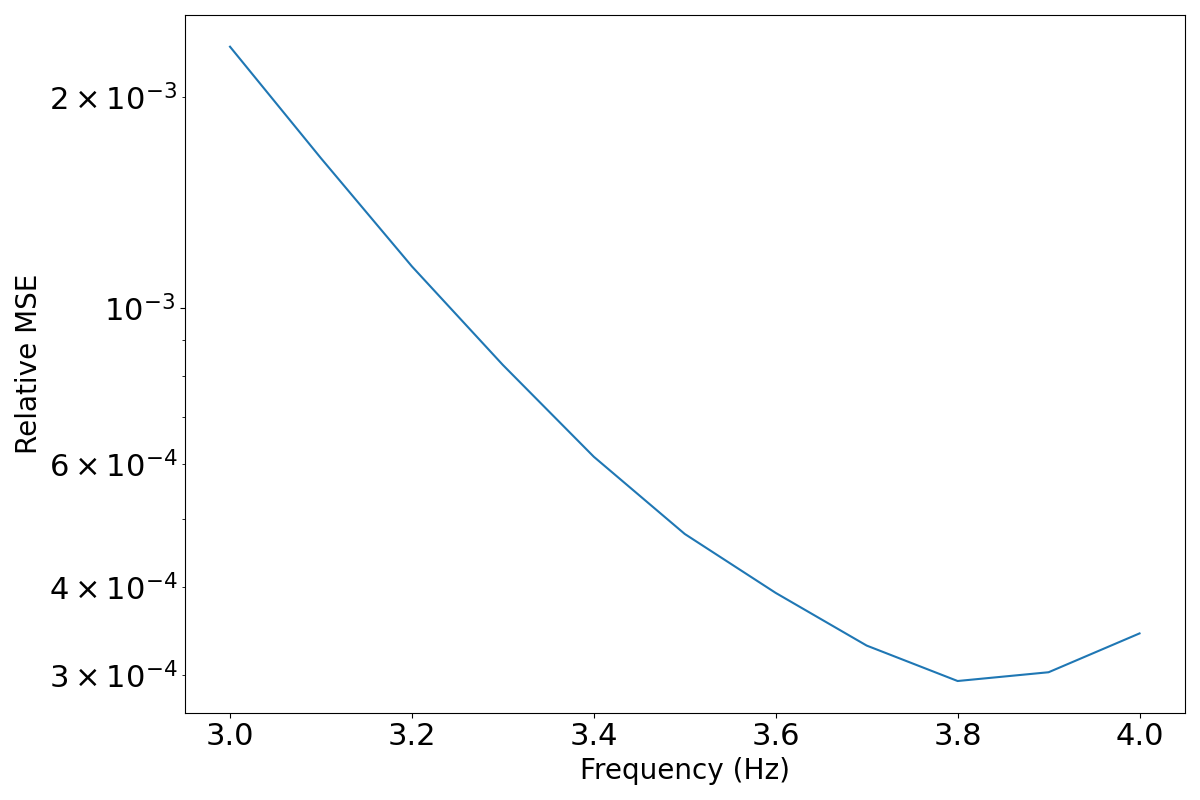}}
    \subfloat[]{\label{fig:gen_freq_9_10}
    \includegraphics[width= 0.32\textwidth, height= 0.2\textheight]{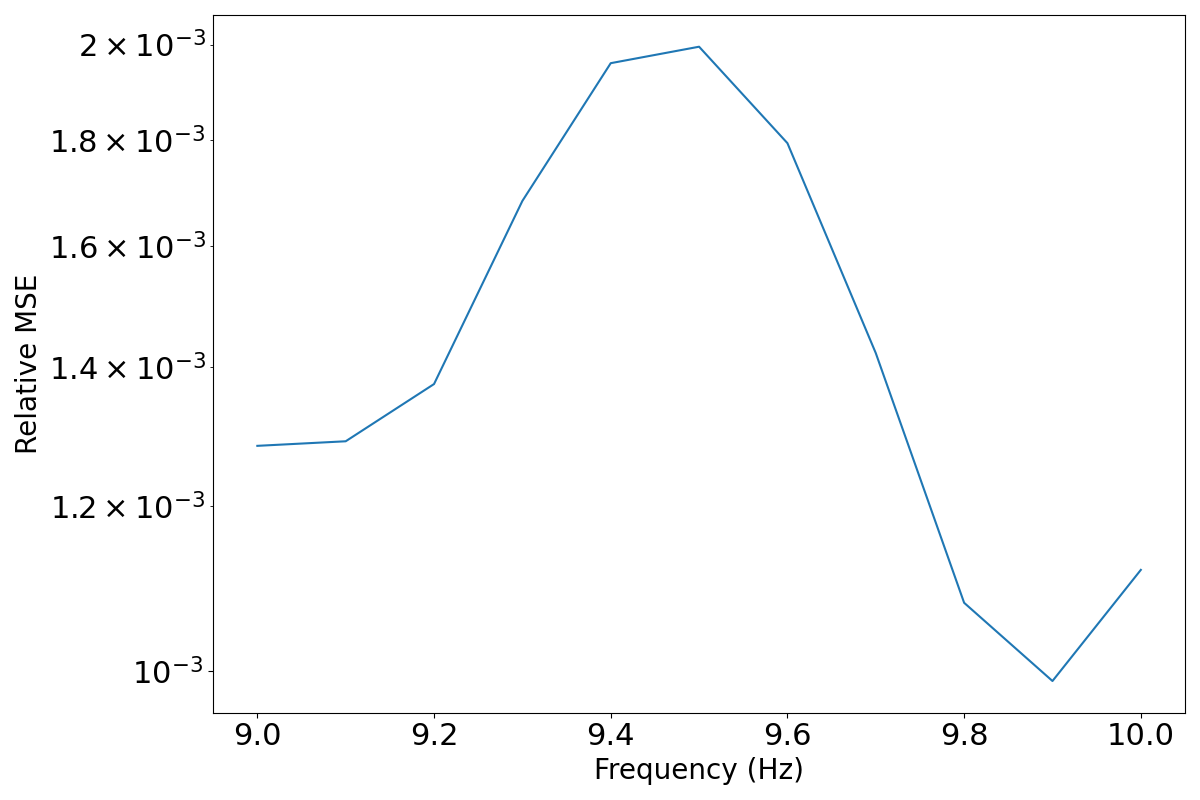}}
    \subfloat[]{\label{fig:gen_freq_15_16}
    \includegraphics[width= 0.32\textwidth, height= 0.2\textheight]{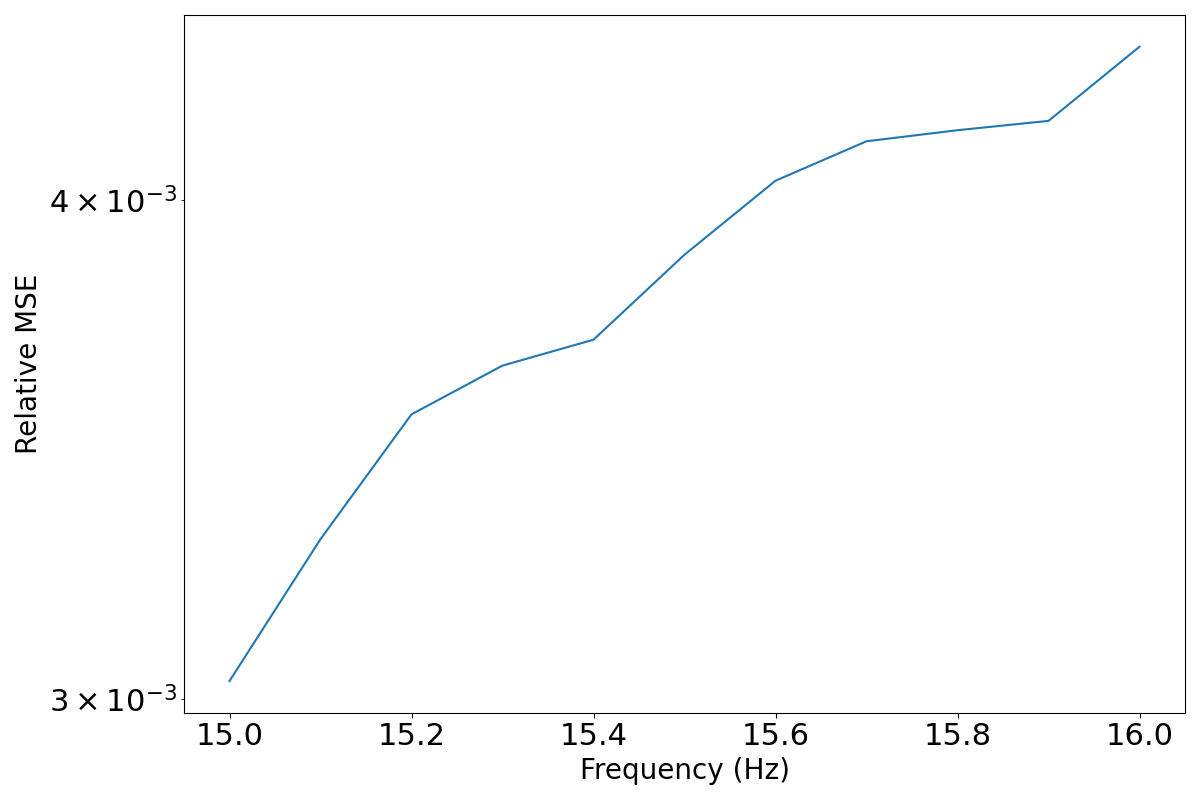}} \\
    \subfloat[]{\label{fig:gen_freq_21_22}
    \includegraphics[width= 0.32\textwidth, height= 0.2\textheight]{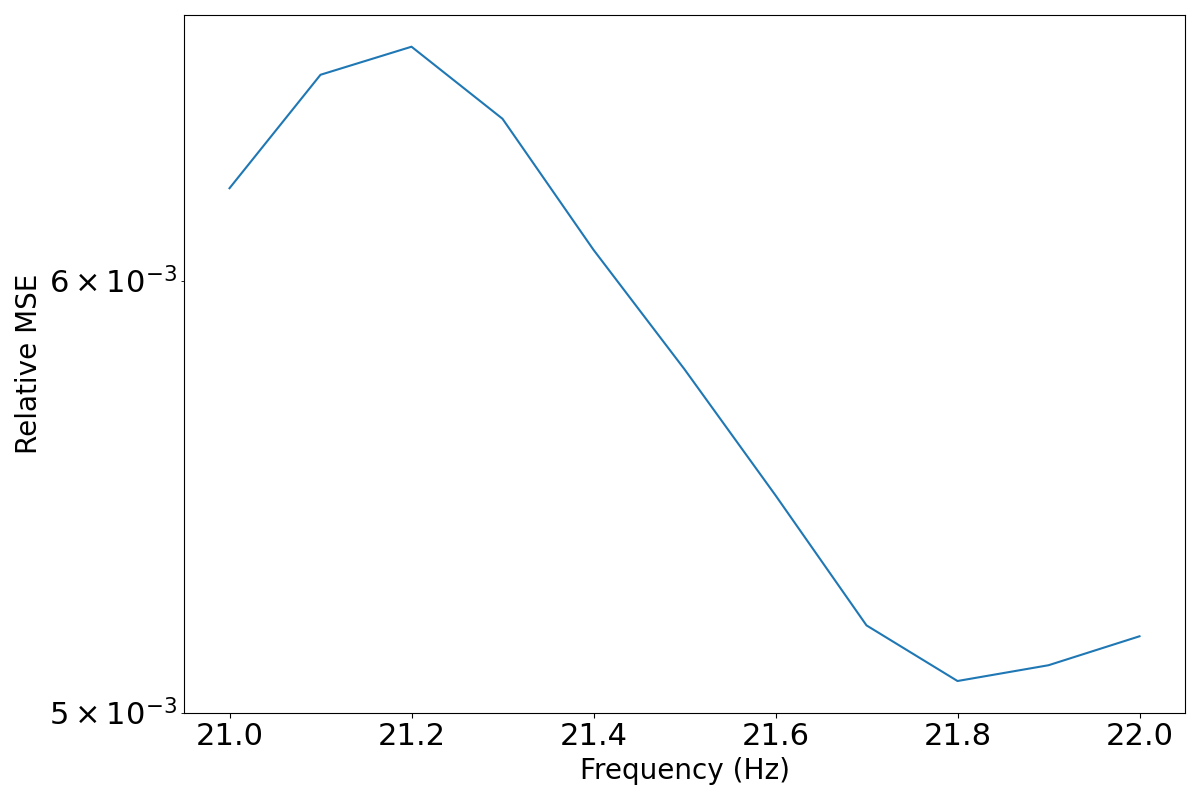}}
    \subfloat[]{\label{fig:gen_freq_27_28}
    \includegraphics[width= 0.32\textwidth, height= 0.2\textheight]{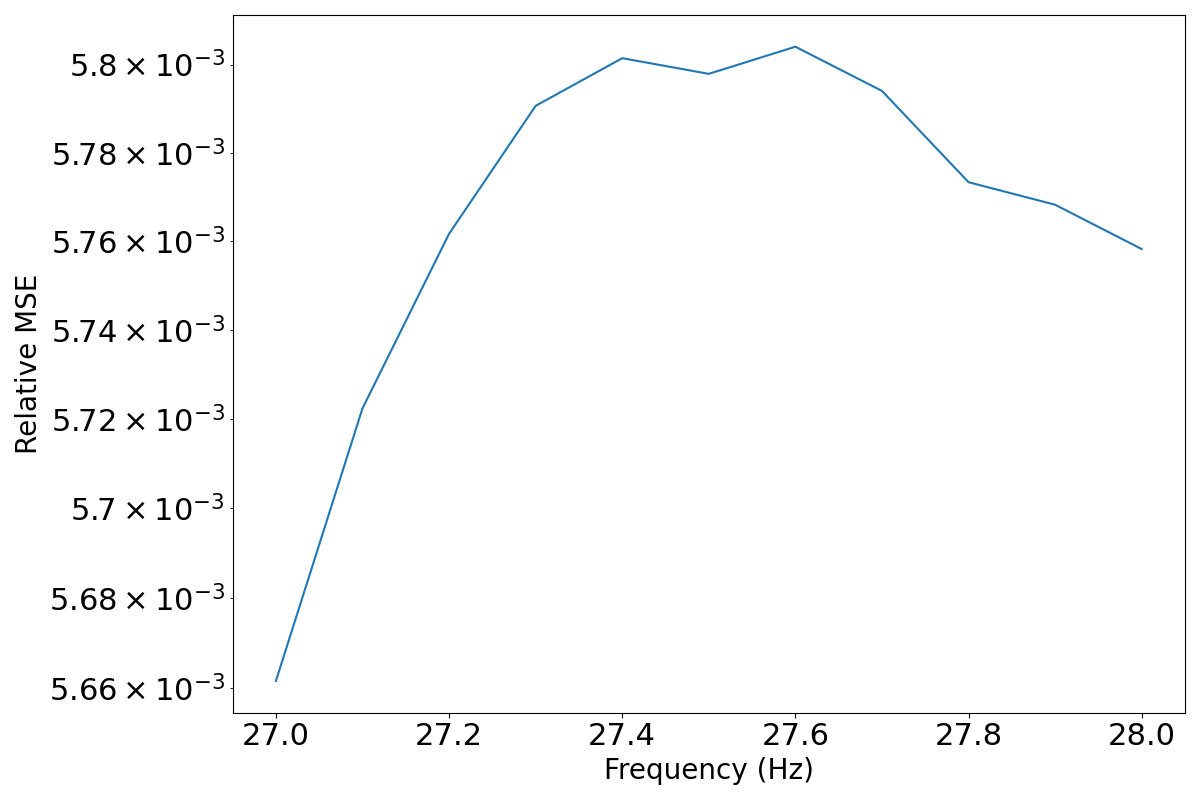}}
    \subfloat[]{\label{fig:gen_freq_33_34}
    \includegraphics[width= 0.32\textwidth, height= 0.2\textheight]{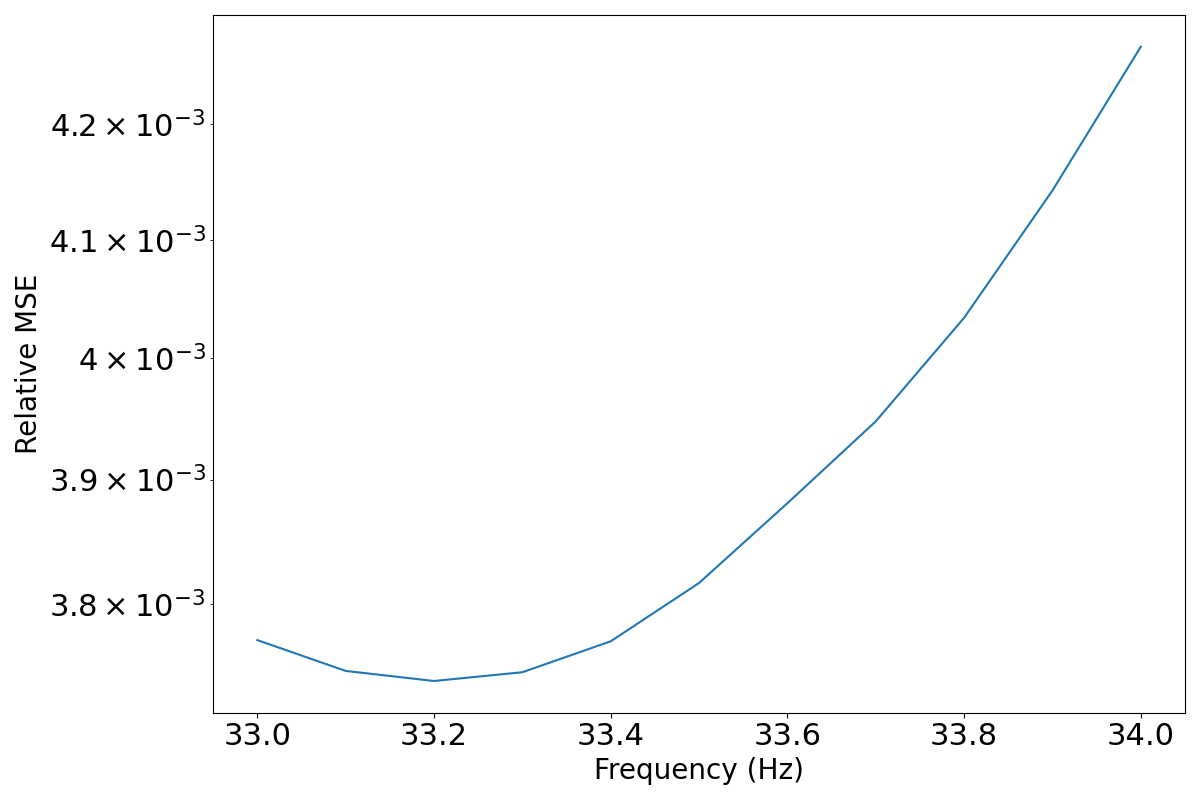}}
    \caption{Generalization on frequency: (a) and (f) examine the extrapolation of the FNO, while the others examine its interpolation. The relative MSE is computed by dividing MSE by the average magnitude of the wavefield.}
    \label{fig:gen_freq}
\end{figure}

More specifically, the FNO is trained on the \emph{CurveFault-A} data of $4$Hz $\sim$ $33$Hz with interval $1$Hz. However, the interval is set to be $0.1$Hz in the testing since we only expect the testing results to be accurate near the training frequencies. In \Cref{fig:gen_freq}, we test whether the FNO generalizes to neighbouring frequencies by measuring the relative MSE corresponding to the wavefields of different frequencies. In particular, $\text{relative MSE} = \text{MSE} / \text{Avg}(\vert \bm u \vert)$ is computed to exclude the impact of the magnitude of the wavefields across the frequencies. Since only $4$Hz belongs to the training frequencies, \Cref{fig:gen_freq_3_4} illustrates how FNO generalizes towards lower frequencies. Similarly, \Cref{fig:gen_freq_33_34} shows the generalization towards higher frequencies. It is not surprising to see in decrease in accuracy if we move far away from the boundary. In comparison, the two boundary frequencies in  \Cref{fig:gen_freq_9_10,fig:gen_freq_15_16,fig:gen_freq_21_22,fig:gen_freq_27_28} are training frequencies. Hence, the error is anticipated to increase as we move towards the middle, and a bell-shaped relative MSE curve should occur. Although \Cref{fig:gen_freq_15_16} turns out to be an exception, the relative MSE remains smaller than that of the right boundary.  
